\documentclass[conference]{IEEEtran}
\usepackage{iftex}
\ifXeTeX
  \usepackage{fontspec}
\else
  \usepackage{times}
\fi
\usepackage[numbers]{natbib}
\usepackage{multicol}
\usepackage{graphicx}
\usepackage{booktabs}
\usepackage{amsmath,amssymb}
\usepackage{amsthm}
\usepackage{algorithm}
\usepackage{algpseudocode}
\usepackage{xcolor}
\ifdefined\UseExternalMain
  \usepackage{xr-hyper}
\fi
\usepackage[bookmarks=true]{hyperref}
\hypersetup{%
  hidelinks,
  pdftitle={Feedback Completion Certificates for Repeated Dwell Tasks with Time Windows},
  pdfauthor={},
  pdfsubject={},
  pdfkeywords={},
  pdfcreator={}}

\newif\ifdraft
\draftfalse
\newif\ifshowproofsketches
\showproofsketchesfalse
\newif\ifcomputationtable
\computationtabletrue

\theoremstyle{plain}
\newtheorem{theorem}{Theorem}
\newtheorem{lemma}[theorem]{Lemma}
\newtheorem{proposition}[theorem]{Proposition}
\newtheorem{corollary}[theorem]{Corollary}
\theoremstyle{definition}
\newtheorem{definition}[theorem]{Definition}
\newtheorem{assumption}{Assumption}
\newtheorem{problem}{Problem}
\theoremstyle{remark}
\newtheorem{remark}[theorem]{Remark}
\theoremstyle{plain}
\newtheorem{slemma}{Lemma}

\newtheorem{scorollary}[slemma]{Corollary}
\newtheorem{sproposition}[slemma]{Proposition}

\newenvironment{proofsketch}{\begin{IEEEproof}[Proof sketch]}{\end{IEEEproof}}

\newenvironment{restated}[2]{%
  \par\addvspace{\medskipamount}\noindent
  \textbf{#1~\ref{#2} (restated).}\ \itshape\ignorespaces}{%
  \par\addvspace{\medskipamount}}

\newcommand{\R}{\mathbb{R}}
\newcommand{\Dset}{\mathbb{D}}
\newcommand{\Ball}{\mathbb{B}_\varepsilon}
\newcommand{\G}{\mathbf{G}}
\newcommand{\F}{\mathbf{F}}
\newcommand{\UntilOp}{\mathbin{\mathbf{U}}}
\newcommand{\dist}{\operatorname{dist}_\infty}
\newcommand{\hrz}{\operatorname{hrz}}
\newcommand{\ybar}{\bar{y}}
\newcommand{\ytil}{\tilde{y}}
\newcommand{\Cov}{\mathcal{C}}
\newcommand{\Qset}{\mathcal{Q}}
\newcommand{\cbar}{\bar{c}}
\newcommand{\qinit}{q^{\mathrm{init}}}
\newcommand{\lo}{\mathrm{lo}}
\newcommand{\hi}{\mathrm{hi}}
\newcommand{\Acc}{\mathrm{Acc}}
\newcommand{\Adm}{\mathrm{Adm}}
\newcommand{\RA}{\mathrm{RA}}
\newcommand{\Win}{\mathrm{Win}}
\newcommand{\Pre}{\mathrm{Pre}}
\newcommand{\Tset}{\mathcal{T}}
\newcommand{\That}{\widehat{\mathcal{T}}}
\newcommand{\Zpost}{\mathcal{Z}_{\mathrm{post}}}
\newcommand{\Lib}{\mathcal{L}}
\newcommand{\Fex}{\mathcal{F}_{\mathrm{exact}}}
\newcommand{\Fsh}{\mathcal{F}_{\mathrm{shared}}}
\newcommand{\Fsk}{\mathcal{F}_{\mathrm{shrink}}}
\newcommand{\Fsi}{\mathcal{F}_{\mathrm{single}}}
\newcommand{\Kpost}{\mathcal{K}_{\mathrm{post}}}
\newcommand{\Succ}{\mathrm{Succ}}
\newcommand{\Blib}{\mathcal{B}}
\newcommand{\PreA}{\mathrm{Pre}^{\forall}}

\newcommand{\EOneRuns}{10}
\newcommand{\EOneForcedOut}{5--12}
\newcommand{\EOneCFFail}{22}
\newcommand{\EOneTailK}{6}
\newcommand{\EOneTailTau}{16}
\newcommand{\EOneTailCTau}{1}
\newcommand{\EOneTailClear}{22}

\newcommand{\EOneTailSeed}{403}
\newcommand{\EOneTailPool}{53}
\newcommand{\ETwoQueries}{60}
\newcommand{\ETwoFeasible}{60}
\newcommand{\ETwoPatches}{4}
\newcommand{\ETwoRegionsBefore}{23}
\newcommand{\ETwoRegionsAfter}{143}
\newcommand{\ETwoPatchS}{64}
\newcommand{\ETwoPatchCertS}{54}
\newcommand{\ETwoLibraryS}{5.6}
\newcommand{\ETwoLibraryFullS}{28}
\newcommand{\ETwoMaxDPMs}{40}
\newcommand{\ETwoStoredFullOrdered}{11487}
\newcommand{\ETwoStoredFullBitset}{15742}
\newcommand{\EThreeUpdateK}{10}
\newcommand{\EThreeOutage}{[30,45)}
\newcommand{\EThreeOutageTicks}{15}
\newcommand{\EThreeBudgetS}{1}
\newcommand{\EThreeUsefulDelta}{1}
\newcommand{\EThreeLookOurs}{2}
\newcommand{\EThreeLookFull}{6}
\newcommand{\EThreeLookWTwo}{2}
\newcommand{\EThreeLimitOursMs}{400}
\newcommand{\EThreeLimitFullMs}{800}
\newcommand{\EThreeLimitWTwoMs}{150}
\newcommand{\EThreeOutcomesOurs}{431 ready, 322 superseded, 113 stale, 81 skipped in queue, 3 late}
\newcommand{\EThreeOutcomesFull}{446 skipped in queue, 210 ready, 198 over budget, 56 superseded}
\newcommand{\EThreeOutcomesWTwo}{556 ready, 226 stale, 145 skipped in queue, 20 superseded, 3 late}
\newcommand{\EThreeConfigs}{10}
\newcommand{\EThreeSwitchable}{5}
\newcommand{\EThreeFullOurs}{\ensuremath{-21}}
\newcommand{\EThreeFullOursLo}{\ensuremath{-26}}
\newcommand{\EThreeFullOursHi}{\ensuremath{-16}}

\newcommand{\EThreeWTwoOurs}{\ensuremath{67}}
\newcommand{\EThreeWTwoOursLo}{\ensuremath{62}}
\newcommand{\EThreeWTwoOursHi}{\ensuremath{72}}

\newcommand{\EThreeWTwoOursUp}{\ensuremath{118}}
\newcommand{\EThreeWTwoOursLow}{\ensuremath{17}}
\newcommand{\EThreeNoSolveOurs}{\ensuremath{118}}
\newcommand{\EThreeNoSolveOursLo}{\ensuremath{115}}
\newcommand{\EThreeNoSolveOursHi}{\ensuremath{121}}

\newcommand{\EThreeOutageOurs}{\ensuremath{2.8}}
\newcommand{\EThreeOutageOursLo}{\ensuremath{-0.8}}
\newcommand{\EThreeOutageOursHi}{\ensuremath{5.6}}

\newcommand{\EThreeOutageFull}{\ensuremath{-2.0}}
\newcommand{\EThreeOutageFullLo}{\ensuremath{-8.3}}
\newcommand{\EThreeOutageFullHi}{\ensuremath{4.2}}

\newcommand{\EThreeUsefulOursMin}{7}
\newcommand{\EThreeUsefulOursMax}{24}

\newcommand{\EThreeRespOursMin}{2}

\newcommand{\EThreeUsefulWTwoMin}{31}
\newcommand{\EThreeUsefulWTwoMax}{37}

\newcommand{\EThreeShFullOurs}{\ensuremath{-33}}
\newcommand{\EThreeShFullOursLo}{\ensuremath{-35}}
\newcommand{\EThreeShFullOursHi}{\ensuremath{-29}}

\newcommand{\EThreeShWTwoOurs}{\ensuremath{22}}
\newcommand{\EThreeShWTwoOursLo}{\ensuremath{-7}}
\newcommand{\EThreeShWTwoOursHi}{\ensuremath{51}}

\newcommand{\EThreeShNoSolveOurs}{\ensuremath{108}}
\newcommand{\EThreeShNoSolveOursLo}{\ensuremath{107}}
\newcommand{\EThreeShNoSolveOursHi}{\ensuremath{109}}

\newcommand{\EThreeUsefulFullFirst}{8}

\newcommand{\EThreeBeffOursMs}{371}
\newcommand{\EThreeBeffFullMs}{1000}

\newcommand{\EThreeShUsefulFullFirst}{5}

\newcommand{\EThreeCostlierOurs}{84}
\newcommand{\EThreeCostlierOursOf}{431}
\newcommand{\EThreeCostlierOursPct}{19}
\newcommand{\EThreeCostlierFull}{7}
\newcommand{\EThreeCostlierFullOf}{210}
\newcommand{\EThreeCostlierFullPct}{3}
\newcommand{\EThreeCostlierWTwo}{19}
\newcommand{\EThreeCostlierWTwoOf}{556}

\newcommand{\EThreeShCostlierOurs}{74}
\newcommand{\EThreeShCostlierOursOf}{399}

\newcommand{\RobotLateMs}{13.6}
\newcommand{\RobotLateCandidates}{11}
\newcommand{\RobotLateInputs}{1}
\newcommand{\RobotTicks}{11550}
\newcommand{\RobotWorkMs}{54}

\newcommand{\IfRuns}{21}
\newcommand{\IfSweep}{14}
\newcommand{\PlatThrustPct}{76}
\newcommand{\PlatMomentPct}{94}
\newcommand{\PlatMomentMaxNm}{0.0056}

\newcommand{\PlatAtLimit}{0}
\newcommand{\ETwoNFiveBefore}{63}
\newcommand{\ETwoNFiveBeforeMs}{45}
\newcommand{\ETwoNFiveAfter}{100}
\newcommand{\ETwoNFiveAfterMs}{139}
\newcommand{\ETwoNTenBefore}{100}
\newcommand{\ETwoNTenBeforeMs}{58}
\newcommand{\ETwoNTenAfter}{100}
\newcommand{\ETwoNTenAfterMs}{152}
\newcommand{\ETwoNThirtyBefore}{100}
\newcommand{\ETwoNThirtyBeforeMs}{87}
\newcommand{\ETwoSTLBefore}{100}
\newcommand{\ETwoSTLBeforeMs}{37}

\newcommand{\ROneAllPass}{all four}
\newcommand{\ROneEdges}{27{,}076}

\newcommand{\ROneKpost}{378}
\newcommand{\ROneKpostCand}{432}

\newcommand{\ROneH}{19}
\newcommand{\ROneD}{2}
\newcommand{\ROneA}{13}
\newcommand{\ROneB}{16}
\newcommand{\ROneEll}{1}
\newcommand{\ROneQzero}{(3,0)}
\newcommand{\ROneRollouts}{200}
\newcommand{\ROneCompleted}{200}
\newcommand{\ROneEvidenceK}{0}
\newcommand{\ROneEvidencePrim}{\ensuremath{v{=}1,\,\theta_{\rm ref}{=}-1/4}}

\newcommand{\ROneAOne}{\ensuremath{\{v{=}1,\,\theta_{\rm ref}{=}1/4\}}}
\newcommand{\ROneATwo}{\ensuremath{\{v{=}1,\,\theta_{\rm ref}{=}-1/4\}}}

\newcommand{\RTwoOrdS}{3.52}
\newcommand{\RTwoOrdMB}{39.8}
\newcommand{\RTwoCompS}{3.21}
\newcommand{\RTwoCompMB}{4.6}
\newcommand{\RTwoOrdTwoS}{6.04}
\newcommand{\RTwoOrdTwoMB}{42.6}
\newcommand{\RTwoCompTwoS}{3.9}
\newcommand{\RTwoCompTwoMB}{9.7}
\newcommand{\RTwoSingleEdges}{0}
\newcommand{\RTwoNominalSingle}{0}
\newcommand{\RTwoFullCertified}{650}
\newcommand{\RTwoSingleCertified}{0}
\newcommand{\RTwoBranchHist}{4, 6 or 9}

\newcommand{\RThreeConfigs}{20}

\newcommand{\RThreeNominalLost}{5}

\newcommand{\RThreeRobustInterv}{100}
\newcommand{\RThreeNominalInterv}{21}
\newcommand{\RThreeSafetyInterv}{20}
\newcommand{\RThreeBudgetMs}{100}

\newcommand{\RThreeOverruns}{0}

\newcommand{\RThreeDistPhys}{9}
\newcommand{\RThreeDistLostK}{8}
\newcommand{\RThreeUncLostK}{5}

\newcommand{\RThreeOutFallback}{7}

\newcommand{\RThreeBranchTicks}{324}
\newcommand{\RThreeDistComplNominal}{2}
\newcommand{\RThreeDistComplSafety}{1}
\newcommand{\PartWbar}{0.1}
\newcommand{\PartA}{13}
\newcommand{\PartRbarWidth}{2}
\newcommand{\PartLHS}{2.6}
\newcommand{\RAdmRegions}{1{,}404}

\newcommand{\RFourEdges}{224{,}094}
\newcommand{\RFourKpost}{1{,}936}
\newcommand{\RFourKpostCand}{3{,}600}

\newcommand{\RFourBuildS}{134}
\newcommand{\RFourDPs}{41}
\newcommand{\RFourA}{14}

\newcommand{\RFourH}{21}
\newcommand{\RFourD}{2}
\newcommand{\RFourSearchLo}{10}
\newcommand{\RFourSearchFail}{13}
\newcommand{\RFourSeeds}{10}
\newcommand{\RFourGateTwo}{10}
\newcommand{\RFourCompleted}{10}
\newcommand{\RFourSafe}{10}
\newcommand{\RFourPost}{10}
\newcommand{\RFourResidualOK}{10}

\newcommand{\RFourMaxWp}{35}
\newcommand{\RFourMaxWnu}{34}
\newcommand{\RFourBDTicks}{150}
\newcommand{\RFourBDOf}{210}
\newcommand{\RFourKeys}{21}

\newcommand{\RFourFineFound}{no}

\newcommand{\RFourWitnessK}{12}
\newcommand{\RFourMujoco}{3.14.0}
\newcommand{\RFourPostTicks}{5}
\newcommand{\RPairPoints}{210}

\newcommand{\RPairCands}{710}
\newcommand{\RPairUnreach}{702}
\newcommand{\RPairBoundary}{0}
\newcommand{\RPairUnknown}{0}

\newcommand{\RPairWit}{8}

\newcommand{\RPairWitK}{5}
\newcommand{\RPairWitSeeds}{4}
\newcommand{\RPairMarginMinMM}{0.22}
\newcommand{\RPairMarginMaxMM}{4.5}

\newcommand{\RPairPrim}{\ensuremath{v_{\rm ref}{=}}(1/4,0)}

\newcommand{\RPairAttempts}{16}

\newcommand{\RPairCompleted}{16}

\newcommand{\RPairLandDistinct}{8}
\newcommand{\RPairLandDisjoint}{0}
\newcommand{\RPairLandNested}{8}
\newcommand{\RPairMissMinMM}{23}
\newcommand{\RPairMissMaxMM}{27}

\newcommand{\EncMonVars}{8}
\newcommand{\EncMonRows}{22}
\newcommand{\EncMonCases}{3660}
\newcommand{\EncQueries}{60}
\newcommand{\EncOptionsTen}{23}
\newcommand{\EncTenVars}{183}
\newcommand{\EncTenRows}{1259}
\newcommand{\EncTenMonVars}{70}
\newcommand{\EncTenMonRows}{200}

\newcommand{\ExtSTLMs}{37}
\newcommand{\ExtSTLMainMs}{37}
\newcommand{\ExtRamMs}{176}
\newcommand{\ExtRamPninetyfiveMs}{287}
\newcommand{\ExtRamVer}{60}

\newcommand{\ExtDen}{60}
\newcommand{\ExtRamSlower}{60}
\newcommand{\ExtPaired}{60}

\newcommand{\RFiveKstar}{12}
\newcommand{\RFiveXstar}{\ensuremath{(\tfrac{25}{4},\tfrac{15}{4},0)}}
\newcommand{\RFiveMargin}{\ensuremath{\tfrac{1}{8}}}
\newcommand{\RFiveA}{3}
\newcommand{\RFiveB}{19}
\newcommand{\RFiveEll}{1}
\newcommand{\RFiveD}{2}
\newcommand{\RFiveH}{28}
\newcommand{\RFiveRegions}{3\,240}
\newcommand{\RFiveEdges}{75\,327}
\newcommand{\RFiveKpost}{313}
\newcommand{\RFivePlainStates}{144}
\newcommand{\RFiveBitsetStates}{576}
\newcommand{\RFiveSetA}{\ensuremath{\{\ensuremath{(-\tfrac{1}{4},-1)},\,\ensuremath{(\tfrac{1}{4},-1)}\}}}
\newcommand{\RFiveSetB}{\ensuremath{\{\ensuremath{(-\tfrac{1}{4},1)},\,\ensuremath{(\tfrac{1}{4},1)}\}}}

\newcommand{\RFiveNprims}{28}

\newcommand{\RFiveWlastA}{0.0082}
\newcommand{\RFiveWlastAB}{0.0085}
\newcommand{\RFiveCertFromA}{11}
\newcommand{\RFiveCertFromAB}{5}
\newcommand{\RFiveDiffN}{26}
\newcommand{\RFiveDiffFailMin}{1}
\newcommand{\RFiveDiffFailMax}{6}
\newcommand{\RFiveProposal}{\ensuremath{(\tfrac{1}{2},0)}}
\newcommand{\RFiveDecA}{\ensuremath{(\tfrac{1}{4},-1)}}
\newcommand{\RFiveDecB}{\ensuremath{(\tfrac{1}{4},1)}}
\newcommand{\RFiveDecAB}{\ensuremath{(\tfrac{1}{2},0)}}
\newcommand{\RFiveDecReset}{\ensuremath{(\tfrac{1}{2},0)}}
\newcommand{\RFiveSeeds}{10}
\newcommand{\RFiveMemRuns}{30}
\newcommand{\RFiveMemCompleted}{30}
\newcommand{\RFiveResetLostK}{12}
\newcommand{\RFiveResetCompA}{0}
\newcommand{\RFiveResetCompB}{0}
\newcommand{\RFiveResetCompAB}{10}

\newcommand{\RFivePKstar}{14}
\newcommand{\RFivePXstar}{\ensuremath{(\tfrac{87}{16},\tfrac{7}{4},\tfrac{1}{2})}}
\newcommand{\RFivePMargin}{\ensuremath{\tfrac{1}{16}}}
\newcommand{\RFivePA}{13}
\newcommand{\RFivePB}{16}
\newcommand{\RFivePEll}{1}
\newcommand{\RFivePCP}{2}
\newcommand{\RFivePCQ}{1}
\newcommand{\RFivePServedP}{\ensuremath{\{2\}}}
\newcommand{\RFivePServedQ}{\ensuremath{\{1,2\}}}
\newcommand{\RFivePDwellP}{11}
\newcommand{\RFivePDwellQ}{12}
\newcommand{\RFivePSetP}{\ensuremath{\{\ensuremath{(1,1)}\}}}
\newcommand{\RFivePSetQ}{\ensuremath{\{\ensuremath{(1,1)},\,\ensuremath{(\tfrac{1}{2},\tfrac{1}{2})}\}}}

\newcommand{\RFivePWlastP}{0.0510}
\newcommand{\RFivePWlastQ}{0.0509}
\newcommand{\RFivePCertFrom}{2}

\newcommand{\RFivePDiffFail}{2}
\newcommand{\RFivePProposal}{\ensuremath{(\tfrac{1}{2},\tfrac{1}{2})}}
\newcommand{\RFivePDecP}{\ensuremath{(1,1)}}

\newcommand{\RFivePMemRuns}{20}
\newcommand{\RFivePMemCompleted}{20}

\newcommand{\RSixRegions}{1\,440}
\newcommand{\RSixEdges}{30\,282}
\newcommand{\RSixH}{30}
\newcommand{\RSixSettings}{11}
\newcommand{\RSixDevSettings}{2}
\newcommand{\RSixTimeLimit}{600}
\newcommand{\RSixRSSLimit}{12}

\newcommand{\RSixAnswersEqual}{11}
\newcommand{\RSixSameReach}{all}
\newcommand{\RSixRatioMin}{6.0}
\newcommand{\RSixRatioMax}{45.9}
\newcommand{\RSixFullChecked}{4}
\newcommand{\RSixChecked}{11}
\newcommand{\RSixCompared}{1\,321\,164}
\newcommand{\RSixFullBitsetOver}{7}
\newcommand{\RSixFullPlainOver}{3}
\newcommand{\RSixBatches}{16}
\newcommand{\RSixOnlineA}{$(2,4)$}

\newcommand{\RSixOnlineSparseA}{9.84}
\newcommand{\RSixOnlineOrdA}{9.87}

\newcommand{\RSixOnlineB}{$(2,16)$}

\newcommand{\RSixOnlineSparseB}{8.35}
\newcommand{\RSixOnlineOrdB}{17.1}

\newcommand{\RSixOnlineC}{$(4,4)$}

\newcommand{\RSixOnlineSparseC}{36.4}
\newcommand{\RSixOnlineOrdC}{128}
\newcommand{\RSixOnlineOrdMiBC}{202}
\newcommand{\RSixOnlineSparseMiBC}{311}
\newcommand{\RSixOrdCoreMin}{68}
\newcommand{\RSixOrdCoreMax}{84}
\newcommand{\RSixVal}[4]{\csname rsix@#1@#2@#3@#4\endcsname}
\expandafter\def\csname rsix@sb@1@4@s\endcsname{0.62}
\expandafter\def\csname rsix@sb@1@4@mib\endcsname{146}
\expandafter\def\csname rsix@sb@1@4@extra\endcsname{19}
\expandafter\def\csname rsix@sc@1@4@s\endcsname{0.64}
\expandafter\def\csname rsix@sc@1@4@mib\endcsname{147}
\expandafter\def\csname rsix@sc@1@4@extra\endcsname{19}
\expandafter\def\csname rsix@sc@1@4@reach\endcsname{85\,104}
\expandafter\def\csname rsix@ord@1@4@s\endcsname{7.39}
\expandafter\def\csname rsix@ord@1@4@mib\endcsname{196}
\expandafter\def\csname rsix@ord@1@4@extra\endcsname{69}
\expandafter\def\csname rsix@fp@1@4@s\endcsname{5.63}
\expandafter\def\csname rsix@fp@1@4@mib\endcsname{131}
\expandafter\def\csname rsix@fp@1@4@extra\endcsname{4}
\expandafter\def\csname rsix@fp@1@4@dom\endcsname{18}
\expandafter\def\csname rsix@fb@1@4@s\endcsname{6.04}
\expandafter\def\csname rsix@fb@1@4@mib\endcsname{134}
\expandafter\def\csname rsix@fb@1@4@extra\endcsname{7}
\expandafter\def\csname rsix@fb@1@4@dom\endcsname{96}
\expandafter\def\csname rsix@sb@2@4@s\endcsname{1.44}
\expandafter\def\csname rsix@sb@2@4@mib\endcsname{160}
\expandafter\def\csname rsix@sb@2@4@extra\endcsname{30}
\expandafter\def\csname rsix@sc@2@4@s\endcsname{1.48}
\expandafter\def\csname rsix@sc@2@4@mib\endcsname{160}
\expandafter\def\csname rsix@sc@2@4@extra\endcsname{30}
\expandafter\def\csname rsix@sc@2@4@reach\endcsname{231\,187}
\expandafter\def\csname rsix@ord@2@4@s\endcsname{9.79}
\expandafter\def\csname rsix@ord@2@4@mib\endcsname{192}
\expandafter\def\csname rsix@ord@2@4@extra\endcsname{62}
\expandafter\def\csname rsix@fp@2@4@s\endcsname{7.63}
\expandafter\def\csname rsix@fp@2@4@mib\endcsname{149}
\expandafter\def\csname rsix@fp@2@4@extra\endcsname{19}
\expandafter\def\csname rsix@fp@2@4@dom\endcsname{324}
\expandafter\def\csname rsix@fb@2@4@s\endcsname{63.4}
\expandafter\def\csname rsix@fb@2@4@mib\endcsname{607}
\expandafter\def\csname rsix@fb@2@4@extra\endcsname{477}
\expandafter\def\csname rsix@fb@2@4@dom\endcsname{9\,216}
\expandafter\def\csname rsix@sb@3@4@s\endcsname{3.32}
\expandafter\def\csname rsix@sb@3@4@mib\endcsname{186}
\expandafter\def\csname rsix@sb@3@4@extra\endcsname{56}
\expandafter\def\csname rsix@sc@3@4@s\endcsname{3.59}
\expandafter\def\csname rsix@sc@3@4@mib\endcsname{189}
\expandafter\def\csname rsix@sc@3@4@extra\endcsname{59}
\expandafter\def\csname rsix@sc@3@4@reach\endcsname{508\,408}
\expandafter\def\csname rsix@ord@3@4@s\endcsname{22.4}
\expandafter\def\csname rsix@ord@3@4@mib\endcsname{185}
\expandafter\def\csname rsix@ord@3@4@extra\endcsname{55}
\expandafter\def\csname rsix@fp@3@4@s\endcsname{45.9}
\expandafter\def\csname rsix@fp@3@4@mib\endcsname{405}
\expandafter\def\csname rsix@fp@3@4@extra\endcsname{274}
\expandafter\def\csname rsix@fp@3@4@dom\endcsname{5\,832}
\expandafter\def\csname rsix@fb@3@4@s\endcsname{budget}
\expandafter\def\csname rsix@fb@3@4@mib\endcsname{budget}
\expandafter\def\csname rsix@fb@3@4@extra\endcsname{budget}
\expandafter\def\csname rsix@sb@4@4@s\endcsname{8.89}
\expandafter\def\csname rsix@sb@4@4@mib\endcsname{235}
\expandafter\def\csname rsix@sb@4@4@extra\endcsname{116}
\expandafter\def\csname rsix@sc@4@4@s\endcsname{9.65}
\expandafter\def\csname rsix@sc@4@4@mib\endcsname{235}
\expandafter\def\csname rsix@sc@4@4@extra\endcsname{116}
\expandafter\def\csname rsix@sc@4@4@reach\endcsname{1\,181\,283}
\expandafter\def\csname rsix@ord@4@4@s\endcsname{130}
\expandafter\def\csname rsix@ord@4@4@mib\endcsname{202}
\expandafter\def\csname rsix@ord@4@4@extra\endcsname{83}
\expandafter\def\csname rsix@fp@4@4@s\endcsname{budget}
\expandafter\def\csname rsix@fp@4@4@mib\endcsname{budget}
\expandafter\def\csname rsix@fp@4@4@extra\endcsname{budget}
\expandafter\def\csname rsix@fb@4@4@s\endcsname{budget}
\expandafter\def\csname rsix@fb@4@4@mib\endcsname{budget}
\expandafter\def\csname rsix@fb@4@4@extra\endcsname{budget}
\expandafter\def\csname rsix@sb@2@2@s\endcsname{1.63}
\expandafter\def\csname rsix@sb@2@2@mib\endcsname{160}
\expandafter\def\csname rsix@sb@2@2@extra\endcsname{30}
\expandafter\def\csname rsix@sc@2@2@s\endcsname{1.68}
\expandafter\def\csname rsix@sc@2@2@mib\endcsname{162}
\expandafter\def\csname rsix@sc@2@2@extra\endcsname{32}
\expandafter\def\csname rsix@sc@2@2@reach\endcsname{242\,947}
\expandafter\def\csname rsix@ord@2@2@s\endcsname{10.1}
\expandafter\def\csname rsix@ord@2@2@mib\endcsname{191}
\expandafter\def\csname rsix@ord@2@2@extra\endcsname{61}
\expandafter\def\csname rsix@fp@2@2@s\endcsname{6.98}
\expandafter\def\csname rsix@fp@2@2@mib\endcsname{139}
\expandafter\def\csname rsix@fp@2@2@extra\endcsname{9}
\expandafter\def\csname rsix@fp@2@2@dom\endcsname{144}
\expandafter\def\csname rsix@fb@2@2@s\endcsname{10.6}
\expandafter\def\csname rsix@fb@2@2@mib\endcsname{161}
\expandafter\def\csname rsix@fb@2@2@extra\endcsname{31}
\expandafter\def\csname rsix@fb@2@2@dom\endcsname{576}
\expandafter\def\csname rsix@sb@2@8@s\endcsname{1.44}
\expandafter\def\csname rsix@sb@2@8@mib\endcsname{161}
\expandafter\def\csname rsix@sb@2@8@extra\endcsname{31}
\expandafter\def\csname rsix@sc@2@8@s\endcsname{1.47}
\expandafter\def\csname rsix@sc@2@8@mib\endcsname{159}
\expandafter\def\csname rsix@sc@2@8@extra\endcsname{29}
\expandafter\def\csname rsix@sc@2@8@reach\endcsname{207\,667}
\expandafter\def\csname rsix@ord@2@8@s\endcsname{12.8}
\expandafter\def\csname rsix@ord@2@8@mib\endcsname{192}
\expandafter\def\csname rsix@ord@2@8@extra\endcsname{62}
\expandafter\def\csname rsix@fp@2@8@s\endcsname{12.0}
\expandafter\def\csname rsix@fp@2@8@mib\endcsname{177}
\expandafter\def\csname rsix@fp@2@8@extra\endcsname{47}
\expandafter\def\csname rsix@fp@2@8@dom\endcsname{900}
\expandafter\def\csname rsix@fb@2@8@s\endcsname{budget}
\expandafter\def\csname rsix@fb@2@8@mib\endcsname{budget}
\expandafter\def\csname rsix@fb@2@8@extra\endcsname{budget}
\expandafter\def\csname rsix@sb@2@16@s\endcsname{1.19}
\expandafter\def\csname rsix@sb@2@16@mib\endcsname{158}
\expandafter\def\csname rsix@sb@2@16@extra\endcsname{28}
\expandafter\def\csname rsix@sc@2@16@s\endcsname{1.15}
\expandafter\def\csname rsix@sc@2@16@mib\endcsname{158}
\expandafter\def\csname rsix@sc@2@16@extra\endcsname{28}
\expandafter\def\csname rsix@sc@2@16@reach\endcsname{160\,627}
\expandafter\def\csname rsix@ord@2@16@s\endcsname{17.1}
\expandafter\def\csname rsix@ord@2@16@mib\endcsname{195}
\expandafter\def\csname rsix@ord@2@16@extra\endcsname{65}
\expandafter\def\csname rsix@fp@2@16@s\endcsname{22.2}
\expandafter\def\csname rsix@fp@2@16@mib\endcsname{276}
\expandafter\def\csname rsix@fp@2@16@extra\endcsname{146}
\expandafter\def\csname rsix@fp@2@16@dom\endcsname{2\,916}
\expandafter\def\csname rsix@fb@2@16@s\endcsname{budget}
\expandafter\def\csname rsix@fb@2@16@mib\endcsname{budget}
\expandafter\def\csname rsix@fb@2@16@extra\endcsname{budget}
\expandafter\def\csname rsix@sb@1@2@s\endcsname{0.64}
\expandafter\def\csname rsix@sb@1@2@mib\endcsname{147}
\expandafter\def\csname rsix@sb@1@2@extra\endcsname{20}
\expandafter\def\csname rsix@sc@1@2@s\endcsname{0.67}
\expandafter\def\csname rsix@sc@1@2@mib\endcsname{147}
\expandafter\def\csname rsix@sc@1@2@extra\endcsname{20}
\expandafter\def\csname rsix@sc@1@2@reach\endcsname{87\,344}
\expandafter\def\csname rsix@ord@1@2@s\endcsname{7.55}
\expandafter\def\csname rsix@ord@1@2@mib\endcsname{196}
\expandafter\def\csname rsix@ord@1@2@extra\endcsname{69}
\expandafter\def\csname rsix@fp@1@2@s\endcsname{5.75}
\expandafter\def\csname rsix@fp@1@2@mib\endcsname{130}
\expandafter\def\csname rsix@fp@1@2@extra\endcsname{3}
\expandafter\def\csname rsix@fp@1@2@dom\endcsname{12}
\expandafter\def\csname rsix@fb@1@2@s\endcsname{5.73}
\expandafter\def\csname rsix@fb@1@2@mib\endcsname{131}
\expandafter\def\csname rsix@fb@1@2@extra\endcsname{3}
\expandafter\def\csname rsix@fb@1@2@dom\endcsname{24}
\expandafter\def\csname rsix@sb@3@8@s\endcsname{2.85}
\expandafter\def\csname rsix@sb@3@8@mib\endcsname{178}
\expandafter\def\csname rsix@sb@3@8@extra\endcsname{48}
\expandafter\def\csname rsix@sc@3@8@s\endcsname{2.98}
\expandafter\def\csname rsix@sc@3@8@mib\endcsname{178}
\expandafter\def\csname rsix@sc@3@8@extra\endcsname{48}
\expandafter\def\csname rsix@sc@3@8@reach\endcsname{423\,736}
\expandafter\def\csname rsix@ord@3@8@s\endcsname{35.1}
\expandafter\def\csname rsix@ord@3@8@mib\endcsname{188}
\expandafter\def\csname rsix@ord@3@8@extra\endcsname{58}
\expandafter\def\csname rsix@fp@3@8@s\endcsname{160}
\expandafter\def\csname rsix@fp@3@8@mib\endcsname{1393}
\expandafter\def\csname rsix@fp@3@8@extra\endcsname{1263}
\expandafter\def\csname rsix@fp@3@8@dom\endcsname{27\,000}
\expandafter\def\csname rsix@fb@3@8@s\endcsname{budget}
\expandafter\def\csname rsix@fb@3@8@mib\endcsname{budget}
\expandafter\def\csname rsix@fb@3@8@extra\endcsname{budget}
\expandafter\def\csname rsix@sb@3@16@s\endcsname{1.93}
\expandafter\def\csname rsix@sb@3@16@mib\endcsname{164}
\expandafter\def\csname rsix@sb@3@16@extra\endcsname{33}
\expandafter\def\csname rsix@sc@3@16@s\endcsname{1.93}
\expandafter\def\csname rsix@sc@3@16@mib\endcsname{164}
\expandafter\def\csname rsix@sc@3@16@extra\endcsname{34}
\expandafter\def\csname rsix@sc@3@16@reach\endcsname{255\,470}
\expandafter\def\csname rsix@ord@3@16@s\endcsname{64.5}
\expandafter\def\csname rsix@ord@3@16@mib\endcsname{192}
\expandafter\def\csname rsix@ord@3@16@extra\endcsname{62}
\expandafter\def\csname rsix@fp@3@16@s\endcsname{budget}
\expandafter\def\csname rsix@fp@3@16@mib\endcsname{budget}
\expandafter\def\csname rsix@fp@3@16@extra\endcsname{budget}
\expandafter\def\csname rsix@fb@3@16@s\endcsname{budget}
\expandafter\def\csname rsix@fb@3@16@mib\endcsname{budget}
\expandafter\def\csname rsix@fb@3@16@extra\endcsname{budget}
\expandafter\def\csname rsix@sb@4@8@s\endcsname{7.07}
\expandafter\def\csname rsix@sb@4@8@mib\endcsname{210}
\expandafter\def\csname rsix@sb@4@8@extra\endcsname{91}
\expandafter\def\csname rsix@sc@4@8@s\endcsname{7.53}
\expandafter\def\csname rsix@sc@4@8@mib\endcsname{212}
\expandafter\def\csname rsix@sc@4@8@extra\endcsname{93}
\expandafter\def\csname rsix@sc@4@8@reach\endcsname{922\,563}
\expandafter\def\csname rsix@ord@4@8@s\endcsname{346}
\expandafter\def\csname rsix@ord@4@8@mib\endcsname{202}
\expandafter\def\csname rsix@ord@4@8@extra\endcsname{83}
\expandafter\def\csname rsix@fp@4@8@s\endcsname{budget}
\expandafter\def\csname rsix@fp@4@8@mib\endcsname{budget}
\expandafter\def\csname rsix@fp@4@8@extra\endcsname{budget}
\expandafter\def\csname rsix@fb@4@8@s\endcsname{budget}
\expandafter\def\csname rsix@fb@4@8@mib\endcsname{budget}
\expandafter\def\csname rsix@fb@4@8@extra\endcsname{budget}

\newcommand{\StmtSpatial}{%
Let $d\in\Dset$. If the tightened Boolean task holds for the shifted bit
signals, $((\gamma,\beta)(\cdot-d),0)\models\varphi_\beta$, then
$(\ytil^{d,\eta},0)\models\varphi$ for every $\eta:\mathbb Z\to\Ball$.}

\newcommand{\StmtCoverage}{%
Fix Boolean signals $(\gamma,\beta)$ on $[-D,H]$.
(a) For every reach task $j$,
$\forall d\in\Dset\ \exists t\in[a_j,b_j]\ \forall r\in[0,\ell_j]:\
\beta_j(t+r-d)=1$ if and only if $\Cov_j(H)=\Dset$.
(b) For every safety task $i$,
$\forall d\in\Dset\ \forall t\in[\alpha_i,\omega_i]:\ \gamma_i(t-d)=1$ if and
only if $\gamma_i(t)=1$ for all $t\in[\alpha_i-D,\omega_i]$.}

\newcommand{\StmtCompile}{%
Let $\xi$ be a Boolean signal on $[-D,H]$ and $\psi$ a Boolean STL formula,
possibly with negation and bounded until $\UntilOp_I$, whose evaluation at time
$0$ reads $\xi$ only on $[0,H]$. Then
\[
\big[\forall d\in\Dset:\ (\xi(\cdot-d),0)\models\psi\big]
\iff (\xi,-D)\models\G_{[0,D]}\psi .
\]
In particular, the tightened Boolean task holds under every offset iff
$((\gamma,\beta),-D)\models\G_{[0,D]}\varphi_\beta$.}

\newcommand{\StmtExpiry}{%
Fix a reach task and a Boolean signal. If offset $d$ is expired at $k$, then
$d\notin\Cov(k')$ for every $k'\ge k$; hence the task is failed at every
$k'\ge k$ and $\Cov(H)\neq\Dset$. Moreover, for $k\ge b+\ell$ the task is not
failed at $k$ if and only if $\Cov(k)=\Dset$.}

\newcommand{\StmtSuffix}{%
Fix a reach task, a Boolean signal and $k\ge-D-1$. If the task is not failed
at $k$, then $\Cov(k)=\{c,c+1,\dots,D\}$ for some $c\in[0,D+1]$, where
$c=D+1$ denotes $\Cov(k)=\varnothing$.}

\newcommand{\StmtCompact}{%
Fix a Boolean signal $\beta$ on $[-D,H]$, let $q_{-D-1}=\qinit$ and
$q_k=\Gamma_k(q_{k-1},\beta(k))$. For every $k\in[-D-1,H]$: $q_k=\bot$ iff
some reach task is failed at $k$; otherwise $q_k=((c_j,r_j))_j\in\Qset_k$
with $\Cov_j(k)=\{c_j,\dots,D\}$ and $r_j=r_j(k)$ for every $j$. The monitor
therefore takes at most $1+\prod_{j}(D+2)(\ell_j+2)$ values.}

\newcommand{\StmtMono}{%
For every $k$, all $q\preceq q'$ in $\Qset\cup\{\bot\}$ and all bit vectors
$\beta\le\beta'$ (componentwise),
$\Gamma_k(q,\beta)\preceq\Gamma_k(q',\beta')$. Moreover, $\Qset_k$ and the
acceptance set $\{q:\Acc(q)\}$ are upward closed in $\Qset\cup\{\bot\}$.}

\newcommand{\StmtSegment}{%
Fix one reach task and a true segment $[p,q]\subseteq[-D,b+\ell]$ of $\beta$
with $q-p\ge\ell$. The witnesses $s\in[p,q-\ell]$ cover $\Dset$ iff
$p\le b-D$ and $q\ge a+\ell$. Consequently, $\beta\in\Fsi$ iff
$\beta$ satisfies
\[
\psi_{\mathrm{single}}=
\begin{cases}
\F_{[a,b-D]}\G_{[0,\ell]}\beta\ \text{at time } 0, & \Delta\ge D,\\
\G_{[b,a+\ell+D]}\beta\ \text{at time } {-D}, & \Delta<D,
\end{cases}
\]
where the second case states $\beta(t)=1$ for all $t\in[b-D,a+\ell]$ in
absolute nominal time; negative times read the recorded history.}

\newcommand{\StmtHierarchy}{%
Fix one reach task with parameters $(a,b,\ell,D)$ and $\Delta=b-a$.
If $\Delta\ge D$, then $\Fsh\subseteq\Fsk=\Fsi\subseteq\Fex$.
If $\Delta<D$, then $\Fsk=\varnothing$ and $\Fsh\subseteq\Fsi\subseteq\Fex$.}

\newcommand{\StmtStrict}{%
Fix one reach task and consider all Boolean signals on $[-D,b+\ell]$. Then
$\Fsi\subsetneq\Fex$ if and only if $D\ge1$ and $\Delta\ge\ell+1$. If $D=0$
or $\Delta\le\ell$, every signal in $\Fex$ has a single true segment whose
witnesses cover $\Dset$.}

\newcommand{\StmtExactMono}{%
If $q\preceq q'$, then $\Tset_k(q)\subseteq\Tset_k(q')$ for every
$k\in[0,H]$.}

\newcommand{\StmtMinPre}{%
Fix $k\in[-D,H-1]$, a label $\sigma\in\{0,1\}^{n_r}$ and a target
$g=((c^*_j,r^*_j))_j\in\Qset_{k+1}$, and let
$\Pre_k(g,\sigma)=\{q\in\Qset_k:\Gamma_{k+1}(q,\sigma)\succeq g\}$.
With $s_j=k+1-\ell_j$, $\lo_j=\max\{0,a_j-s_j\}$ and
$\hi_j=\min\{D,b_j-s_j\}$, let $X_j=\{(c^*_j,0)\}$ if $\sigma_j=0$ and
$r^*_j=0$, $X_j=\varnothing$ if $\sigma_j=0$ and $r^*_j>0$, and, if
$\sigma_j=1$,
$X_j=\{(c^*_j,\max\{0,r^*_j-1\})\}\cup
\{(\min\{D+1,\hi_j+1\},\ell_j):\lo_j\le\hi_j,\ \lo_j\le c^*_j\}$.
Then $\Pre_k(g,\sigma)$ is upward closed in $\Qset$,
$X_1\times\dots\times X_{n_r}\subseteq\Pre_k(g,\sigma)$, and
\[
\min\Pre_k(g,\sigma)=\min\,(X_1\times\dots\times X_{n_r}).
\]
Thus, for a fixed time, label and target generator, each task contributes
at most two candidates, all candidates are valid predecessors, and there are
at most $2^{n_r}$ of them.}

\newcommand{\StmtDPEq}{%
For every $k\in[0,H]$, $\rho\in\mathcal R$ and $q\in\Qset\cup\{\bot\}$,
$\Win_k(\rho,q)=1$ iff $m\preceq q$ for some $m\in M_k(\rho)$. Hence the plain
recursion \eqref{eq:win-H}--\eqref{eq:win-k} and the ordered recursion
\eqref{eq:ordered} certify identical sets on the same library.}

\newcommand{\StmtSound}{%
Let Assumption~\ref{ass:convex} hold, let $\Lib$ be verified, and let
$\That_k(q)=\bigcup\{P_\rho:\Win_k(\rho,q)=1\}$. Then, for all $k\in[0,H]$:
(a) $q\preceq q'$ implies $\That_k(q)\subseteq\That_k(q')$;
(b) if $k<H$, $z\in P_\rho$ and $\Win_k(\rho,q)=1$, then $\Adm_k(z,q)$ holds
and some edge $e$ with $\rho_e=\rho$ satisfies $v=F_ez+g_e\in\mathcal V$,
$z^+=Az+Bv\in P_{\rho'_e}$, $\sigma_e\le\chi(Cz^+)$, and
$z^+\in\That_{k+1}(\Gamma_{k+1}(q,\beta))$ for every $\beta\ge\sigma_e$;
(c) $\That_H(q)\neq\varnothing$ implies $\Acc(q)$, and
$\That_H(q)\subseteq\Zpost$;
(d) $\That_k(q)\subseteq\Tset_k(q)$.}

\newcommand{\StmtRobPre}{%
Fix $k\in[-D,H-1]$, an edge with branches $(\rho_b,\sigma^b)$, $b=1,\dots,B$,
$B\ge1$, and targets $g^b=((c^b_j,r^b_j))_j\in\Qset_{k+1}$, and let
$\PreA_k=\{q\in\Qset_k:\Gamma_{k+1}(q,\sigma^b)\succeq g^b\ \forall b\}$.
For task $j$ let $I^0_j=\{b:\sigma^b_j=0\}$ and $I^1_j=\{b:\sigma^b_j=1\}$. If
$r^b_j>0$ for some $b\in I^0_j$, let $X_j=\varnothing$. Otherwise let
$C^0_j=\min\{\cbar_{j,k},\min_{b\in I^0_j}c^b_j\}$ (the inner minimum is
$+\infty$ if $I^0_j=\varnothing$); let $X_j=\{(C^0_j,0)\}$ if
$I^1_j=\varnothing$, and otherwise, with $C^1_j=\min_{b\in I^1_j}c^b_j$,
$R^1_j=\max_{b\in I^1_j}r^b_j$ and $s_j,\lo_j,\hi_j$ as in
Lemma~\ref{lem:minpre},
$X_j=\{(\min\{C^0_j,C^1_j\},\max\{0,R^1_j-1\})\}\cup
\{(\min\{C^0_j,D+1,\hi_j+1\},\ell_j):\lo_j\le\hi_j,\ \lo_j\le C^1_j\}$.
Then $\PreA_k$ is upward closed in $\Qset$,
$X_1\times\dots\times X_{n_r}\subseteq\PreA_k$, and
$\min\PreA_k=\min\,(X_1\times\dots\times X_{n_r})$. In particular, for every
$B$ and every fixed target vector there are at most $2^{n_r}$ candidates.}

\newcommand{\StmtSuffixSim}{%
Let $\iota(q)=((\{c_j,\dots,D\},r_j))_j$ map compact states to
explicit (served set, run length) states, and let $\Gamma^{\mathrm{bit}}_k$
be the explicit update that adds the offsets served by a witness completing
at $k$ and returns failure if some offset is expired at $k$. For every $k$,
$q\in\Qset$ and bit vector $\beta$: $\Gamma^{\mathrm{bit}}_k(\iota(q),\beta)$
is failure iff $\Gamma_k(q,\beta)=\bot$, and otherwise equals
$\iota(\Gamma_k(q,\beta))$. Moreover $\iota(q)$ is not failed at $k$ iff
$q\in\Qset_k$.}

\newcommand{\StmtRobCert}{%
Let $\Blib$ be a branching library, $\Kpost$ a set of regions and $\Win$ the
recursion \eqref{eq:rwin-H}--\eqref{eq:rwin-k}.
(a) For every $k\in[0,H]$ and $\rho$, $\Win_k(\rho,\cdot)$ is upward closed in
$\Qset\cup\{\bot\}$, and $\Win_k(\rho,q)=1$ iff $m\preceq q$ for some
$m\in M_k(\rho)$, where $M_H(\rho)=\{((0,0))_j\}$ if $\RA_H(\rho)$ and
$\rho\in\Kpost$ ($\varnothing$ otherwise) and
$M_k(\rho)=\min\bigcup\{\min\PreA_k((g^b)_b):e\in A(\rho),\
g^b\in M_{k+1}(\rho_b)\}$ if $\RA_k(\rho)$ ($\varnothing$ otherwise).
(b) The explicit recursion over (served set, run length) states, with
$\Gamma^{\mathrm{bit}}$ and the same library, satisfies
$\Win^{\mathrm{bit}}_k(\rho,\iota(q))=\Win_k(\rho,q)$ for all $k$, $\rho$ and
$q\in\Qset$.
(c) If every edge has a single branch, \eqref{eq:rwin-k} is
\eqref{eq:win-k} and $M_k$ is the set of Theorem~\ref{thm:dp-eq}.}

\newcommand{\StmtRobLoop}{%
Let $\Blib$ be verified (Definition~\ref{def:blib}), let $\Kpost$ satisfy
\eqref{eq:kpost} with witness edges $e_{\rm post}(\rho)$, and let $x_0\in
P_{\rho_0}$, with $q_0$ obtained by running $\Gamma$ on the actual bits of the
recorded history and of $x_0$, satisfy $\Win_0(\rho_0,q_0)=1$; let the actual
pre-mission states $x_t$, $t\in[-D,-1]$, satisfy every tightened safety
predicate active at $t$. Let the closed
loop at each $k<H$ apply $u_k=\kappa_e(x_k)$ for an edge $e\in A(\rho_k)$ with
$\Win_{k+1}(\rho_b,\Gamma_{k+1}(q_k,\sigma^b))=1$ for every branch $b$, then
select any branch $b$ with $x_{k+1}\in P_{\rho_b}$ and
$\sigma^b\le\chi(h(x_{k+1}))$, and set $\rho_{k+1}=\rho_b$,
$q_{k+1}=\Gamma_{k+1}(q_k,\chi(h(x_{k+1})))$; for $k\ge H$ let it apply
$\kappa_{e_{\rm post}(\rho_k)}$ and select branches in the same way. Then, for
every admissible disturbance sequence, such an edge and such a branch exist at
every step, $\Win_k(\rho_k,q_k)=1$ for $k\in[0,H]$, the state and input
constraints hold, $\Acc(q_H)$, and $\rho_k\in\Kpost$ for every $k\ge H$.
Consequently the tightened safety tasks hold under every offset $d\in\Dset$,
and every reach task $j$ is completed under offset $d$ by nominal time
$b_j-d+\ell_j$.}

\newcommand{\StmtRobMono}{%
Let $\Blib\subseteq\Blib'$ be branching libraries on the same regions, with
the same $\RA$ and $\Kpost$. Then $\Win^{\Blib}_k\le\Win^{\Blib'}_k$
pointwise for every $k$. In particular, adding verified multi-successor edges
to a single-successor library never shrinks the certified set; the
enlargement is relative to this fixed abstraction.}

\newcommand{\StmtPostFP}{%
Let $\Kpost$ be the greatest fixed point of
$\mathcal K\mapsto\{\rho\in\mathcal K:\exists e\in A(\rho)\ \forall
b:\ \rho_b\in\mathcal K\}$ below $\mathcal K^0$, computed by the decreasing
iteration from $\mathcal K^0$. For a verified $\Blib$, the switching law that
applies a witness edge $e_{\rm post}(\rho)$ in the current region keeps
$x_k\in\bigcup_{\rho\in\Kpost}P_\rho$ for all $k$ and every admissible
disturbance sequence.}

\newcommand{\StmtBoxObstruction}{%
For the unicycle \eqref{eq:unicycle} with $\mathcal W\supseteq
[-\bar w_x,\bar w_x]\times[-\bar w_y,\bar w_y]\times\{0\}$, $\bar w_x,\bar
w_y>0$, no feedback $\kappa:X\to U$ renders a set $X=B\times\Theta$, with $B$ a
closed box in position and $\Theta$ nonempty, robustly invariant.}

\newcommand{\StmtRF}{%
Let Assumptions~\ref{ass:tube}, \ref{ass:post}, \ref{ass:convex} and
\ref{ass:onesided} hold, let $\Lib$ be verified, and let $N\ge1$. Let
$k\in[0,H-2]$ and let
$(v^*,\hat\beta^*)$ be feasible for $\mathcal P_k(z_k,q_k)$, with predicted
states $z^*_{i|k}$ and monitor states $q^*_{i|k}$. Suppose that
(i) the nominal center is not reset, $z_{k+1}=z^*_{1|k}$, and
(ii) the executed monitor processes the executed nominal output with exact
tightened bits, $q_{k+1}=\Gamma_{k+1}(q_k,\chi(Cz_{k+1}))$.
Then $q^*_{1|k}\preceq q_{k+1}$ and $\mathcal P_{k+1}(z_{k+1},q_{k+1})$ is
feasible. A feasible solution reuses $v^*_{i|k}$ and $\hat\beta^*_{i|k}$ for
the overlapping steps and, if $k+N_k<H$, appends the input
$F_ez^*_{N_k|k}+g_e$ and the label $\sigma_e$ of a library edge given by
Theorem~\ref{thm:sound}(b).}

\newcommand{\StmtClosedLoop}{%
Let the assumptions of Theorem~\ref{thm:rf} and
Assumption~\ref{ass:history} hold, and let $H\ge1$. Assume that an initial
feasible candidate satisfying Assumption~\ref{ass:onesided} has been obtained
for $\mathcal P_0(z_0,q_0)$, with the $z_0$, $q_0$ and recorded history used at
task time $0$, and let the closed loop follow Algorithm~\ref{alg:closedloop}.
Then, for every disturbance sequence in $\mathcal W$:
(i) for every $k\in[0,H-1]$, the candidate selected at time $k$ is feasible for
$\mathcal P_k(z_k,q_k)$;
(ii) $x_k\in\mathcal X$ for $k\in[0,H]$ and $u_k\in\mathcal U$ for
$k\in[0,H-1]$;
(iii) $\Acc(q_H)$, i.e., $q_H\neq\bot$ and $c_j(q_H)=0$ for all $j$, and
$z_H\in\Zpost$;
(iv) $(\ytil^{d,\eta},0)\models\varphi$ for all $d\in\Dset$ and all
$\eta:\mathbb Z\to\Ball$. In particular, the sampled closed-loop output
$y_t=Cx_t$, $t\in[0,H]$, satisfies $\varphi$.}

\draftfalse
\showproofsketchestrue

\begin{document}

\title{Feedback Completion Certificates for Repeated Dwell Tasks with Time Windows}

\ifdefined\RSSAnonymousReview
  \author{Author Names Omitted for Anonymous Review}
\else
  \newcommand*{\RSSaff}[1]{\textsuperscript{#1}}
\pdfstringdefDisableCommands{\def\RSSaff#1{}}
\newcommand*{\RSSauthors}{Jiaxun Sun\RSSaff{1,*}, Yuqi Yang\RSSaff{2},
Senlan Yao\RSSaff{1}, Hengyu Xue\RSSaff{3}, Hongming Fu\RSSaff{4}
and Zhimu Zhou\RSSaff{4}}
\hypersetup{pdfauthor={\RSSauthors}}
\author{\IEEEauthorblockN{\RSSauthors}
\IEEEauthorblockA{\RSSaff{1}Department of Mechanical and Process Engineering, ETH Zurich, Switzerland\\
\RSSaff{2}Quicrobot\\
\RSSaff{3}School of Engineering, University of Birmingham, United Kingdom\\
\RSSaff{4}School of Artificial Intelligence, Shanghai Jiao Tong University, China\\
\RSSaff{*}Corresponding author.\\
jiaxsun@ethz.ch, saint.yangyuqi@gmail.com, senyao@ethz.ch,\\
hxx634@bham.ac.uk, fuhongming@sjtu.edu.cn, piekeniuszwu@gmail.com}}

\fi

\maketitle

\begin{abstract}
An inspection robot may owe several measurements at the same station, each
with a different window in which a dwell can begin. We ask what a feedback
controller must retain about earlier visits to guarantee all remaining
deadlines under bounded disturbances. For a common spatial predicate and
dwell length, removing redundant windows orders the remaining obligations
by both endpoints. On every history without an expired obligation, the
served windows form a suffix. One boundary and one capped run length
therefore preserve the future satisfaction of the original task. On a
verified library of branching motion primitives we use this state to
recover a feedback policy. For one task type a scalar threshold per run
length suffices, with work linear in the number of listed branches per
time layer for fixed dwell. Multiple types require a joint recursion.
On a quadrotor inspection library the threshold sweep reduced the cost
of constructing a reusable full-domain policy, while a sparse Boolean
baseline answered a single initial query faster. In a two-station MuJoCo patrol, three task-aware filters all met every
window in twelve paired runs each. Of eight initially certified runs, the
completion filter retained the certificate in all eight, compared with
three for safety-only filtering.

\end{abstract}

\IEEEpeerreviewmaketitle

\section{Introduction}\label{sec:window-intro}
An inspection robot checks two stations several times during a shift.
The first reading at one station might be allowed in a narrow start window
and the next reading in a wider one. A reading takes several consecutive
samples. When the robot leaves a station to do other work, its position and
the clock do not tell the controller which visits it still owes. Even a
flag recording that the station has been visited loses the distinction
between an early and a late reading. The controller needs enough of the
executed history to decide whether it can still meet each deadline after
the next disturbance.
Figure~\ref{fig:s1-mechanisms}(a) shows what that distinction costs:
at the same state and clock, returning from A or from B leaves disjoint
sets of certified actions. Figure~\ref{fig:s1-mpc-return} shows how a
short-horizon controller can certify a return that lies beyond its plan.

We study finite lists of such obligations known before execution. A
measurement may count for every window containing its start. This convention
fits an inspection record that can satisfy overlapping requirements, or a
single request whose actual service window is one of several candidates.
It also gives the usual constant-clock-offset interpretation when the
candidate windows are equal-width translations. Separate work orders that
each consume a measurement require a different assignment rule.

Our first result describes exactly what to remember. Removing a window that
contains another preserves the task. Among the retained windows the served
ones are consecutive in endpoint order on every history that can still meet
its deadlines. A boundary index and the current dwell length therefore
determine the response to every continuation. Second, we use this order to
solve a feedback game on verified motion primitives. With one task type the
winning boundary for each run length satisfies a scalar backward recursion.
Several types share the same physical game but may compete for the next
action and require a joint state. Third, we compare executable policies on
identical queries and test the role of memory and branching in closed-loop
robot simulation. We report the cost of constructing a certificate alongside
the cost of querying it.

\begin{figure}[t]
\centering
\includegraphics[width=\columnwidth]{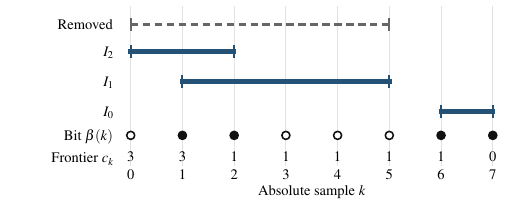}
\caption{Intervals constrain visit starts. The dashed interval contains two
retained windows. Filled circles mark true samples. With a two-sample dwell,
the visit ending at sample 2 serves $I_2$ and $I_1$. The visit ending at 7
serves $I_0$. The bottom trace records the first served index.}
\label{fig:window-history}
\end{figure}
\section{Window Histories and Feedback Completion}\label{sec:window-theory}

\subsection{What a completed visit establishes}
A task type $g$ specifies one Boolean signal $\beta_g$ and a dwell
length $\ell_g\ge0$. Its known integer intervals $I_{gi}=[a_{gi},b_{gi}]$
restrict the \emph{start} of a visit. Together with sampled safety
requirements $S_t$, the objective is
\begin{equation}\label{eq:win-objective}
  x_t\in S_t\ (t\text{ active}),\quad
  \forall g,i\ \exists s\in I_{gi}:\ 
     \bigwedge_{j=0}^{\ell_g}[\beta_g(s+j)=1].
\end{equation}
The last admissible completion sample is $b_{gi}+\ell_g$. Several
obligations may use the same witness $s$. A fixed clock offset
$d\in[0,D]$ gives the equal-width family $[a-d,b-d]$. Finite service
windows can also have different widths and gaps. All candidate windows
are enforced if they represent uncertainty unobserved by the policy.
Sharing requires identical output map, predicate (including tightening
and boundary convention), bit source and dwell length. Absolute indices
may be negative. Equivalently one can shift the origin of the corresponding
future-time STL formula to the earliest sample.

For one type, discard duplicates and intervals strictly containing another
interval. Let the remaining $M$ intervals have decreasing left endpoints,
indexed from zero. Their right endpoints then decrease strictly as well.
An empty family contributes no obligation. A listed interval with $a>b$
makes the task infeasible. At time $k$ a window is served if a witness in it
has finished by $k$. An unserved window has expired when $k\ge b_i+\ell$.
Write $J(s)=\{i:s\in I_i\}$ and
$e_k=\min(\{i:b_i+\ell\le k\}\cup\{M\})$.

\begin{theorem}[Shared history]\label{thm:window-history}
Removal preserves expiry failure at every sample and satisfaction of
the original conjunction. Until failure, the served retained windows form
a suffix $\{c,\ldots,M-1\}$. The pair $(c,r)$, with $r$ the trailing run
of ones capped at $\ell+1$, determines acceptance of every future bit
continuation at the same time. For $G$ types the full representation has at
most $1+\prod_g(M_g+1)(\ell_g+2)$ states, including absorbing failure.
\end{theorem}
\begin{IEEEproof}[Proof sketch]
A witness in a retained subinterval serves every raw interval containing
it, and its deadline is no later. If a served interval $i$ has a witness
starting at $s$ but $j>i$ remains unserved, then $a_j<a_i\le s$ forces
$s>b_j$. By the end of that dwell the unserved interval $j$ has already
expired. Thus a live served set is a suffix. Its boundary records every
completed obligation; the final $r$ true samples determine which unfinished
dwells can extend across the current time. Reading the next bit updates
both quantities deterministically, giving the same answer for every
continuation of any two prefixes with equal state.
\end{IEEEproof}

After sample $k$, the formal valid domain is $Q_k=\{0\le c\le e_k,\
0\le r\le\ell+1\}$ in integers. To process sample $k+1$, let
$r'=\min(r+1,\ell+1)$ for bit one and $r'=0$ for bit zero. If
$r'=\ell+1$ and $J(k+1-\ell)$ is nonempty, set
$c'=\min(c,\min J(k+1-\ell))$; otherwise set $c'=c$.
Return failure if $c'>e_{k+1}$. This order admits a visit ending on
its deadline. Each further one after saturation yields another witness.
The joint update $\Gamma_{k+1}$ acts coordinatewise; any failure is
absorbing. At $L-1$ initialize $(M_g,0)$ for every type, for
$L\le0$ covering all retained releases and required past safety checks.
Read the recorded samples $L,\ldots,-1$ and the initial sample $0$.
Take $H\ge0$ covering the original deadlines and safety horizon.

\subsection{Feedback game on a verified library}
Let $x^+\in F(x,u)$. At $k$, $\RA_k(\rho)$ certifies the state and
active safety constraints throughout region $P_\rho$. Each edge $e$ from
$\rho$ supplies an admissible control law and a nonempty branch list
$(\rho_b,\sigma_b)$, where $\sigma_b$ is a lower label. For every allowed
successor under that law, verification must establish at least one branch
with both $x^+\in P_{\rho_b}$ and $\sigma_b\le\beta(x^+)$.
The post-mission set $\Kpost$ admits a verified invariant controller.
For $x^+=f(x,u,w)$, take $F(x,u)=\{f(x,u,w):w\in\mathcal W(x,u)\}$.
We seek one causal policy that satisfies~\eqref{eq:win-objective} for
every allowed disturbance sequence.
Let $q=((c_g,r_g))_{g=1}^G$ be a live joint state. At $H$,
$\Win_H(\rho,q)=1$ exactly when $\RA_H(\rho)$, $\rho\in\Kpost$
and $c_g=0$ for every $g$. Failure or an invalid source loses at
every time. For $k<H$, put $q_b^+=\Gamma_{k+1}(q,\sigma_b)$. Then
\begin{equation}\label{eq:s1-joint-game}
 \Win_k(\rho,q)=\RA_k(\rho)\ \land
 \bigvee_{e\in A(\rho)}\;
 \bigwedge_{b\in\Succ(e)}\!\Win_{k+1}(\rho_b,q_b^+).
\end{equation}
One edge must cover every successor before the disturbance is known.
Each observed branch can then choose its own next edge. The two-station
controller solves this joint game; for one task type
Theorem~\ref{thm:window-threshold} reduces it to a scalar recursion.

For a single type write $T_{k,\rho}(r)$ for the largest winning $c$ at
run $r$, or $-1$ for none. At $H$ store $0$ in admissible post regions
and $-1$ elsewhere. For a branch, compute the next run $r'_b$ from its
label and put $t_b=T_{k+1,\rho_b}(r'_b)$, $J=J(k+1-\ell)$. Then
\begin{align}
 p_b&=\begin{cases}
 -1,&t_b=-1,\\
 e_k,&r'_b=\ell+1,\ J\ne\varnothing,\ \min J\le t_b,\\
 \min(e_k,t_b),&\text{otherwise},
 \end{cases}\label{eq:window-branch}\\
 T_{k,\rho}(r)&=\max_{e\in A(\rho)}\min_{b\in\Succ(e)} p_b .
 \label{eq:window-backward}
\end{align}
An inadmissible source or empty action set gives $-1$.

\begin{theorem}[Single-type threshold]\label{thm:window-threshold}
For every formal state in $Q_k$, the explicit library game wins exactly
when $c\le T_{k,\rho}(r)$. An edge attaining a nonnegative maximum in
\eqref{eq:window-backward} certifies all such $c$. With window indices
and admissibility precomputed, the recursion takes
$O((\ell+2)((H+1)|\mathcal R|+H\sum_e|\Succ(e)|))$ arithmetic
operations and $O((H+1)|\mathcal R|(\ell+2))$ threshold and policy entries.
\end{theorem}
\begin{IEEEproof}[Proof sketch]
For a nonempty $J(s)=[u,v]$ at $s=k+1-\ell$, integer endpoints give
$v+1=e_k$: both count the intervals with $b_i+\ell>k$. All valid sources
satisfy $c\le e_k$, so an eligible new visit moves the boundary to
$\min(c,u)$. If $u\le t_b$, every source boundary passes; otherwise only
$c\le t_b$ passes. Intersecting these prefixes across branches takes
their minimum, and taking their union across actions takes their maximum.
Induction from the terminal layer proves the claim.
\end{IEEEproof}

For several types the winning set can have incomparable frontiers.
For example, take two zero-dwell types with one start window $[0,1]$
each. After sample zero one action can serve A at sample one and a
second can serve B. A prefix with A already served wins, as does one
with B served. A prefix with neither served loses. Both its coordinates
occur separately in winning states. The joint relation is therefore
essential even with a single sample left.

A sparse implementation searches the joint game from a query. An
ordered implementation stores the minimal winning pairs in the order
$(c,r)\preceq(d,t)$ iff $c\ge d$ and $r\le t$, componentwise across
types. For illustration, fix a one-bit branch whose target is $(d,t)$.
On a valid source, domination of this target requires one of
$(d,\max(0,t-1))$ or $(e_k,\ell)$ as a lower bound in completion
order. The second candidate is available only when $J\ne\varnothing$
and $\min J\le d$. The first candidate uses coverage
already present. The second finishes a new dwell. A zero-bit branch
requires $t=0$ and has generator $(d,0)$. Grouping branches with equal
bits intersects their requested frontiers and run bounds. After this
intersection at most two candidates remain per type for each fixed tuple
of branch targets. Choosing a target for every branch can still require
many tuples.

\begin{theorem}[Feedback completion]\label{thm:window-feedback}
Given verified edges, a safe prehistory and a winning actual initial
region and history, choosing a certified edge and then a branch matching
the observed region and lower label preserves a certificate at every step.
Every admissible disturbance sequence satisfies \eqref{eq:win-objective}
at its original deadlines and reaches a post region at $H$.
\end{theorem}

\begin{IEEEproof}[Proof sketch]
Terminal acceptance is upward closed. Increasing a source state or a bit
cannot shorten the next run or undo an earned visit, so backward induction
makes each winning set upward closed. Every actual successor is covered by
a verified branch. Updating with its measured bits therefore dominates the
winning lower-label update on that branch. Induction to $H$ proves safety
and completion; the interval reduction returns the guarantee to each raw
deadline.
\end{IEEEproof}

For observed-state feedback the update uses the observed bits. A tube
MPC instance instead uses the exact bits of its executed nominal centre
and a tightening that transfers these bits to the physical output.
Exactness of the game compares controllers on the declared labelled
library; its physical guarantee concerns the verified sampled model.

\section{Using the Certificate}\label{sec:window-implementation}
The filter accepts a proposed edge when it passes every branch test
in~\eqref{eq:s1-joint-game}. A failed or late proposal triggers the stored
winning edge for the current tick, region and history. For one type,
$c\le\min_b p_b$ recovers every permitted edge.
After execution, a measured successor in $P_{\rho_b}$ with actual bits
at least $\sigma_b$ determines a valid continuation.
\ifdefined\WindowStageOne\else
This deadline rule requires a winning starting pair and a backup available
before the input deadline. For one type any maximizer in~\eqref{eq:window-backward}
supplies that backup. For multiple types each edge must pass the joint
successor test, since two separate station tests may select different edges.
\fi

\begin{figure}[tb]
\centering
\includegraphics[width=\columnwidth]{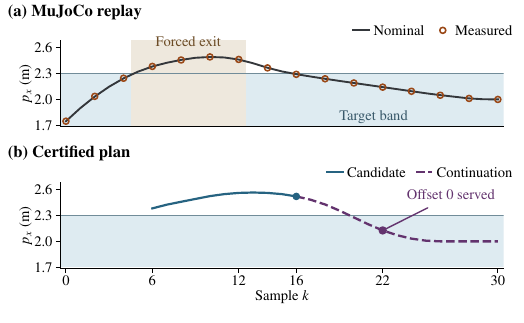}
\caption{A three-sample dwell must start in each window $[3-d,20-d]$,
$d\in\{0,\ldots,8\}$. (a) Measured and nominal quadrotor positions in
MuJoCo. Every admissible nominal input leaves the centre outside the
tightened target at samples \EOneForcedOut{}. (b) The plan adopted at
$k=\EOneTailK$ ends at
$t=\EOneTailTau$ with $c=\EOneTailCTau$: offset zero is still owed.
The model-certified continuation serves it at its deadline, $k=\EOneTailClear$.}
\label{fig:s1-mpc-return}
\end{figure}

\emph{Short-horizon MPC.} Raman et al.~\cite{raman2014mpc} encode STL
constraints in mixed-integer MPC. Sadraddini and Belta~\cite{sadraddini2015robust}
treat bounded additive disturbances. Time-interval decomposition links
STL subtasks on disjoint horizons~\cite{yu2022decomp}. Backward-reachable
terminal sets also support infinitely recurring tasks~\cite{vlahakis2024recurring}.
Here the terminal set depends on which finite windows the candidate has
already served. For a linear tube, propagate the nominal centre
$z^+=Az+Bv$ across ticks without resetting it to the measurement, and
apply $u=v+K(x-z)$~\cite{mayne2005tube}. For additive
disturbances in $\mathcal W$, let $(A+BK)\mathcal E\oplus\mathcal W
\subseteq\mathcal E$. Tighten state and input sets by $\mathcal E$ and
$K\mathcal E$, respectively, and output predicates by $C\mathcal E$
for $y=Cx$. Compute
$\Win^z$ on the nominal library with tightened predicates and constraints.
At tick $k$ plan to $t=\min(k+N,H)$, $N\ge1$. Enforce tightened input
and sampled state/safety constraints. Update the executed history $q_k$
along the plan with lower bounds on its tightened bits and impose
\begin{equation}\label{eq:s1-window-terminal}
 z_{t\mid k}\in\widehat{\mathcal T}_t(q_{t\mid k}),\qquad
 \widehat{\mathcal T}_t(q)=\!\bigcup_{\rho:\Win_t^z(\rho,q)=1}\!P_\rho .
\end{equation}
\begin{corollary}[Short-horizon completion]\label{cor:s1-window-mpc}
Assume valid tightenings for an invariant tube containing the initial error,
a verified initial plan for the actual centre and safe history, and a verified
one-step extension from each winning nominal pair before $H$.
Executing verified plans under~\eqref{eq:s1-window-terminal} preserves
feasibility and meets every original deadline.
\end{corollary}
\begin{IEEEproof}
The executed centre's true tightened bits dominate the planned bits.
Shift the remaining inputs at their original times and recompute history.
Monotonicity preserves feasibility and terminal membership. Append one
certified step if $t<H$. A plan ending at $H$ simply shortens.
Store this plan as backup. At $H$ all windows are served, and the tube
tightening transfers sampled safety and the dwell witnesses to the plant.
\end{IEEEproof}
In Fig.~\ref{fig:s1-mpc-return}, requiring all visits to finish inside
the ten-step plan would discard this feasible candidate. At its endpoint,
condition~\eqref{eq:s1-window-terminal} certifies the return that is still due.

\section{Related Work}\label{sec:window-related}
A dwell with allowed starts $[a,b]$ is expressible in
STL~\cite{maler2004monitoring} or as a hold within $[a,b+\ell]$ in
Time Window Temporal Logic~\cite{vasile2017twtl}. Temporal robustness
measures tolerance to common or predicate-specific shifts~\cite{rodionova2023tecs}
and to shifts in both directions~\cite{rodionova2023leftright}.
Synthesis can exploit redundancy across time shifts~\cite{yu2023atr}.
Spatiotemporal monitors quantify tradeoffs between spatial and temporal
perturbations~\cite{schoen2026str}. Here even a deterministic schedule
can give visits unequal windows. Uncertain travel durations pose a
different planning problem~\cite{linard2024mitl}.

Partial formula evaluation and robustness-to-go carry prefix information
into STL control~\cite{ilyes2025rcpe,ilyes2026rotogo}. Feasibility can
also be indexed by a subset of remaining obligations~\cite{yu2024monitor}.
Theorem~\ref{thm:window-history} shows that before failure the served
windows of each type form a suffix, represented by one frontier.
\ifdefined\WindowStageOne\else
For example, a predicate with zero dwell reduces to repeated eventualities
over that same predicate. The frontier specifies which remain to be served
when the controller chooses an edge with several possible successors.
\fi

STL runtime assurance and value functions filter proposed
actions~\cite{baird2023rta,so2026value}. Reachability gives completion
guarantees under disturbances~\cite{jiang2024completion}. Feedback
refinement relates symbolic policies to physical systems~\cite{reissig2017frr},
and antichain algorithms exploit ordered states~\cite{dewulf2006antichains,cadilhac2022acacia}.
For one dwell type, our order yields a scalar winning threshold and a
feedback action for every frontier it covers.

\section{Experiments}\label{sec:window-experiments}
\subsection{Correctness against independent oracles}
We compared the monitor to a direct enumeration of dwell starts in each
original window, including duplicated and nested intervals, zero dwell,
negative indices and completion on the deadline. The outputs agreed on
33,356 prefixes. Exhaustive future extensions gave 10,800 agreements on
the remaining language. Explicit source-state enumeration also agreed
on the complete set of minimal predecessors for 3,508 fixed-target
queries. On 16 seeded branching graphs, explicit coverage and the policy
implementations agreed on 5,856 legal states and 16,704 state--action
checks. The original clock-offset graphs also reproduced their
complete action sets at the recorded branch and history witnesses.
These checks supplement the proof on the entire formal domain.

\subsection{What does sharing save?}
Every method uses the same canonical windows, labelled branches, admissible
regions and terminal conditions. The sparse methods memoize the reachable
subproblem for each query. They store either one run per outstanding clause,
a shared frontier, a bit mask of pending obligations, or a reduced ordered
Boolean decision diagram for the residual formula. The diagram compiles an
OR over possible dwell starts per clause and shares identical Boolean
subfunctions. It uses the same sparse game solver. The ordered method stores
minimal winning states for the full domain. The scalar threshold method
also constructs the full policy and applies only to a single type.

The workload matters even on the same verified quadrotor graph. With 160
regions, 344 edges and 4.39 successor branches per edge on average,
19,734 of 26,880 time--region--history queries were winning for station A.
For complete-domain coverage the threshold method took 18.36\,ms
against 41.07\,ms for the sparse decision diagram; a single cold query took
18.41\,ms and 1.53\,ms, respectively. The physical graph, including
verification and post-set construction, cost 623\,ms before either
method could run. Including this shared step, producing a full policy took about
641 and 664\,ms, respectively.

\begin{table}[t]
\centering
\caption{Single-station quadrotor library. Median wall time in ms for an
executable policy from one start, five cached batches of 12 queries, and
all 26,880 formal states. Each workload includes policy construction.
Physical verification is common to all rows and excluded.}
\label{tab:window-bench}
\begin{tabular}{@{}lrrr@{}}
\toprule
Method & One start & Five batches & Full domain \\
\midrule
Per-window sparse & 1.77 & 2.36 & 59.50 \\
Grouped sparse & 2.01 & 2.57 & 64.40 \\
Bit-mask sparse & 1.72 & 2.48 & 51.19 \\
BDD progression & 1.53 & 1.99 & 41.07 \\
Ordered & 56.50 & 56.78 & 56.53 \\
Threshold & 18.41 & 18.53 & 18.36 \\
\bottomrule
\end{tabular}

\end{table}
\begin{figure}[t]
\centering
\includegraphics[width=\columnwidth]{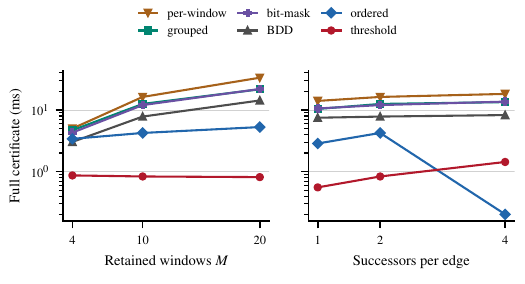}
\caption{Full-domain policies on synthetic one-type graphs at a fixed
horizon of 36 samples. Left: two successors per edge. Right: ten retained
windows. In the four-successor case only 21 of 6,072 queries win; the
ordered recursion terminates many branches early.}
\label{fig:window-scaling}
\end{figure}
Across seven synthetic settings and the robot library, the methods agreed on
all matched initial and batch queries and their full action sets. Seven
complete domains were also checked. The four-type full enumeration had
4,602,096 states and exceeded the prespecified 45,000-state coverage limit.
Its sparse batch queries all lost. On the synthetic two-successor graphs,
raising the number of retained windows from four to twenty at a fixed
horizon left the threshold build near 0.8--0.9\,ms. The sparse diagram
required about 3--14\,ms for a full policy (Fig.~\ref{fig:window-scaling}).
At four successors the mostly losing game favored ordered recursion.
On the verified library the sparse diagram won on all five cached batches
combined as well as on the isolated start (Table~\ref{tab:window-bench}).

\subsection{Repeated inspection in robot simulation}
Both the revisit in Fig.~\ref{fig:s1-mpc-return} and the patrol use
MuJoCo~\cite{todorov2012mujoco} and the Bitcraze Crazyflie~2 model from
MuJoCo Menagerie~\cite{menageriePinned}. On patrol the quadrotor must return to two stations while also
tracking a patrol reference that can be redirected during the mission.
Each measurement requires two consecutive samples. The allowed start
windows are $[0,3]$ and $[14,20]$ at A and $[6,10]$ and $[23,29]$ at B.
One decision tick comprises eight outer-loop samples of 0.2\,s.
Each method receives the current reference change at the same tick.

Three filters rank the same five longitudinal velocity primitives using
the same one-step tracking cost, unfinished-inspection cost and control
penalty. They allow, respectively, robust completion actions, robust
safety actions or nominal completion actions. A fourth controller (Full-H)
minimizes the sum of the stage costs on a deterministic cell-centroid
graph with all original window constraints and replans after each
observation. Its primitive grid is finite and its planning model is nominal.
The other arms use whole-cell transition covers.

\begin{figure}[t]
\centering
\includegraphics[width=\columnwidth]{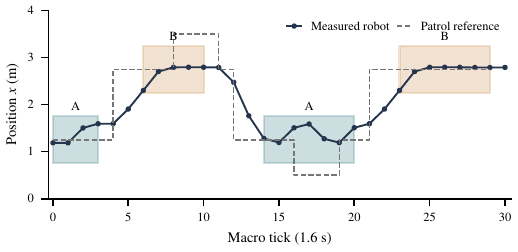}
\caption{Simulated patrol from $x_0=1.1875$\,m with wind seed 101. Shaded
rectangles show the station extent and allowed dwell-start samples. Points
show measured longitudinal position at each decision tick.}
\label{fig:window-robot}
\end{figure}

\ifdefined\WindowStageOne
Table~\ref{tab:window-robot-results} pairs three start positions with four
wind seeds. Figure~\ref{fig:window-robot} shows one patrol. All 48 runs
reached tick 30. Three filters met every window on all twelve runs;
Full-H missed the first B window four times. Of the eight robust-certified
starts, four also completed under Full-H. Safety-only and nominal each
retained robust membership on three of eight starts. Requiring every
executed action to be certified kept all three safety-only runs and
excluded every nominal run.
The four successes from uncertified starts under the completion filter count
as observed completions.
\else
The paired trial crosses three fixed start positions with four wind seeds for
each arm. All 48 patrols reach the 30-tick horizon. The robust, safety-only
and nominal filters each complete all four windows in 12/12 runs. Full-H
completes 8/12; each miss occurs at the first B window. Eight of the twelve initial pairs carry a robust certificate; four of
these eight also complete under Full-H. Among the certified starts, the
numbers that retain robust membership throughout are 8, 3, 3 and 0.
Requiring every \emph{executed action} also to be certified gives 8, 3, 0
and 0. The four successful robust-filter runs from uncertified initial
states count as observed completions only.
\fi

The robust filter's mean total cost is 19.33, compared with 21.19 for
safety-only. The paired difference is $-1.85$ (descriptive 95\% seed-block
interval $[-2.03,-1.76]$, conditional on the three starts).
Full-H costs less: on the eight runs where both methods finish every
inspection its mean advantage over the robust filter is 6.70.
All 56 trajectories, including four single-station threshold runs and four
branch-mechanism runs, stay within the sampled state, input and residual
bounds. The largest absolute residual component divided by its allowed
bound is 0.343 over 13,440 fine steps.

\begin{table}[t]\centering
\caption{Paired two-station patrols. Arm init.: starts feasible for that method.
Mission: all windows met. Retain and Exec.: robust membership and a certified
action at every tick, out of eight robust-certified starts. Fallback counts
ticks with an empty primary action set, out of 360 per arm.}
\label{tab:window-robot-results}
\begin{tabular}{lrrrrr}
\toprule
Method & Arm init. & Mission & Retain & Exec. & Fallback \\
\midrule
Completion & 8/12 & 12/12 & 8/8 & 8/8 & 12 \\
Safety only & 12/12 & 12/12 & 3/8 & 3/8 & 0 \\
Nominal & 8/12 & 12/12 & 3/8 & 0/8 & 4 \\
Full-H grid & 12/12 & 8/12 & 0/8 & 0/8 & 80 \\
\bottomrule
\end{tabular}

\end{table}

\paragraph{What changes the action?}
In a separate four-seed mechanism test, all runs begin at the same measured
state with the same history and execute hover. Two wind seeds lead to a
western successor and two to an eastern successor. The shared tracking
objective selects $+0.25$\,m/s in the west and hover in the east.
The complete next-action sets are identical, $\{-0.25,0,+0.25\}$\,m/s,
and both proposals pass. The robot executes branch-dependent feedback,
but either command would have retained the certificate in either cell.

A different comparison on the declared outer model isolates the effect
of history at a fixed state and clock. At tick 7, two certified prefixes
end at exactly the same four-dimensional state. One has finished B's
first dwell; the other has one true sample at B. The complete action sets
are $\{0,+0.25\}$ and $\{0\}$\,m/s, respectively. Both the run length
and the window coverage differ. This witness concerns modeled trajectories.

\subsection{When history and feedback change the permitted action}
\begin{figure}[!htb]
\centering
\includegraphics[width=\columnwidth]{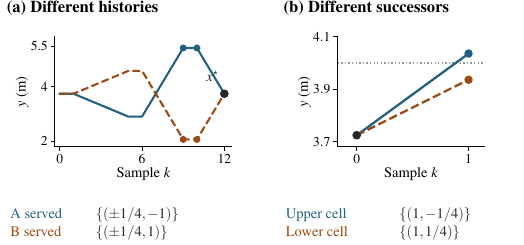}
\caption{Complete certified primitive sets $(v,\theta_{\rm ref})$ in two
fixed-offset unicycle examples. (a) The A-only and B-only prefixes end at
the same full state $x^\star=(25/4,15/4,0)$ at $k=12$, with zero bits and
runs. Colored dots mark dwell samples and the black dot marks the common
endpoint. The frontiers $(0,3)$ and $(3,0)$ give disjoint action sets.
(b) Markers show the midpoints of verified $y$-enclosures after the same
primitive $(1,-1/4)$. Each side of $y=4$~m has a different singleton
next-action set. Both panels use the declared transition model.}
\label{fig:s1-mechanisms}
\end{figure}

Figure~\ref{fig:s1-mechanisms} separates two decisions on verified
unicycle libraries. In Fig.~\ref{fig:s1-mechanisms}(a), the current
measurement cannot distinguish two legal histories. Whichever primitive a
controller selects using only the state, clock, bits and runs, it cannot
certify that primitive for both histories. A one-station example also
distinguishes partial service: at $k=14$ the same state, visited flag,
bit and zero run admit only $(1,1)$ at $c=2$, and also $(1/2,1/2)$ at $c=1$.
In Fig.~\ref{fig:s1-mechanisms}(b) both successors have the same updated
history $(c,r)=(3,0)$. The first primitive is certified because its next
action can depend on the observed cell, as allowed by~\eqref{eq:s1-joint-game}.

On 20 ordinary unicycle configurations all three filters finish the task. Disjoint
sets demonstrate where feedback is necessary for certification; their
frequency depends on the task and the transition cover.
\ifdefined\WindowStageOne\else

In the experiment of Fig.~\ref{fig:s1-mechanisms}(a), a third legal prefix
serves both stations and reaches the same endpoint, clock, bits and runs.
Its certified set contains both of the plotted sets.

The physical abstraction affects the outcome. In the original quadrotor
branch experiment ten wind seeds follow identical branches. An outer-model
pair with disjoint next-action sets reaches distinct cells with nested
sets in MuJoCo. In fixed-offset scaling tests, explicit bitset and compact
sparse caches visit equally many states; whole-domain ordered recursion is
\RSixRatioMin--\RSixRatioMax{} times slower than sparse compact on the matched
initial-region queries. An optimized full-horizon STL encoding is fastest in
the original linear feasibility comparison, and the sign of the replanning
cost difference changes with geometry and execution window. These checks
limit extrapolation to other graphs and control objectives.
\fi

\ifdefined\WindowStageOne\else
\subsection{Which comparison answers which question?}
\begin{table}[!ht]\centering
\caption{Matched inputs and scope of the policy comparisons.}
\label{tab:window-interfaces}
\begin{tabular}{@{}p{0.26\columnwidth}p{0.66\columnwidth}@{}}
\toprule
Comparison & Shared inputs and evaluated outputs \\
\midrule
Exact game solvers & Same graph, labels, deadlines and post set. Compare
membership and the complete certified action set for each query. \\
Three filters & Same five primitives, proposer, costs and wind phase.
Compare task completion and the robust eligibility of each executed action. \\
Full-H grid & Same primitive menu and information about the patrol.
Compare realized outcomes under its nominal cell-centre forecast. \\
\bottomrule
\end{tabular}\end{table}
The distinction between retaining membership and taking a certified action
matters empirically. A nominal action can leave a winning pair on the
observed branch even though another allowed branch loses. Three nominal
patrols remained in the robust winning set through $H$, yet none used
a robust-certified action on every step. Conversely, some simulations met
all deadlines even from a start that the whole-cell library could not
certify. These outcomes measure different properties of an abstraction.

The synthetic two-type game at four retained windows per type illustrates
why no single representation dominates. Its ordered full-domain build
took 5.51\,ms; the sparse diagram took 41.62\,ms. At four types the
predefined full-query cap bound the evaluation. On the verified one-type
graph the ordered construction examined 8,912 grouped target tuples;
the scalar sweep examined none. In the repeated application timing run,
full-workload process peaks were approximately 36.3\,MiB for threshold
and 35.1\,MiB for the diagram, including the interpreter, inputs and
separate audit. The measured threshold advantage is construction time.
It gives a precomputed lookup over the domain without establishing a
memory or closed-loop timing advantage.

\fi
\section{Scope and Conclusion}\label{sec:window-limits}
A finite family of inspection windows can have unequal widths, overlapping
releases and different deadlines. Reuse of a visit is what permits the
frontier representation. Independent work orders, arrivals after the start
of the task, and different precision requirements at a station can change
the remaining language. The threshold sweep is useful when a policy for every legal history
must be available before execution. A sparse solver was cheaper both
from one prescribed initial state and on five cached query batches. For several
types the joint winning set may have many incomparable histories.
Finally, the guarantee is relative to the labelled library and its stated
disturbance envelope. A measured simulation trajectory tests that interface;
establishing it for hardware requires additional evidence.\label{body-end}

\appendices
\section{Window Histories, Predecessors and Policies}\label{sec:window-proofs}

\subsection{Notation and initialization}
\begin{table}[!ht]\centering
\caption{Symbols for the general-window game. Time $k$ follows processing of sample $k$.}
\begin{tabular}{@{}lp{0.69\columnwidth}@{}}
\toprule
Symbol & Meaning \\
\midrule
$K_g,M_g$ & Original and inclusion-minimal interval counts for type $g$. \\
$I_{gi},\ell_g$ & Allowed start interval and number of further consecutive true samples required. \\
$L,H$ & Beginning of recorded history and terminal task sample. \\
$c,r$ & First served index in decreasing endpoint order, and capped true-run length. \\
$J(s),e_k$ & Window indices containing start $s$, and maximum live boundary after $k$. \\
$P_\rho,\sigma_b$ & Physical region and Boolean label guaranteed by a verified branch. \\
$T_{k,\rho}(r)$ & Largest winning boundary for a single type; $-1$ denotes an empty winning set. \\
\bottomrule
\end{tabular}\end{table}
All intervals below contain integers. For one type with a fixed effective
predicate, write $K$ for the raw number of windows, $M$ for the number
remaining, $I_i=[a_i,b_i]$ for a retained interval and $\ell+1$ for the
number of consecutive true samples required. For a bit prefix ending at $k$,
define directly from the raw history
\[
 D_I(k)=\{s\in I:s+\ell\le k,\
             \beta(s)=\cdots=\beta(s+\ell)=1\}.
\]
A window is served iff $D_I(k)\ne\varnothing$. Its expiry failure means
$D_I(k)=\varnothing$ and $k\ge\max I+\ell$.
Failure here checks expired deadlines; a prefix may already admit no
winning continuation because of its current run, safety or the dynamics.
The independent safety requirements are checked at their own active times.

\begin{IEEEproof}[Proof of Theorem~\ref{thm:window-history}]
If $I'\subseteq I$, then $D_{I'}(k)\subseteq D_I(k)$ for every $k$.
Every raw interval contains a minimal one, so serving all minimal
intervals serves every raw interval. If a raw interval $I$ has expired
unserved, its contained minimal interval $I'$ is also unserved and its
deadline is at most that of $I$. Conversely, an expired unserved retained
interval is itself raw. Expiry is irrevocable: after $b_i+\ell$ no new
witness in $I_i$ can finish. These arguments concern the conjunction;
a removed window can be served before any interval retained inside it.

After duplicates are removed, equal left or right endpoints would give
containment. Order the retained intervals by strictly decreasing $a_i$.
If $i<j$ and $b_i\le b_j$, then $I_j$ contains $I_i$, a contradiction.
Thus $b_i$ decreases strictly. Put
\begin{align*}
 u(s)&=\min(\{i:a_i\le s\}\cup\{M\}),\\
 v(s)&=\max(\{i:b_i\ge s\}\cup\{-1\}).
\end{align*}
The release condition gives a suffix, the right-endpoint condition a
prefix. Thus $J(s)=[u(s),v(s)]$ when $u(s)\le v(s)$, and is empty
otherwise. Suppose window $i$ is served by $s$ and $j>i$ is unserved.
Then $a_j<a_i\le s$ and $s\notin I_j$ imply $s>b_j$. The unserved
window $j$ had therefore expired by $s+\ell\le k$. On a nonfailed
prefix this is impossible. The served set is consequently a suffix,
possibly empty ($c=M$) or full ($c=0$). Its expiry check is $c\le e_k$
because the expired deadlines form the suffix beginning at $e_k$.

The update from $k$ to $k+1$ uses the integer-time identity
\begin{equation}\label{eq:window-adj}
 v(k+1-\ell)+1=e_k .
\end{equation}
Indeed $b_i\ge k+1-\ell$ holds exactly when $b_i+\ell> k$.
The number of such indices is both sides of \eqref{eq:window-adj},
including the empty and full prefix cases. For any valid source,
$c\le e_k$; hence every nonempty new witness interval intersects or
abuts the old suffix (including adjacency to an empty suffix).
A saturated next run finishes exactly one candidate start at this sample.
The explicit union of covered indices is then the suffix starting at
$\min(c,u)$ if $J$ is nonempty. Otherwise it is unchanged. Testing expiry
after this union gives the stated update on every formal source in $Q_k$.
An invalid source cannot be supplied as a live state, and failure maps
to itself for every later bit.

No valid retained witness starts before $L$; true samples before $L$
are therefore unnecessary to establish a valid start. Initializing the
trailing count at zero at $L-1$ counts precisely runs in the supplied
history. Induction from $L$ gives the stated suffix, capped run and expiry.
For every continuation the deterministic update extends the prefix exactly;
at $H$ all retained intervals must be served, which is equivalent to all
raw obligations. Taking the product over types preserves this equivalence.
Distinct prefixes with the same state may disagree about a removed window
or the identity of a past witness; neither can change the remaining
conjunction. The size bound follows by counting the formal pairs for
each type and one common failure. It counts neither a minimal automaton
nor the number of history-consistent states actually visited.
\end{IEEEproof}

An empty family is a true conjunction and is removed; if all families
are empty the game retains only safety and post constraints. An interval
with $a>b$ has no witness, even with an identically true signal.
Recorded history also covers all pre-mission safety samples. When
window data are changed after execution begins, a previously compressed
history need not initialize the changed task.

\paragraph{One visit and one witness}
A common start is possible on a given signal iff some $s\in\bigcap_i I_i$
starts a true dwell. The completed witnesses within one true segment
$[p,q]$ have starts $[p,q-\ell]$ when $q-p\ge\ell$. It serves every
window iff $q-p\ge\ell$ and $I_i\cap[p,q-\ell]\ne\varnothing$ for all
$i$ (nonempty family). For example, disjoint singleton windows
$[0,0],[2,2]$ with $\ell=1$ can both use the uninterrupted segment $[0,3]$.
A requirement for physically separated visits needs constraints that
exclude every such uninterrupted segment. The fixed-offset baseline
relations that use a common width or shift bound apply to
$[a-d,b-d]$, not to arbitrary interval families.

\subsection{Order and exact robust predecessors}
Write $(c,r)\preceq(d,s)$ when $c\ge d$ and $r\le s$, take the
componentwise product over types, and put $\bot$ below every live state.
For valid formal sources, the next run is nondecreasing in both current
run and bit. When one update yields a witness, any dominating source and
larger bit also yield a witness at the same start. The frontier map is
either the identity or $c\mapsto\min(c,u)$, both nondecreasing in the
ordinary order on $c$; gaining a witness only decreases $c$.
If the lesser state survives expiry, the greater one does too.
Thus $\Gamma$ is monotone in state and label on the full formal domain.
One can extend it by mapping all out-of-domain sources to $\bot$:
each $Q_k$ is upward closed. Terminal acceptance is upward closed as well.

\paragraph{One branch}
Fix $k$, $s=k+1-\ell$, a target $(d,t)\in Q_{k+1}$, and let
$u=u(s)$, $v=v(s)$. The desired predecessors are live sources whose
updated pair dominates $(d,t)$. A zero branch has no predecessors when
$t>0$ and otherwise has generator $(d,0)$. For a one branch form
\[
 (d,\max(0,t-1)),\qquad
 (e_k,\ell)\quad\text{if }J(s)\ne\varnothing\text{ and }u\le d .
\]
Keep only distinct minimal candidates in completion order (delete $y$
if $x\preceq y$ and $x\ne y$). Indeed $d\le e_{k+1}\le e_k$.
The updated run dominates $t$ iff $r\ge\max(0,t-1)$. If $c\le d$
this already suffices. Otherwise only a newly completed witness can
suffice; it requires $r\ge\ell$, nonempty $J$ and $u\le d$.
Equation~\eqref{eq:window-adj} supplies adjacency for every valid $c$.
The target bound $d\le e_{k+1}$ also ensures no successful predecessor
is rejected by the next expiry check.

\paragraph{Several branches and types}
For a fixed edge and one target per branch, work in one coordinate with
targets $(d_b,t_b)$. Write $I_z=\{b:\sigma_b=z\}$ for $z=0,1$. If a zero
branch requests $t_b>0$, the feasible set is empty. Otherwise put
$C_0=\min(\{e_k\}\cup\{d_b:b\in I_0\})$. Zero branches together
impose $c\le C_0$. When $I_1$ is empty the sole candidate is $(C_0,0)$.
Otherwise let $C_1=\min_{b\in I_1}d_b$ and
$R_1=\max_{b\in I_1}t_b$.
All one branches have the same updated state in this coordinate, so
their targets jointly request $c'\le C_1$ and $r'\ge R_1$.
Intersecting the preceding one-branch cases with $c\le C_0$ gives
\begin{align*}
 &(\min(C_0,C_1),\max(0,R_1-1)),\\
 &(C_0,\ell)\qquad\text{if }J(s)\ne\varnothing\text{ and }u\le C_1 .
\end{align*}
Again minimize. For multiple types take the Cartesian product of
per-type candidates and retain its minima, since coordinate conditions
for a fixed target tuple are conjunctive. This supplies at most $2^G$
candidates per tuple. To recurse on antichains, take the union over
edges and one winning generator per successor branch, then retain the
minima of that union. Finite upward-closed sets equal the upward closure
of their minima; backward induction therefore recovers the explicit
winning set. Different branches can require different generators.
Tuple enumeration in this general construction may be expensive.

\paragraph{Why independent coordinate thresholds can fail}
Take two distinct types with $\ell_1=\ell_2=0$, one window $[0,1]$ each
and $H=1$. At $k=0$ let one edge ensure bits $(1,0)$ at the next
sample and another ensure $(0,1)$, all regions admissible and post.
An initial history which has served either type is winning. A history
which has served neither is losing, although its individual coordinates
each occur in a winning history. This rules out reconstructing the
winning product solely from its coordinate projections.

\subsection{Threshold computation and action recovery}
\begin{IEEEproof}[Proof of Theorem~\ref{thm:window-threshold}]
Terminal acceptance gives $T_H(r)=0$ at admissible post regions and
$-1$ otherwise, even when $H=0$. For the induction step assume exact
successor thresholds. Upon fixing a run and an edge, the next run on each
branch is independent of $c$. A negative successor threshold rules out
the branch for all $c$. Otherwise, a new witness in a nonempty $J$ yields
$c'=\min(c,u)$, so $c'\le t_b$ holds for every $c\le e_k$ if $u\le t_b$,
and exactly for $c\le t_b$ if $u>t_b$. Without such a witness $c'=c$.
This gives \eqref{eq:window-branch}, including expiry because
$t_b\le e_{k+1}$. The intersection across branches is the prefix through
their minimum; the union across edges is the prefix through the largest
of these minima. No action witnesses an empty edge set. This proves
the formula for all formal states, including history-inconsistent ones.
A maximizing edge covers its entire prefix. The complete certified
action set at $c$ consists of all outgoing edges with branch minimum
at least $c$ at an admissible source. It can be recovered by scanning
the edges without materializing a policy for every $c$.

For each run, each time layer initializes $|\mathcal R|$ entries and
performs constant arithmetic work per branch for each outgoing edge.
Storing the maximizer and threshold costs a constant number of entries
per $(k,\rho,r)$, with no edge needed at $H$. An antichain for one
region and time has at most $\ell+2$ elements because all states
with the same run are comparable. Threshold entries are integers
in $[-1,M]$. These are arithmetic-operation and entry bounds; bit
width grows with the range, and the verified library is an input.
Two value layers suffice for the initial feasibility decision alone;
recovering the time-indexed policy from them may require recomputation.
\end{IEEEproof}

\begin{algorithm}[t]
\caption{Single-type backward sweep on a fixed library}
\label{alg:window-sweep}
\begin{algorithmic}[1]
\Require canonical time indices, verified branches, $\RA$, $\Kpost$, $H$
\State initialize $T_{H,\rho}(r)$ as above for all $\rho,r$
\For{$k=H-1,\ldots,0$}
  \For{each $\rho,r$}
    \State set $T_{k,\rho}(r)\gets-1$; clear the edge witness
    \If{$\RA_k(\rho)$}
      \For{each edge $e\in A(\rho)$}
        \State $v\gets\min_{b\in\Succ(e)}p_b$ by \eqref{eq:window-branch}
        \If{$v>T_{k,\rho}(r)$}
          \State store $v$ and $e$ for this $(k,\rho,r)$
        \EndIf
      \EndFor
    \EndIf
  \EndFor
\EndFor
\end{algorithmic}
\end{algorithm}

Window preprocessing is separate. For example, sorting by decreasing
release and increasing right endpoint on ties permits a linear scan
after the sort: discard the next interval if its right endpoint is
at least the smallest previously seen right endpoint. A previously
retained interval realizing that minimum witnesses the containment.
Otherwise retain the interval and update the minimum. Moving indices
through the two strictly ordered endpoint arrays then tabulate $u,v,e$
by time in $O(M+H-L+1)$ per type. Thus the single-type preprocessing
can be implemented in $O(K\log(K+1)+H-L+1)$ arithmetic operations.
Certifying physical edges, computing $\RA$ and answering online queries
incur additional costs.

\subsection{Lifting the policy to the executed system}
\begin{IEEEproof}[Proof of Theorem~\ref{thm:window-feedback}]
Induct backward for upward closure. At the terminal time it follows from
acceptance. If some edge wins for a given state, increasing that state
in completion order preserves admissibility and improves every branch
update. Every branch still wins at the next layer by induction.
The game on explicit retained sets and runs is identical under suffix
decoding of every valid formal pair by \eqref{eq:window-adj}. For raw
windows, exactness holds on states generated by a common bit prefix:
Theorem~\ref{thm:window-history} gives the same rejection and terminal
acceptance under every extension. Induction on the future tree preserves
the existential action and universal branch decisions because both
recursions use the same labels, regions, safety and post conditions.

At a winning region and live state take any certified edge. By verification
its law is admissible. For each actual disturbance there is at least one
branch for the measured successor with both region containment and
$\sigma_b\le\beta(x^+)$. The lower-label update wins on that branch;
processing the realized bits gives a dominating state by monotonicity
and still wins. Induction forward from the initialized state gives
admissibility until $H$ and acceptance in $\Kpost$. A verified invariant
post law can maintain the post constraints thereafter. Historical safety
was checked during initialization. Each minimal window has a witness
at its own deadline, and each deleted window contains such a witness by
its own deadline. This proves the original task for every disturbance.

In a proposal filter, use a currently certified proposed edge when
available and otherwise use the stored winning edge for the current tick,
region and monitor state. Any delayed proposal requires revalidation for
the decision to which it is applied. A branch matching only the measured
region can have an incorrect lower label and cannot justify the induction.
\end{IEEEproof}

\paragraph{Tube MPC instantiation}
For a linear tube with invariant error set, propagate the centre with its
executed nominal dynamics. Assume the checked nominal plan satisfies
tightened safety and one-sided labels, terminates in a certified nominal
region, and the nominal certifying library supplies a verified one-step
extension from each winning pair. Actual \emph{nominal-centre} predicate
bits at the executed successor dominate the planned lower bits, so the
executed monitor state dominates the planned one. Shift the overlapping
plan using the same absolute times, nominal inputs and lower bits.
Its monitor suffix still dominates the old planned suffix. If its endpoint
is $H$, upward closure preserves that terminal certificate; otherwise
append the one-step verified extension at its endpoint. Thus a verified
plan for the next tick exists after each executed step before $H$.
At the last step the terminal state accepts; each true tightened nominal
predicate implies the intended physical predicate for every state in the
tube. Historical nominal bits and tightening bounds must be available
on the active pre-mission samples. This proof does not mix measured
nonlinear bits with nominal-centre bits in the same run.

\paragraph{Integer monitor rows for an arbitrary window family}
At successor time $t$, impose valid source $0\le c\le e_{t-1}$ and
$0\le r\le\ell+1$. For a claimed bit $z$, use integral next states
$c',r'$ and auxiliary binaries $u,w$:
\begin{align*}
  & (\ell+1)u\le r\le\ell+u,\quad (\ell+1)w\le r'\le\ell+w,\\
  &0\le r'\le (\ell+1)z,\quad r'\le r+1-u,\\
  &r'\ge r+1-u-(\ell+1)(1-z).
\end{align*}
These rows force $u=[r=\ell+1]$, $w=[r'=\ell+1]$ and the required
capped run. If $J(t-\ell)=\varnothing$, impose $c'=c$. Otherwise let
$d=\min J(t-\ell)$, introduce binaries $v,h$ and impose
\begin{align*}
 &(d+1)v\le c\le d+(M-d)v,\\
 &h\le w,\quad h\le v,\quad h\ge w+v-1,\\
 &c-Mh\le c'\le c,\qquad |c'-d|\le M(1-h).
\end{align*}
Here $v=[c>d]$ and $h=w\wedge v$. Thus $c'=d$ exactly when an eligible
witness lowers the frontier; otherwise $c'=c$ (which can also equal $d$).
Finally impose $0\le c'\le e_t$. By \eqref{eq:window-adj} no additional
adjacency indicator is needed on this source domain. The integer rows
are feasible for given source and bit exactly when the monitor survives,
and then return its unique next pair. This step needs at most two new
integer variables and four auxiliary binaries besides the supplied bit.
Geometry, terminal selection and the plant equations have separate rows.

\section*{Repeated-window certificate benchmark protocol}
The seven synthetic settings use eight regions on a ring, three actions per region, a four-region terminal set,
and a fixed horizon of 36. Each additional successor is a distinct shift around the ring. Every type has
$\ell=1$. Four partially overlapping sectors define the potential predicates. Opposite sectors compete.
Every method receives the same canonical intervals, edge order, lower labels and admissibility table.
Each synthetic type includes one duplicate and one containing interval in its raw inputs.
The verified application uses the two original inspection windows of station A.
One query workload starts from the specified initial sample. A second consumes five consecutive batches of
12 queries drawn from fixed lower-label walks, retaining all caches. After timing, every permitted edge is
enumerated for each query. For full-domain preparation the sparse methods populate their policy from all
legal formal states, including unreachable histories. The eager representations cover the same domain
directly. A separate readback pass compares membership for all 88,704 formal queries in the seven
enumerated settings. Five solvers apply to multiple types, and six to one type.
\begin{table*}[t]\centering
\caption{Full-domain construction in the initial campaign (median wall ms, three builds per cell).
The matched column counts winning queries across five batches of 12. $B$ is the actual number of successors
per edge on the ring and the mean on the application. The four-type full sweep exceeded the declared
45,000-state enumeration limit. The threshold only applies to one type.}
\label{tab:window-all-builds}
\begin{tabular}{@{}lrrrrrrrrrr@{}}\toprule
Game & $G$ & $M$/type & $B$ & Matched & Per-window & Grouped & Bit mask & BDD & Ordered & Threshold \\
\midrule
Ring & 1 & 4 & 2 & 60 & 4.99 & 4.68 & 4.29 & 3.03 & 3.40 & 0.87 \\
Ring & 1 & 10 & 2 & 30 & 16.23 & 12.50 & 11.97 & 7.80 & 4.23 & 0.83 \\
Ring & 1 & 20 & 2 & 42 & 33.27 & 21.59 & 21.83 & 14.22 & 5.27 & 0.81 \\
Ring & 1 & 10 & 1 & 33 & 13.99 & 10.36 & 10.56 & 7.47 & 2.86 & 0.55 \\
Ring & 1 & 10 & 4 & 0 & 18.17 & 13.38 & 13.63 & 8.24 & 0.20 & 1.43 \\
Ring & 2 & 4 & 2 & 49 & 99.90 & 86.37 & 72.60 & 41.62 & 5.51 & --- \\
Ring & 4 & 4 & 2 & 0 & --- & --- & --- & --- & --- & --- \\
Verified robot & 1 & 2 & 4.39 & 18 & 57.86 & 65.53 & 53.06 & 41.15 & 55.67 & 17.92 \\
\bottomrule\end{tabular}\end{table*}
The first campaign ran on an arm64 macOS host with CPython 3.11.16. It used one worker and one BLAS
thread at a time. Other project processes could use the host. Seed 83017 fixes the histories, tasks
and job order. A separate process ran each case, method and workload; three repetitions rebuilt the
policy. All 399 repetitions completed within ten seconds and 600 MiB per process. Peak RSS includes
the interpreter, inputs, repeated builds and audit; on the robot full-domain workload it ranged from
35 to 39 MiB. Runtime distributions and all labelled edges accompany the raw results.
The application ordered recursion examined 8,912 joined target tuples from 12,563 ungrouped tuples,
with at most two generators in a region at a time. The threshold recursion examined zero target tuples.
In a separate preprocessing ablation the same per-window sparse solver received either all original
clauses or their inclusion-minimal subset and replayed the entire input history in both arms. All
eight sets of 60 answers agreed. Across synthetic settings the ratio of raw to canonical median wall
time ranged from 0.95 to 1.95 for five batches. Input reduction helps some configurations more than others.

\section*{Repeated inspection protocol and scope of the interface}
A macro action commands one of five longitudinal speeds,
$\{-1/2,-1/4,0,1/4,1/2\}$\,m/s, with a fixed lateral feedback law.
The two inspection strips have longitudinal extents $[0.75,1.75]$ and
$[2.25,3.25]$\,m. The workspace, velocity and input constraints are
checked on the eight fine outer-model samples within a macro primitive.
The measured initial longitudinal positions are chosen by resetting the
simulator at 1.1875, 1.6875 or 3.0625\,m. Four fixed wind seeds (101--104)
are crossed with all three starts and four methods. Each start is repeated
under all seeds and has the same nominal initial bits. The physical graph
has 160 regions, 344 verified actions and 134 post regions. A separate
single-station run uses the threshold policy and the four wind seeds.
The ordinary two-station runs instead use the joint game.

At each sample the shared stage cost is
$(x^+-p_k)^2+0.45(x^+-a_k)^2+0.06(v_{\rm ref}-v_x)^2$,
where $x^+$ is the next longitudinal position, $p_k$ the current patrol
target and $a_k$ the station with the earliest unfinished deadline,
or $p_k$ if no inspection remains. The current target changes at prescribed
samples; a controller sees each redirect upon issue. Costs are evaluated
on the measured trajectory. The three filters share the one-step proposer.
The full-horizon comparator minimizes the sum of this cost on nominal
cell-centre transitions, using exact original-window masks, a terminal post
constraint and all announced reference information. It replans after
the next measurement. This differs from both the whole-cell nominal
completion filter and the continuous-input linear STL reference in the
fixed-offset study.

The model-history comparison matches the full four-dimensional state at
tick 7. Both modeled prefixes start from the same state, and their final
disturbances are within the declared componentwise bounds. The partial B
visit has changed both the run and the coverage. In the distinct MuJoCo
branch experiment, the four wind seeds 111--114 share an initial hover.
The first two tracking preferences are designed to distinguish measured
successors near a cell boundary. The two possible next decisions pass the
certificate on both branches. Exact action-set equality was checked by
original-window enumeration in addition to the grouped policy.

The robot simulations stop at or before the terminal sample. The invariant
post condition is checked on the model. Residual checks on the measured
state and input trace refer to sampled outer dynamics, with contact checks
in simulation. A finite set of traces does not certify a disturbance bound
on hardware. An elapsed-time overrun in the simulator is checked after
the decision routine returns; wall-clock measurements therefore describe
the implementation and cannot establish hard real-time preemption.

Wind seeds are the units for paired resampling: for each of four resampled
seeds retain the entire paired block of three fixed starts. The descriptive
95\% percentile intervals are conditional on these starts and on the small
collection of observed wind phases; multiple ticks within a trajectory are
never counted as independent runs. Completion, safety-only, nominal and
Full-H respectively have mean total costs 19.33, 21.19, 21.37 and 14.03
over all twelve configurations, including the four missed Full-H missions.
On configurations where both missions finish every window, the robust
filter costs 6.70 more than Full-H (paired interval $[6.13,7.28]$).
Respectively, 12, 0, 4 and 80 decisions use a backup because the primary
action set is empty. These counts are out of 360 decisions per arm.
No optimizer outage is injected. The 95th percentiles for action selection
are 0.075, 0.018, 0.063 and 7.194\,ms. A separate replay of observed branch,
bit and monitor updates takes 0.031--0.032\,ms at the 95th percentile.
No decision exceeds the 1.6-s budget in the 1,440 ordinary-patrol ticks.
Individual percentiles of these two timing scopes cannot be added to infer
an end-to-end latency percentile.

An initial comparison incorrectly required a robust backup before even
calling the nominal full-horizon optimizer. That dispatch rule truncated
four Full-H runs at tick 7. The corrected experiment uses an available
nominal plan and checks backup availability only when needed. Every
original configuration was rerun with the same seeds, costs and geometry;
the corrected trajectories are the ones summarized here.

\section*{Fixed-offset specialization and supporting experiments}
The following sections give the equal-width specialization, the linear tube
implementation, its verification assumptions and the original clock-offset
experiments. In these sections $d$ indexes a translation of a common
window. The strict shared-witness comparisons are specific to that family.
\section{Problem Formulation}\label{sec:problem}

\subsection{Nominal system and tube interface}
We consider the constrained linear system
\begin{equation}\label{eq:sys}
x_{k+1}=Ax_k+Bu_k+w_k,\qquad y_k=Cx_k,
\end{equation}
with $x_k\in\R^n$, $u_k\in\R^m$, $y_k\in\R^p$, polytopic constraints
$x_k\in\mathcal X$, $u_k\in\mathcal U$, and disturbances $w_k$ in a compact set
$\mathcal W$. We use the standard tube interface~\cite{mayne2005tube,mayne2006output}:
a nominal state evolves as $z_{k+1}=Az_k+Bv_k$ under a nominal input $v_k$, the
applied input is $u_k=v_k+K(x_k-z_k)$, and the error $e_k=x_k-z_k$ obeys
$e_{k+1}=(A+BK)e_k+w_k$.

\begin{assumption}[Tube interface]\label{ass:tube}
The set $\mathcal E$ satisfies $(A+BK)\mathcal E\oplus\mathcal W\subseteq\mathcal E$
and $e_0\in\mathcal E$. The nominal constraints are
$\mathcal Z=\mathcal X\ominus\mathcal E$ and
$\mathcal V=\mathcal U\ominus K\mathcal E$, and
$C\mathcal E\subseteq\Ball=\{\eta\in\R^p:\|\eta\|_\infty\le\varepsilon\}$.
\end{assumption}

The interface is standard (nominal case $\mathcal W=\mathcal E=\{0\}$); the
nominal center is never reset to the measured state, and all arguments below
refer to it.

\subsection{Timing offsets and the guaranteed signal family}
Time is discrete, and all time intervals are integer intervals. Let $D\in\mathbb N$,
$\Dset=[0,D]$, and let $\ybar_t=Cz_t$ be the nominal output for $t\ge0$.

\begin{assumption}[Pre-mission history]\label{ass:history}
A recorded nominal output $\ybar_t$ is available for $t\in[-D,-1]$; the actual
pre-mission outputs satisfy $y_t-\ybar_t\in\Ball$ for $t\in[-D,-1]$ (for
example, because $e_t\in\mathcal E$ already held before the mission); and
$\ybar_t$ satisfies the tightened safety predicates of
Definition~\ref{def:tight} at every $t\in[-D,-1]$ at which they are active
(Section~\ref{sec:certs}).
\end{assumption}

Without such a history, the windows must be shifted or $D$ reduced; a
fictitious history must not be substituted, since offset $d$ reads $y_{t-d}$
with $t-d<0$ for $t<d$.

\begin{definition}[Guaranteed signal family]\label{def:family}
For $d\in\Dset$ and $\eta:\mathbb Z\to\Ball$, let
$\ytil^{d,\eta}(t)=\ybar_{t-d}+\eta(t)$ for $t\in[0,H]$.
\end{definition}

The offset $d\ge0$ is constant and common to all outputs: a fixed
misalignment of at most $hD$ seconds between task clock and trajectory, as for
a start command with bounded latency. Componentwise shifts, input or
measurement delays and time-varying retiming (Remark~\ref{rem:jitter}) are
different models and are not covered.

\subsection{Task fragment}
\begin{definition}[Reach--dwell--avoid fragment]\label{def:task}
For closed sets $S_i,R_j\subseteq\R^p$ and integers $0\le\alpha_i\le\omega_i$,
$0\le a_j\le b_j$ and $\ell_j\ge0$, let
\begin{equation}\label{eq:task}
\varphi=\bigwedge_{i=1}^{n_s}\G_{[\alpha_i,\omega_i]}\mu_{S_i}\ \wedge\
\bigwedge_{j=1}^{n_r}\F_{[a_j,b_j]}\G_{[0,\ell_j]}\mu_{R_j},
\end{equation}
where $(y,t)\models\mu_S$ iff $y(t)\in S$, and $\G$ and $\F$ carry the usual
discrete-time semantics~\cite{maler2004monitoring}. Let
$\hrz(\varphi)=\max\{\max_i\omega_i,\max_j(b_j+\ell_j)\}$,
$\Delta_j=b_j-a_j$, and choose a terminal certification horizon
$H\ge\hrz(\varphi)$. Each task keeps its window (offset $d$ of reach task $j$
expires at nominal time $b_j-d+\ell_j$); only the post-mission terminal
condition is imposed at $H$.
\end{definition}

Predicate sets are closed; an avoid requirement is a closed safety set with
an explicit clearance (Section~\ref{sec:tight}), and a conjunction of two
eventualities imposes no order.

\subsection{Spatial tightening and the Boolean abstraction}\label{sec:tight}
\begin{definition}[Tightening]\label{def:tight}
Tightened sets satisfy $\bar S_i\oplus\Ball\subseteq S_i$ and
$\bar R_j\oplus\Ball\subseteq R_j$. The bit map
$\chi:\R^p\to\{0,1\}^{n_r}$ is $\chi_j(y)=1$ iff $y\in\bar R_j$. For
$t\in[-D,H]$, the nominal bit signals are $\beta_j(t)=\chi_j(\ybar_t)$ and
$\gamma_i(t)=1$ iff $\ybar_t\in\bar S_i$. The Boolean task $\varphi_\beta$ is
\eqref{eq:task} with $\mu_{S_i}$ and $\mu_{R_j}$ replaced by $\gamma_i$ and
$\beta_j$.
\end{definition}

For $\{y:Gy\le f\}$ the largest choice is
$\{y:Gy\le f-\varepsilon|G|\mathbf 1\}$. The tightening is sufficient, not the
exact robust semantics.
\emph{Obstacles.} For a closed obstacle $\mathcal O_i$ and a clearance
$\delta_i>0$, $S_i=\{y:\dist(y,\mathcal O_i)\ge\delta_i\}$ and
$\bar S_i\subseteq\{y:\dist(y,\mathcal O_i)\ge\delta_i+\varepsilon\}$ in the
$\ell_\infty$ distance; the supplement gives a polyhedral choice.

\begin{lemma}[Spatial soundness]\label{lem:spatial}
\StmtSpatial
\end{lemma}
\ifshowproofsketches
\begin{IEEEproof}
If $\beta_j(t-d)=1$, then $\ybar_{t-d}\in\bar R_j$ and
$\ytil^{d,\eta}(t)\in\bar R_j\oplus\Ball\subseteq R_j$; the same holds for
$\gamma_i$. Since \eqref{eq:task} contains no negation, satisfaction is
preserved.
\end{IEEEproof}
\fi

\begin{problem}\label{prob:main}
Given \eqref{eq:sys}, Assumptions~\ref{ass:tube}--\ref{ass:history}, the task
\eqref{eq:task}, and a prediction horizon $N\ge1$ that may be much shorter than
$H$, design a receding-horizon controller such that, once a verified feasible
candidate for its first optimization problem is available, for every
disturbance sequence the constraints hold and
$(\ytil^{d,\eta},0)\models\varphi$ for all $d\in\Dset$ and all
$\eta:\mathbb Z\to\Ball$.
\end{problem}

Since $0\in\Dset$ and $Ce_t\in\Ball$, the actual sampled output then satisfies
$\varphi$ under every offset. Guarantees concern sampling instants of the
linear model; inter-sample behaviour and the nonlinear vehicle are empirical
evidence only.

\begin{remark}[Fixed offsets are not jitter]\label{rem:jitter}
For $\F_{[1,2]}\mu$, $\Dset=\{0,1\}$ and bits $1,0,1$ at $t=0,1,2$, each fixed
offset succeeds, but $d(1)=0$, $d(2)=1$ reads the bit at $t=1$ twice and
fails.
\end{remark}

\subsection{Disturbed nonlinear systems and causal policies}\label{sec:nonlinear}
The compact state does not use linearity; Section~\ref{sec:branching} also
treats a known discrete-time system
\begin{equation}\label{eq:nlsys}
x_{k+1}=f(x_k,u_k,w_k),\quad w_k\in\mathcal W(x_k,u_k),\quad u_k\in U(x_k),
\end{equation}
with measured state, output $h(x_k)$ and bits $\chi_j(h(x))$. The objective
is a causal policy $u_k=\pi_k(x_{-D},\dots,x_k)$ with
\[
\exists\pi\ \ \forall(w_0,\dots,w_{H-1})\ \ \forall d\in\Dset:\ \
((\gamma,\beta)^{\pi,w}(\cdot-d),0)\models\varphi_\beta,
\]
over admissible disturbance sequences and the recorded history; the
completing witness may differ between realizations and offsets, and the
future must be feedback (counterexamples in the supplement). Guarantees
concern the declared discrete system.

\section{History-Consistent Completion State}\label{sec:state}

We work with the tightened bits (Lemma~\ref{lem:spatial} transfers the
results) and drop the index $j$ for one reach task.

\subsection{Offset coverage}
A \emph{dwell witness} is a start time $s\ge-D$ with
$\beta(s)=\dots=\beta(s+\ell)=1$; it is \emph{completed} at $s+\ell$. Under
offset $d$ the witness appears at task time $s+d$, so it serves exactly the
offsets
\begin{equation}\label{eq:I}
I(s)=\{d\in\Dset:\ a\le s+d\le b\}=[a-s,\,b-s]\cap\Dset .
\end{equation}

\begin{definition}[Coverage, run length, failure]\label{def:coverage}
The coverage at time $k$ is $\Cov(k)=\bigcup\{I(s): s\text{ a witness},\
s+\ell\le k\}$. The run length $r(k)$ is the number of consecutive ones of
$\beta$ ending at $k$, truncated at $\ell+1$. Offset $d$ is \emph{expired} at
$k$ if $d\notin\Cov(k)$ and $k\ge b-d+\ell$; the task is \emph{failed} at $k$
if some offset is expired.
\end{definition}

Only executed samples enter $\Cov(k)$. One ``visited'' flag is insufficient:
for $\F_{[1,1]}\mu$, $\Dset=\{0,1\}$ and bits $0,1$, only offset $0$ is
served.

\begin{lemma}[Coverage semantics]\label{lem:L1}
\StmtCoverage
\end{lemma}
\ifshowproofsketches
\begin{IEEEproof}
(a) Substitute $s=t-d$: every $d$ needs a witness $s$ with $a\le s+d\le b$,
i.e., $d\in I(s)$, and every such witness has $s\le b$ and is completed by
$b+\ell\le H$. (b) $\{t-d:t\in[\alpha,\omega],d\in\Dset\}=[\alpha-D,\omega]$.
\end{IEEEproof}
\fi

\begin{proposition}[Exact compilation of the tightened task]\label{prop:compile}
\StmtCompile
\end{proposition}
\ifshowproofsketches
\begin{IEEEproof}
Boolean STL is shift invariant,
$(\xi(\cdot-d),t)\models\psi\iff(\xi,t-d)\models\psi$, by induction on $\psi$;
the supplement treats every operator, including bounded until.
With $\tau=D-d$ the left side reads $\forall\tau\in[0,D]:(\xi,-D+\tau)\models\psi$.
\end{IEEEproof}
\fi

Generic STL-MPC with history
handling~\cite{ilyes2025rcpe,ilyes2026rotogo,yu2022decomp} therefore applies
directly and is a baseline; the rest of this section quantifies what the
structure of \eqref{eq:task} saves.

\subsection{Coverage is a suffix}
\begin{lemma}[Irrevocable expiry]\label{lem:expiry}
\StmtExpiry
\end{lemma}

\begin{theorem}[Suffix coverage]\label{thm:suffix}
\StmtSuffix
\end{theorem}
\ifshowproofsketches
\begin{IEEEproof}
Let $d\in\Cov(k)$ be served by a witness $s$ with $a\le s+d\le b$ and
$s+\ell\le k$, and let $d'\in\Dset$ with $d'>d$. Then $s+d'>s+d\ge a$, so
$d'\notin I(s)$ forces $s+d'>b$, i.e., $b-d'+\ell<s+\ell\le k$. If also
$d'\notin\Cov(k)$, then $d'$ is expired at $k$, contradicting non-failure.
Hence $d\in\Cov(k)$ implies $[d,D]\subseteq\Cov(k)$.
\end{IEEEproof}
\fi

Larger offsets see the trajectory later in task time, so their deadlines
come first.

\subsection{Compact monitor}
Theorem~\ref{thm:suffix} replaces the $(D+1)$-bit coverage by one integer.
For task $j$ let
$\Qset_j=[0,D+1]\times[0,\ell_j+1]$, let $\Qset=\prod_j\Qset_j$, and add an
absorbing failure state $\bot$, to which all failed histories are mapped;
their coverage need not be a suffix. Let
$\cbar_{j,k}=\min\{D+1,\max\{0,b_j+\ell_j-k\}\}$.

\begin{definition}[Compact monitor]\label{def:compact}
For task $j$, time $k$, state $(c,r)\in\Qset_j$ and bit $\beta\in\{0,1\}$, let
$r^+=\min\{\ell+1,r+1\}$ if $\beta=1$ and $r^+=0$ otherwise. If $r^+\le\ell$,
let $c^+=c$. If $r^+=\ell+1$, let $s=k-\ell$, $\lo=\max\{0,a-s\}$,
$\hi=\min\{D,b-s\}$, and
\begin{equation}\label{eq:compact-update}
c^+=\begin{cases}\min\{c,\lo\}, & \lo\le\hi\ \text{and}\ \hi\ge c-1,\\
c, & \text{otherwise.}\end{cases}
\end{equation}
Set $\Gamma_{j,k}((c,r),\beta)=\bot$ if $c^+>\cbar_{j,k}$, and $(c^+,r^+)$
otherwise. The joint update $\Gamma_k(q,\beta)$ applies $\Gamma_{j,k}$ to each
component and returns $\bot$ if any component does; $\Gamma_k(\bot,\cdot)=\bot$.
The monitor starts from $\qinit=((D+1,0))_j$ at time $-D-1$ and processes
$q_t=\Gamma_t(q_{t-1},\chi(\ybar_t))$.
\end{definition}

The test checks expiry of the largest uncovered offset $c^+-1$, which has the
earliest deadline. Let $\Qset_k=\{q\in\Qset:c_j\le\cbar_{j,k}\ \forall j\}$;
non-failure does not imply that a satisfying continuation exists, which
Section~\ref{sec:certs} certifies.

\begin{corollary}[Equivalence and size]\label{cor:compact}
\StmtCompact
\end{corollary}
\ifshowproofsketches
\begin{proofsketch}
Induction on $k$. If no witness completes at $k$, the coverage is unchanged. If
$s=k-\ell$ completes, $\Cov(k)=[c,D]\cup[\lo,\hi]$. This is the suffix
$[\min\{c,\lo\},D]$ when $\hi\ge c-1$; otherwise offset $c-1$ stays uncovered
with $b-(c-1)+\ell<k$, which the test detects. The supplement gives the full
argument.
\end{proofsketch}
\fi

Arbitrary bitsets need $2^{n_r(D+1)}\prod_j(\ell_j+2)$ states: the dependence
on $D$ drops from exponential to linear per task, still exponential in $n_r$.
This counts encodable states, not running time.

\subsection{Order and monotonicity}
\begin{definition}[Completion order]\label{def:order}
For $q,q'\in\Qset$, $q\preceq q'$ iff $c_j\ge c'_j$ and $r_j\le r'_j$ for all
$j$; moreover $\bot\preceq q$ for all $q$. $\Acc(q)$ holds iff $q\neq\bot$ and
$c_j=0$ for all $j$.
\end{definition}

The order is a product of $2n_r$ chains.

\begin{lemma}[Monotonicity]\label{lem:mono}
\StmtMono
\end{lemma}
\ifshowproofsketches
\begin{proofsketch}
Componentwise, $r^+$ is nondecreasing in $r$ and $\beta$; the right side of
\eqref{eq:compact-update} is nondecreasing in $c$ and never exceeds $c$; the
failure test is monotone in $c^+$. The statement holds on all of $\Qset$, not
only on reachable states.
\end{proofsketch}
\fi

\subsection{When more than one visit is needed}
\begin{definition}[Semantic baselines]\label{def:baselines}
For one reach task and a bit signal $\beta$ on $[-D,b+\ell]$:
$\beta\in\Fex$ (exact coverage, possibly by \emph{separated visits}) iff
$\Cov(b+\ell)=\Dset$;
$\beta\in\Fsh$ (shared task time) iff some $t\in[a,b]$ has $\beta=1$ on
$[t-D,t+\ell]$, so that one task time serves all offsets;
$\beta\in\Fsk$ (\emph{single witness}) iff some $s\in[a,b-D]$ has $\beta=1$ on
$[s,s+\ell]$, i.e., $(\beta,0)\models\F_{[a,b-D]}\G_{[0,\ell]}\beta$;
$\beta\in\Fsi$ (\emph{single continuous visit}) iff the witnesses inside one
true segment $[p,q]$ of $\beta$ cover $\Dset$.
\end{definition}

A single continuous visit may need several witnesses (e.g., $a=b=2$, $D=1$,
$\ell=1$, $\beta=1$ on $[1,3]$); a true segment $[p,q]$ covers $\Dset$ iff
$q-p\ge\ell$, $p\le b-D$ and $q\ge a+\ell$, which gives an STL form of $\Fsi$
(supplement).

\begin{proposition}[Baseline hierarchy]\label{prop:hierarchy}
\StmtHierarchy
\end{proposition}

\begin{figure}[t]
\centering
\includegraphics[width=\columnwidth]{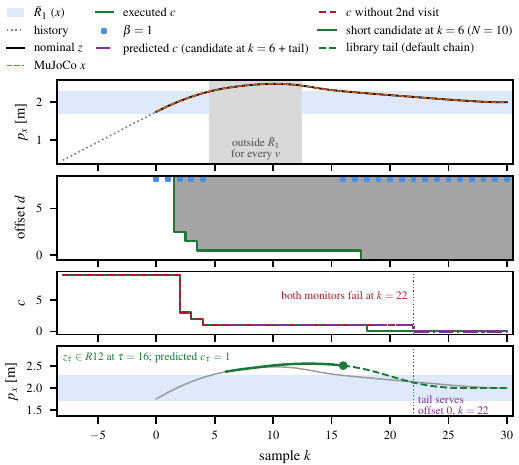}
\caption{E1 on one closed-loop MuJoCo run; $\F_{[3,20]}\G_{[0,2]}\mu_{R_1}$,
$D=8$, $H=30$. (a)~Every admissible nominal input leaves $\bar R_1$ at samples
\EOneForcedOut{} (gray). (b)~Served offsets form the suffix $[c,D]$; the first
visit serves $[1,D]$, the second $\{0\}$. (c)~Without the second visit both
monitors fail at $k=\EOneCFFail$. (d)~The candidate applied at
$k=\EOneTailK$ ($N=10$) ends at $\tau=\EOneTailTau$ with offset~0 unserved;
the certified library tail serves it at $k=\EOneTailClear$ (selection rule in
the supplement).}
\label{fig:coverage}\label{fig:mechanism}
\end{figure}

\begin{proposition}[When exact coverage adds signals]\label{prop:strict}
\StmtStrict
\end{proposition}
\ifshowproofsketches
\begin{proofsketch}
Witnesses from distinct maximal true segments are at least $\ell+2$ apart. The
intervals \eqref{eq:I} of two starts leave no gap only if the starts differ by
at most $\Delta+1$, so for $\Delta\le\ell$ a single segment must cover $\Dset$.
For $D\ge1$ and $\Delta\ge\ell+1$, the segments $[a-D,a-1+\ell]$ and
$[b,b+\ell]$ cover $[1,D]$ and $\{0\}$, respectively, and neither covers
$\Dset$ alone (Fig.~\ref{fig:coverage}).
\end{proofsketch}
\fi

The signals that $\Fex$ adds to $\Fsi$ are those served only by separated
visits; with dynamics they matter only if every single witness and continuous
visit is excluded, and for $\Delta\le\ell$ the single-visit encoding is exact.

\section{Completion Certificates}\label{sec:certs}

A physically invariant set does not certify a deadline, so the terminal
object must depend on absolute time and on the completion state.

\subsection{Exact completion family}
Let $\bar S_k=\bigcap\{\bar S_i:k\in[\alpha_i-D,\omega_i]\}$, with $\bar S_k=\R^p$
if no safety task is active at $k$; by Lemma~\ref{lem:L1}(b) these are exactly
the times at which safety must hold for some offset. Define
\[
\Adm_k(z,q)\iff z\in\mathcal Z,\ Cz\in\bar S_k,\ q\in\Qset_k .
\]

\begin{assumption}[Post-mission set]\label{ass:post}
$\Zpost\subseteq\mathcal Z$ is control invariant for the nominal dynamics with
inputs in $\mathcal V$, and a controller rendering it invariant is available.
\end{assumption}

Entry into $\Zpost$ is a design restriction imposed on all baselines.

\begin{definition}[Exact completion family]\label{def:exact-family}
Let $\Tset_H(q)=\{z\in\Zpost:Cz\in\bar S_H\}$ if $\Acc(q)$ and
$\Tset_H(q)=\varnothing$ otherwise. For $k<H$,
\[
\begin{aligned}
\Tset_k(q)=\big\{z:{}&\Adm_k(z,q),\ \exists v\in\mathcal V:\ z^+=Az+Bv,\\
&z^+\in\Tset_{k+1}\big(\Gamma_{k+1}(q,\chi(Cz^+))\big)\big\}.
\end{aligned}
\]
\end{definition}

This recursion, in the spirit of~\cite{yu2022decomp,vlahakis2024recurring},
defines the problem and is not a contribution.

Each $\Tset_k$ is monotone: $q\preceq q'$ implies
$\Tset_k(q)\subseteq\Tset_k(q')$ (supplement).

\subsection{Verified transition library}\label{sec:library}
\begin{assumption}[Convex tightened sets]\label{ass:convex}
Each $\bar R_j$ is a convex polytope. Each $\bar S_i$ is a finite union of
closed convex polyhedra, called cells. The sets $\mathcal Z$, $\mathcal V$ and
$\Zpost$ are convex polytopes.
\end{assumption}

\begin{definition}[Verified transition library]\label{def:library}
A library $\Lib$ consists of finitely many regions $P_\rho$, $\rho\in\mathcal R$,
each a convex polytope, and edges $e=(\rho_e,\rho'_e,F_e,g_e,\sigma_e)$, each
with a source region, a successor region, an affine law $v=F_ez+g_e$ and a label
$\sigma_e\in\{0,1\}^{n_r}$. The library is \emph{verified} if every edge
satisfies, on the whole source region $P_{\rho_e}$:
\begin{itemize}
\item[(V1)] $F_ez+g_e\in\mathcal V$ for all $z\in P_{\rho_e}$;
\item[(V2)] $(A+BF_e)P_{\rho_e}+Bg_e\subseteq P_{\rho'_e}$;
\item[(V3)] $\sigma_{e,j}=1$ only if
$C\big((A+BF_e)P_{\rho_e}+Bg_e\big)\subseteq\bar R_j$.
\end{itemize}
Region $\rho$ is admissible at $k$, written $\RA_k(\rho)$, if
$P_\rho\subseteq\mathcal Z$ and, for every safety task active at $k$, $CP_\rho$
lies in a single cell of $\bar S_i$.
\end{definition}

Targets are convex, so (V1)--(V3) are checked at the vertices of $P_\rho$ or
by linear programming (nonconvex safe sets never through vertices). An edge
certifies its entire source region.

\subsection{Plain and ordered backward recursion}
On the formal domain $\Qset\cup\{\bot\}$ (the full chain product, not only the
reachable states) define $\Win_k(\rho,q)\in\{0,1\}$ by
\begin{align}
\Win_H(\rho,q)&=[\Acc(q)\wedge\RA_H(\rho)\wedge P_\rho\subseteq\Zpost],
\label{eq:win-H}\\
\Win_k(\rho,q)&=[q\in\Qset_k\wedge\RA_k(\rho)]\notag\\
&\quad\wedge\bigvee_{e:\rho_e=\rho}\Win_{k+1}\big(\rho'_e,\Gamma_{k+1}(q,\sigma_e)\big).
\label{eq:win-k}
\end{align}
The plain algorithm evaluates \eqref{eq:win-H}--\eqref{eq:win-k} at every
$(k,\rho,q)$, i.e., $O(H\,|\Lib|\,|\Qset|)$ monitor updates. The ordered one
stores only $M_k(\rho)=\min\{q:\Win_k(\rho,q)=1\}$: $M_H(\rho)=\{((0,0))_j\}$
if $\RA_H(\rho)$ and $P_\rho\subseteq\Zpost$, and, if $\RA_k(\rho)$
($\varnothing$ otherwise),
\begin{equation}\label{eq:ordered}
M_k(\rho)=\min{\textstyle\bigcup}\{\min\Pre_k(g,\sigma_e):\rho_e=\rho,\ g\in M_{k+1}(\rho'_e)\},
\end{equation}
with the closed-form predecessors below.

\begin{lemma}[Minimal predecessors]\label{lem:minpre}
\StmtMinPre
\end{lemma}

The bound $2^{n_r}$ is per generator and edge, not a bound on $|M_k(\rho)|$.

\begin{theorem}[Plain and ordered recursions coincide]\label{thm:dp-eq}
\StmtDPEq
\end{theorem}
\ifshowproofsketches
\begin{IEEEproof}
Backward induction on $k$. At $H$, $\Acc(q)$ holds iff
$q\succeq((0,0))_j$. For $k<H$, by \eqref{eq:win-k} and the induction
hypothesis, $\Win_k(\rho,q)=1$ iff $\RA_k(\rho)$ holds and
$q\in\Pre_k(g,\sigma_e)$ for some edge $e$ from $\rho$ and some
$g\in M_{k+1}(\rho'_e)$; the case $\Gamma_{k+1}(q,\sigma_e)=\bot$ is excluded
because $\bot$ dominates no $g$. Each $\Pre_k(g,\sigma_e)$ is upward closed in
the finite poset $\Qset$ (Lemma~\ref{lem:minpre}), so $q$ belongs to it iff $q$
dominates one of its minimal elements. Taking minima of the union preserves
this property.
\end{IEEEproof}
\fi

\begin{theorem}[Soundness of the library family]\label{thm:sound}
\StmtSound
\end{theorem}
\ifshowproofsketches
\begin{proofsketch}
(a) follows from Theorem~\ref{thm:dp-eq}. For (b), choose an edge attaining the
disjunction in \eqref{eq:win-k}. (V1)--(V3) give $v\in\mathcal V$,
$z^+\in P_{\rho'_e}$ and $\sigma_e\le\chi(Cz^+)$, and Lemma~\ref{lem:mono}
together with (a) lifts $\Win_{k+1}(\rho'_e,\Gamma_{k+1}(q,\sigma_e))=1$ to
every $\beta\ge\sigma_e$. (c) is \eqref{eq:win-H}. (d) follows by backward
induction from (b) and (c) with $\beta=\chi(Cz^+)$.
\end{proofsketch}
\fi

Ordering loses nothing relative to the plain recursion on the \emph{same}
library, but says nothing about the library's loss relative to $\Tset_k$;
antichains~\cite{dewulf2006antichains,cadilhac2022acacia} in $2n_r$ chains
can be large, so no worst-case gain is claimed.

\subsection{Branching libraries for disturbed systems}\label{sec:branching}
For \eqref{eq:nlsys}, one state and input can reach different regions that
need different continuations, which a single-successor library cannot express.

\begin{definition}[Verified branching library]\label{def:blib}
A branching library $\Blib$ has finitely many regions $P_\rho$ and edges
$e\in A(\rho_e)$, each with a feedback law $\kappa_e$ defined on the whole
source region and a nonempty finite set
$\Succ(e)=\{(\rho_b,\sigma^b)\}_{b=1}^{B_e}$ of successor regions with
labels. It is \emph{verified} if, for every edge,
(B1) $\kappa_e(x)\in U(x)$ for all $x\in P_{\rho_e}$, and
(B2) for all $x\in P_{\rho_e}$ and all $w\in\mathcal W(x,\kappa_e(x))$ there is
a branch $b$ with $x^+=f(x,\kappa_e(x),w)\in P_{\rho_b}$ and
$\sigma^b\le\chi(h(x^+))$. $\RA_k(\rho)$ requires $P_\rho$ to satisfy the
state and safety constraints active at $k$.
\end{definition}

Branches are matched only after $x^+$ is observed. Our verifier bisects the
source region and $\mathcal W$ jointly in rational interval arithmetic and
adds one branch per grid region that a sub-enclosure meets, labelled on the
intersection.

On $\Qset\cup\{\bot\}$ the robust recursion is an AND/OR game,
\begin{align}
\Win_H(\rho,q)&=[\Acc(q)\wedge\RA_H(\rho)\wedge\rho\in\Kpost],
\label{eq:rwin-H}\\
\Win_k(\rho,q)&=[q\in\Qset_k\wedge\RA_k(\rho)]\notag\\
&\quad\wedge\bigvee_{e\in A(\rho)}\ \bigwedge_{b\in\Succ(e)}
\Win_{k+1}\big(\rho_b,\Gamma_{k+1}(q,\sigma^b)\big).
\label{eq:rwin-k}
\end{align}
The recursion is standard; what is specific is that its predecessors stay
closed form on the compact state.

\begin{proposition}[Robust minimal predecessors]\label{prop:robpre}
\StmtRobPre
\end{proposition}

For each fixed target vector, the candidate-count upper bound decreases from
$2^{n_rB}$ to $2^{n_r}$. The $\prod_b
|M_{k+1}(\rho_b)|$ target vectors remain, and branches sharing a label are
joined first.

Let $\iota(q)=((\{c_j,\dots,D\},r_j))_j$ map compact states to explicit
(served set, run length) states, and let $\Gamma^{\mathrm{bit}}_k$ add the
offsets served by a witness completing at $k$ and fail if an offset is
expired. Then $\Gamma^{\mathrm{bit}}_k(\iota(q),\beta)$ fails iff
$\Gamma_k(q,\beta)=\bot$ and otherwise equals $\iota(\Gamma_k(q,\beta))$
(supplement).

\begin{theorem}[Robust certificates]\label{thm:robcert}
\StmtRobCert
\end{theorem}

\begin{theorem}[Branch-feedback closed loop]\label{thm:robloop}
\StmtRobLoop
\end{theorem}

Adding verified multi-successor edges to a library on the same regions never
shrinks the certified set (supplement).

\emph{Post-mission set.} With bidirectional position disturbances, no single
position box is robustly invariant for the unicycle below (supplement). We
therefore take the terminal set as a safety
fixed point on the verified graph: starting from the regions inside a
declared static safe set $Z_{\rm post}$,
\begin{equation}\label{eq:kpost}
\mathcal K^{i+1}=\{\rho\in\mathcal K^i:\exists e\in A(\rho)\ \forall
b\in\Succ(e):\ \rho_b\in\mathcal K^i\},
\end{equation}
until a fixed point $\Kpost$, with one witness edge $e_{\rm post}(\rho)$ per
retained region.

$\Kpost$ is the greatest such fixed point, and applying $e_{\rm post}$ in
the current region keeps the state in $\Kpost$ under every admissible
disturbance sequence (supplement).

\emph{Instance.} The prototype is the declared discrete unicycle
\begin{equation}\label{eq:unicycle}
\begin{aligned}
p_{k+1}&=p_k+hv_k(\cos\theta_k,\sin\theta_k)+(w_{x,k},w_{y,k}),\\
\theta_{k+1}&=\theta_k+h\omega_k+w_{\theta,k},
\end{aligned}
\end{equation}
with box disturbances and primitives $v$ constant,
$\omega=k_\theta(\theta_{\rm ref}-\theta)$; $\theta$ ranges over a bounded
interval that does not wrap.

\section{Receding-Horizon Controller and Guarantees}\label{sec:mpc}

\subsection{Online problem}
Let $N,H\ge1$ and $N_k=\min\{N,H-k\}$. At $k<H$, with executed nominal state
$z_k$ and monitor state $q_k$, the controller solves $\mathcal P_k(z_k,q_k)$:
\begin{equation}\label{eq:ocp}
\begin{aligned}
\min\ &\textstyle\sum_{i=0}^{N_k-1}L(z_{i|k},v_{i|k})\\
\text{s.t. }\ & z_{0|k}=z_k,\quad q_{0|k}=q_k,\\
& z_{i+1|k}=Az_{i|k}+Bv_{i|k},\\
& \hat\beta_{i+1|k}\in\{0,1\}^{n_r},\\
& \hat\beta_{i+1|k,j}=1\Rightarrow Cz_{i+1|k}\in\bar R_j,\\
& q_{i+1|k}=\Gamma_{k+i+1}(q_{i|k},\hat\beta_{i+1|k}),\\
& v_{i|k}\in\mathcal V,\ \ \Adm_{k+i}(z_{i|k},q_{i|k}),
  \quad i\in[0,N_k-1],\\
& z_{N_k|k}\in\That_{k+N_k}(q_{N_k|k}).
\end{aligned}
\end{equation}
Membership $z\in\That_k(q)$ is a disjunction over regions and generators of
linear conditions under binary selectors. The prediction need not complete the
task; the terminal certificate carries the remaining obligations. $L$ is
arbitrary, and no stability property is claimed.

\begin{assumption}[One-sided encoding]\label{ass:onesided}
(a) Every candidate used by the controller satisfies all constraints of
\eqref{eq:ocp}, including the implications
$\hat\beta_{i|k,j}=1\Rightarrow Cz_{i|k}\in\bar R_j$; this is checked after the
solve. (b) The sets $\bar R_j$ in \eqref{eq:ocp} are the ones used for the
labels in (V3) and by the executed monitor, including on the recorded history.
\end{assumption}

The converse implication is not assumed. Big-M rows
$G_{jh}Cz\le f_{jh}+M_{jh}(1-\hat\beta_j)$ for $\bar R_j=\{y:G_jy\le f_j\}$
must be redundant on $\mathcal Z$ when $\hat\beta_j=0$ (constants in the
supplement). Solver tolerances are not certificates, hence the check in
Assumption~\ref{ass:onesided}(a); a solver bit $0$ where the monitor records
$1$ is absorbed by monotonicity (Theorem~\ref{thm:rf}).

\emph{Monitor encoding.} Per task and step, $\Gamma$ uses integers $c',r'$,
binaries for $r=\ell+1$, $r'=\ell+1$, $c\le\hi+1$ and $c\ge\lo+1$, and two
products ($r'=\hat\beta\min\{r+1,\ell+1\}$; $c'=\min\{c,\lo\}$ exactly when a
dwell completes at a start adjacent to the served suffix); the hard row
$c'\le\cbar_{j,t}$ excludes $\bot$. With constants from $c\in[0,D+1]$,
$r\in[0,\ell+1]$, this takes at most \EncMonVars{} variables and
\EncMonRows{} rows per task and step, independent of $D$ and $\ell$; each
terminal option $(\rho,m)$ adds one selector. In E2 ($N=10$), $\mathcal P_0$
has \EncTenVars{} variables and \EncTenRows{} rows, of which
\EncTenMonVars/\EncTenMonRows{} belong to the monitor (all rows and solver
settings in the supplement).

\begin{algorithm}[t]
\caption{Closed loop}\label{alg:closedloop}
\begin{algorithmic}[1]
\Require $N\ge1$, $H\ge1$; verified library and $M_k(\rho)$; $\mathcal E$, $K$
  (Assumption~\ref{ass:tube}); recorded history $\ybar_{-D},\dots,\ybar_{-1}$
\State choose $z_0$ with $x_0-z_0\in\mathcal E$; run the monitor with exact
  bits on $\ybar_{-D},\dots,\ybar_{-1},Cz_0$ to obtain $q_0$
\State by the first input deadline, have a candidate for
  $\mathcal P_0(z_0,q_0)$, for the actual nominal state, monitor state,
  recorded history and task clock, verified as in
  Assumption~\ref{ass:onesided}(a); store it
\For{$k=0,\dots,H-1$}
  \State attempt a new candidate for $\mathcal P_k(z_k,q_k)$; select it if it
    passes the check of Assumption~\ref{ass:onesided}(a), else select the
    stored candidate
  \State apply $u_k=v_{0|k}+K(x_k-z_k)$
  \State $z_{k+1}\gets Az_k+Bv_{0|k}$ \Comment{never reset to $x_{k+1}$}
  \State $q_{k+1}\gets\Gamma_{k+1}(q_k,\chi(Cz_{k+1}))$ \Comment{exact bits}
  \If{$k<H-1$}
    \State shift the selected candidate, recompute its monitor states from
      $q_{k+1}$, append a library edge if $k+N_k<H$ (Theorem~\ref{thm:rf}),
      and store the result
  \EndIf
\EndFor
\State switch to the post-mission controller (Assumption~\ref{ass:post}),
  keeping the feedback $K(x_k-z_k)$
\end{algorithmic}
\end{algorithm}

\subsection{Guarantees}
\begin{theorem}[Recursive feasibility]\label{thm:rf}
\StmtRF
\end{theorem}
\ifshowproofsketches
\begin{IEEEproof}
Write $n=N_k$. By Assumption~\ref{ass:onesided}(a),
$\hat\beta^*_{1|k}\le\chi(Cz^*_{1|k})=\chi(Cz_{k+1})$, so Lemma~\ref{lem:mono}
gives $q^*_{1|k}=\Gamma_{k+1}(q_k,\hat\beta^*_{1|k})\preceq q_{k+1}$. For the
overlapping steps, set $v_{i|k+1}=v^*_{i+1|k}$ and
$\hat\beta_{i|k+1}=\hat\beta^*_{i+1|k}$. Then $z_{i|k+1}=z^*_{i+1|k}$, and by
induction with Lemma~\ref{lem:mono}, $q_{i|k+1}\succeq q^*_{i+1|k}$. The input,
state, safety and one-sided constraints are inherited because absolute times
coincide, and $q_{i|k+1}\in\Qset_{k+1+i}$ because $\Qset_{k+1+i}$ is upward
closed. If $k+n=H$, then $N_{k+1}=n-1\ge1$, and
$z^*_{n|k}\in\That_H(q^*_{n|k})\subseteq\That_H(q_{n-1|k+1})$ by
Theorem~\ref{thm:sound}(a). If $k+n<H$, then $N_{k+1}=n$, and
Theorem~\ref{thm:sound}(a) gives $z^*_{n|k}\in P_\rho$ with
$\Win_{k+n}(\rho,q_{n-1|k+1})=1$. Theorem~\ref{thm:sound}(b) then supplies an
edge $e$. Appending $v_{n-1|k+1}=F_ez^*_{n|k}+g_e$ and
$\hat\beta_{n|k+1}=\sigma_e$ satisfies $\Adm_{k+n}$, the input constraint and
the one-sided implication, and the terminal constraint holds with
$\beta=\sigma_e$.
\end{IEEEproof}
\fi

\begin{theorem}[Closed-loop guarantees]\label{thm:closedloop}
\StmtClosedLoop
\end{theorem}
\ifshowproofsketches
\begin{proofsketch}
(i) follows by induction from the verified initial candidate: if the candidate
selected at $k\le H-2$ is feasible, Theorem~\ref{thm:rf} makes the stored
shifted candidate feasible for $\mathcal P_{k+1}(z_{k+1},q_{k+1})$. (ii) is the
tube argument: $e_k\in\mathcal E$, so
$x_k\in\mathcal Z\oplus\mathcal E\subseteq\mathcal X$ and
$u_k\in\mathcal V\oplus K\mathcal E\subseteq\mathcal U$. (iii) The candidate
selected at $H-1$ gives $z_H\in\That_H(q^*_{1|H-1})\neq\varnothing$, so
$\Acc(q^*_{1|H-1})$ holds by Theorem~\ref{thm:sound}(c); since
$q^*_{1|H-1}\preceq q_H$, $\Acc(q_H)$ follows. (iv) $q_H$ is the compact
monitor of the executed nominal bits, so Corollary~\ref{cor:compact} and
Lemma~\ref{lem:L1}(a) give every reach task under every offset. $\Adm_k$ for
$k<H$, $\RA_H$ at $H$ and Assumption~\ref{ass:history} give safety through
Lemma~\ref{lem:L1}(b), and Lemma~\ref{lem:spatial} transfers both to
$\ytil^{d,\eta}$.
\end{proofsketch}
\fi

\emph{Implementation boundaries.} Theorem~\ref{thm:closedloop} assumes a
verified initial candidate for the actual $z_0$, $q_0$, history and task
clock, which computation never restarts; Theorem~\ref{thm:rf} guarantees a
backup, not solver success, and no cost decrease. An early attempt is used
only if $z_k$, $q_k$ and the executed inputs match exactly. A quadrotor
needs a certified inner-loop error in $\mathcal E$, which empirical errors
are not.

\subsection{Completion filter on a branching certificate}\label{sec:filter}
On a verified branching library, the certificate itself is an executable
feedback policy (Theorem~\ref{thm:robloop}), and a performance controller can
propose actions. At $(k,\rho,q)$ with $\Win_k(\rho,q)=1$, the admissible set
is
\begin{multline*}
A^{\rm comp}_k(\rho,q)=\{e\in A(\rho):\\
\Win_{k+1}(\rho_b,\Gamma_{k+1}(q,\sigma^b))=1\ \forall b\in\Succ(e)\},
\end{multline*}
evaluated by enumerating every outgoing edge. A continuous proposal is
projected to the nearest library primitive, executed if its edge lies in
$A^{\rm comp}_k$, and otherwise replaced by the nearest primitive that does;
permissiveness is relative to the library. Every member of $A^{\rm comp}_k$
preserves the induction of Theorem~\ref{thm:robloop}, so the guarantees hold
for any proposer. All steps share one wall-clock budget per tick, after which
the witness edge runs; a late proposal is adopted only if its tick, region
and monitor state are current.

\section{Evaluation}\label{sec:eval}

\emph{Platform and libraries.} MuJoCo~\cite{todorov2012mujoco} simulates a
Crazyflie model~\cite{menageriePinned} (modified actuators, not a hardware
calibration) at 500\,Hz under an attitude and height loop; the outer model is
a planar double integrator per axis ($h=0.2$\,s) with tube gain $K$ and
bounded wind. Regions are translates $c\oplus T$ of one tube shape with laws
$v=g+K(z-c)$ (E1--E3) or grid cells with fixed feedback primitives (R1--R6).
Every edge is verified in exact arithmetic on its whole source region, for
all of $\mathcal W$ where disturbed; HiGHS candidates are used only after
exact validation, and every run is reported. On \IfRuns{} untuned runs the
tube interface held at every sample (finite empirical evidence). A change of
model, $\mathcal W$, obstacles or primitives requires re-verification; with
the library unchanged, a new window or $D$ needs only a new recursion, new
targets new labels, and a new terminal set or controller its terminal
conditions.

\emph{E1: mechanism.} In $\F_{[3,20]}\G_{[0,2]}\mu_{R_1}$ with $D=8$,
$H=30$, the vehicle enters $R_1$ at 0.8\,m/s. An exact lower bound over all
admissible nominal inputs places it outside $\bar R_1$ at samples
\EOneForcedOut{} (Fig.~\ref{fig:mechanism}a), excluding every single witness
and continuous visit. All \EOneRuns{} runs complete with the interface
intact; on nominal, measured and counterfactual histories both monitors agree
at every sample (Fig.~\ref{fig:mechanism}b), and without the second visit
both fail at $k=\EOneCFFail$ (Fig.~\ref{fig:mechanism}c).

\emph{E2 and E3 (supplement).} An exact full-horizon reference verifies
\ETwoFeasible{} of \ETwoQueries{} test queries; library-certified horizons
$N=5,10,30$ verify \ETwoNFiveBefore, \ETwoNTenBefore{} and
\ETwoNThirtyBefore\,\% within 2\,s. A full-horizon STL encoding verifies
\ETwoSTLBefore\,\% and plans fastest; it drops the converse half of every
equivalence of~\cite{raman2014mpc}, whose two-sided form, in our
implementation, verifies \ExtRamVer/\ExtDen{} at median \ExtRamMs\,ms. In a branching scene
($H=107$), re-solving lowers the revealed cost relative to the verified backup
alone by \EThreeNoSolveOurs{} and \EThreeShNoSolveOurs{} on two geometries,
and the backup carries an optimizer outage of \EThreeOutageTicks{} samples;
against full-horizon replanning the sign depends on the execution window and
geometry.

\subsection{R1--R6: robust completion}\label{sec:eval-robust}
\emph{Setup.} The system is \eqref{eq:unicycle} with $h=1$, $|w_x|,|w_y|\le0.1$\,m
and $|w_\theta|\le0.05$\,rad, a grid of $0.5$\,m by $1/4$\,rad cells and 28
primitives (four speeds, seven $\theta_{\rm ref}$, deadbeat heading feedback).
R1--R3 use an $8\times5$\,m workspace, one obstacle, a recorded history,
$\F_{[\ROneA,\ROneB]}\G_{[0,\ROneEll]}\mu_{R}$, $D=\ROneD$ and $H=\ROneH$.

\emph{R1: prototype gates} (\ROneAllPass{} passed). (1) \ROneEdges{} edges
are verified on whole cells for all of $\mathcal W$; $\Kpost$ keeps
\ROneKpost{} of \ROneKpostCand{} regions. (2) $\Gamma$ on the actual history
gives $q_0=\ROneQzero$; the start is certified and
\ROneCompleted/\ROneRollouts{} sampled rollouts complete in $\Kpost$. (3) At
$k=\ROneEvidenceK$, two admissible disturbances send the whole start
enclosure under \ROneEvidencePrim{} strictly inside different cells, whose
complete feasible sets $A_1=\ROneAOne$ and $A_2=\ROneATwo$ are disjoint
(figure in the supplement). (4) Explicit-bitset, compact and ordered
recursions agree at every $(k,\rho,q)$. \emph{R2} (supplement): the
single-successor sub-library certifies \RTwoSingleCertified{} regions
against \RTwoFullCertified{}, and as $2\bar wa\ge|\bar R|$ no product-box
partition does better.

\emph{R3: filter in closed loop.} A shared task-aware CEM proposer feeds a
robust filter, a nominal certificate ($w=0$) and a robust safety-only
filter, with seeded disturbances and a \RThreeBudgetMs\,ms budget
(\RThreeOverruns{} overruns). In all
\RThreeConfigs{} frozen configurations every arm completes; the nominal
certificate is lost in \RThreeNominalLost, and the robust filter intervenes
most (\RThreeRobustInterv{} against \RThreeNominalInterv{} and
\RThreeSafetyInterv). In four preregistered cases (supplement), with a
proposer pursuing a distant safe waypoint only the robust arm completes, and
through an optimizer outage it completes on certified fallbacks.

\emph{R4: quadrotor.} On the platform's declared outer model (1\,s ticks,
$\tfrac18$\,m cells), \RFourEdges{} edges are verified on whole cells,
inputs and intermediate samples included. From a
measured history, \RFourCompleted/\RFourSeeds{} MuJoCo wind seeds
start certified, complete safely, stay in $\Kpost$ and keep every residual in
$\mathcal W$, with identical branches. The outer model admits
\RPairWit/\RPairCands{} candidate pairs with disjoint certified next-action
sets; in MuJoCo these reach different regions with nested sets.

\begin{figure}[t]
\centering
\includegraphics[width=\columnwidth]{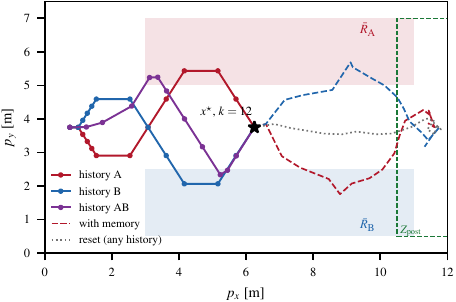}
\caption{R5: three legal histories end exactly at $x^\star$ at $k=\RFiveKstar$;
dashed: continuations with memory after A (red) and B (blue), dotted: the
all-monitor-memory reset controller (identical for every history), one seed.}
\label{fig:r5}
\end{figure}

\emph{R5: same state, different histories.} On a mirror-symmetric
two-target version of the scene, three legal prefixes end exactly at one
state $x^\star$ at $k=\RFiveKstar$ with equal bits and dwell counters; each
last disturbance is proved in $\mathcal W$ by interval arithmetic. After
visiting only A, the
complete certified set at $x^\star$ is \RFiveSetA{} $(v,\theta_{\rm ref})$;
after only B it is \RFiveSetB, disjoint from it; after both, all
\RFiveNprims{} primitives. One common proposal
\RFiveProposal{} is therefore replaced by \RFiveDecA{}, by \RFiveDecB{} or
kept (Fig.~\ref{fig:r5}). Resetting all monitor memory at $x^\star$ (clock
kept) leaves no certified action. With memory, \RFiveMemCompleted/\RFiveMemRuns{} seeded
continuations, certified from $k=\RFiveKstar$ (the legal prefixes are not
certified throughout), complete; the reset controller, judged by an unreset monitor,
completes \RFiveResetCompA/\RFiveSeeds{} after A and
\RFiveResetCompB/\RFiveSeeds{} after B (for this proposer and fallback only).
Partial coverage alone also matters: on the R1 library, two legal prefixes
end at one state at $k=\RFivePKstar$ with served offsets \RFivePServedP{}
and \RFivePServedQ{}, and certified sets \RFivePSetP{} $\subsetneq$
\RFivePSetQ{} (supplement).

\emph{R6: what do the compact and ordered states cost?} On an obstacle-free
four-target scene ($n\le4$ tasks, $D\le16$), all methods answer the same
matched sparse queries (all regions at $k=0$) and agree; bitset and compact
states are equally many in every
setting (Theorem~\ref{thm:suffix}), with no clear resource gain; the main
difference is in whole-domain tables, where the full bitset table exceeds the
time or memory budget in \RSixFullBitsetOver{} of
\RSixSettings{} settings. The ordered recursion covers the whole domain; it
is \RSixRatioMin--\RSixRatioMax{} times slower than sparse compact, at best
draws level over \RSixBatches{} online batches and saves memory only at
$(n,D)=(3,4)$ and with four tasks.

\section{Proofs for Section~\ref{sec:problem}}

\begin{restated}{Lemma}{lem:spatial}
\StmtSpatial
\end{restated}
\begin{IEEEproof}
For Boolean signals $\xi,\xi'$ on a common domain write $\xi\le\xi'$ if every
component satisfies $\xi_p(t)\le\xi'_p(t)$ at every $t$. A negation-free
Boolean STL formula $\psi$ is monotone: $\xi\le\xi'$ and $(\xi,t)\models\psi$
imply $(\xi',t)\models\psi$. This follows by induction on $\psi$, because
atomic propositions, conjunction, disjunction, $\G_I$ and $\F_I$ all preserve
$\le$.

Let $\xi=(\gamma,\beta)(\cdot-d)$, and let $\xi'$ collect the truth values of
the original predicates along $\ytil^{d,\eta}$, i.e., $\xi'_{S_i}(t)=1$ iff
$\ytil^{d,\eta}(t)\in S_i$ and $\xi'_{R_j}(t)=1$ iff $\ytil^{d,\eta}(t)\in R_j$,
for $t\in[0,H]$. If $\beta_j(t-d)=1$, then $\ybar_{t-d}\in\bar R_j$, and since
$\eta(t)\in\Ball$,
\[
\ytil^{d,\eta}(t)=\ybar_{t-d}+\eta(t)\in\bar R_j\oplus\Ball\subseteq R_j ,
\]
so $\xi'_{R_j}(t)=1$. The same argument applies to $\gamma_i$ and $S_i$. Hence
$\xi\le\xi'$ on $[0,H]$. Since $(\xi,0)\models\varphi_\beta$ and $\varphi$ is
negation-free with the same structure, $(\ytil^{d,\eta},0)\models\varphi$.
\end{IEEEproof}

\emph{Admissible tightenings.} The Pontryagin difference $R_j\ominus\Ball$ is
the largest choice of $\bar R_j$; for $\{y:Gy\le f\}$ it is
$\{y:Gy\le f-\varepsilon|G|\mathbf 1\}$, and for a predicate $h_j\ge0$ with
$\ell_\infty$-Lipschitz constant $L_j$ the set $\{y:h_j(y)\ge L_j\varepsilon\}$
is admissible.

\emph{Obstacles.} With $S_i=\{y:\dist(y,\mathcal O_i)\ge\delta_i\}$ and
$\bar S_i\subseteq\{y:\dist(y,\mathcal O_i)\ge\delta_i+\varepsilon\}$, both
sets are closed and exclude contact with $\mathcal O_i$, and
$\dist(y+\eta,\mathcal O_i)\ge\dist(y,\mathcal O_i)-\|\eta\|_\infty$ gives
$\bar S_i\oplus\Ball\subseteq S_i$. For a polytopic obstacle
$\{y:A_iy\le b_i\}$ with rows $a_{ih}\neq0$, the half-spaces
$\{y:a_{ih}y\ge b_{ih}+(\delta_i+\varepsilon)\|a_{ih}\|_1\}$ keep this
clearance, so their union is an admissible $\bar S_i$ (conservative near
vertices). Nonconvex sets are never checked through vertices.

\emph{Quantifier order for disturbed systems.} Two mistakes must be avoided
in the objective of Section~\ref{sec:nonlinear}. For $x^+=u+w$,
$u,w\in\{-1,1\}$ and goal $x^+=0$, every known $w$ admits $u=-w$ but no $u$
works for both, so $\forall w\,\exists u$ is unsound. For $x_1=w_0\in\{-1,1\}$,
$x_2=x_1+u_1$, no fixed $u_1$ reaches $0$ while $u_1=-x_1$ does, so the
future must be feedback.

\section{Proofs for Section~\ref{sec:state}}

Throughout, a single reach task has parameters $(a,b,\ell)$, window length
$\Delta=b-a$, and bit signal $\beta$ on $[-D,H]$ with $H\ge b+\ell$. Witnesses
start at $s\ge-D$ by definition.

\subsection{Coverage semantics and compilation}

\begin{restated}{Lemma}{lem:L1}
\StmtCoverage
\end{restated}
\begin{IEEEproof}
(a, $\Rightarrow$) Fix $d\in\Dset$ and $t_d\in[a,b]$ with $\beta(t_d+r-d)=1$
for all $r\in[0,\ell]$, and let $s=t_d-d$. Then $s\ge a-D\ge-D$ and
$\beta(s)=\dots=\beta(s+\ell)=1$, so $s$ is a witness. Moreover,
$a\le s+d=t_d\le b$ gives $d\in I(s)$, and $s+\ell\le b-d+\ell\le H$. Hence
$d\in\Cov(H)$.

(a, $\Leftarrow$) If $d\in\Cov(H)$, some witness $s$ has $d\in I(s)$. Put
$t=s+d\in[a,b]$. Then $\beta(t+r-d)=\beta(s+r)=1$ for all $r\in[0,\ell]$.

(b) The indices read by the left side are
$\{t-d:t\in[\alpha,\omega],\ d\in[0,D]\}=[\alpha-D,\omega]$, which lies in
$[-D,H]$ because $\alpha\ge0$ and $\omega\le H$.
\end{IEEEproof}

\begin{restated}{Proposition}{prop:compile}
\StmtCompile
\end{restated}
\begin{IEEEproof}
Boolean STL formulas are generated by
$\psi::=p\mid\neg\psi\mid\psi\wedge\psi\mid\G_I\psi\mid\F_I\psi
\mid\psi\UntilOp_I\psi$ with
$(\xi,t)\models p$ iff $\xi_p(t)=1$, $(\xi,t)\models\G_I\psi$ iff
$(\xi,t')\models\psi$ for all $t'\in t+I$, dually for $\F_I$, and
$(\xi,t)\models\psi_1\UntilOp_I\psi_2$ iff some $s\in t+I$ has
$(\xi,s)\models\psi_2$ and $(\xi,u)\models\psi_1$ for every integer
$u\in[t,s-1]$. Another endpoint convention for $\UntilOp_I$ changes only the
range of $u$ and is handled identically. We claim
\begin{equation}\label{eq:pf-shift}
(\xi(\cdot-d),t)\models\psi\iff(\xi,t-d)\models\psi
\end{equation}
whenever the evaluations on both sides read only defined samples. For atomic
$p$, both sides state $\xi_p(t-d)=1$. Negation and conjunction are immediate.
For $\G_I$,
\[
\begin{aligned}
(\xi(\cdot-d),t)\models\G_I\psi
&\iff\forall t'\in t+I:\ (\xi,t'-d)\models\psi\\
&\iff(\xi,t-d)\models\G_I\psi ,
\end{aligned}
\]
using the induction hypothesis and $t'-d\in(t-d)+I$; $\F_I$ is analogous.
For $\UntilOp_I$, $(\xi(\cdot-d),t)\models\psi_1\UntilOp_I\psi_2$ holds iff some
$s\in t+I$ has $(\xi(\cdot-d),s)\models\psi_2$ and
$(\xi(\cdot-d),u)\models\psi_1$ for every integer $u\in[t,s-1]$. By the
induction hypothesis, this reads $(\xi,s-d)\models\psi_2$ and
$(\xi,u-d)\models\psi_1$. Substituting $s'=s-d\in(t-d)+I$ and
$u'=u-d\in[t-d,s'-1]$ gives exactly
$(\xi,t-d)\models\psi_1\UntilOp_I\psi_2$. All evaluations are restricted to
defined samples.
Since $\psi$ read at time $0$ uses only $[0,H]$, the signal $\xi(\cdot-d)$ is
read on $[-d,H-d]\subseteq[-D,H]$. Applying \eqref{eq:pf-shift} with $t=0$ and
substituting $\tau=D-d$,
\[
\begin{aligned}
&\forall d\in[0,D]:\ (\xi,-d)\models\psi\\
&\quad\iff\forall\tau\in[0,D]:\ (\xi,-D+\tau)\models\psi\\
&\quad\iff(\xi,-D)\models\G_{[0,D]}\psi .
\end{aligned}
\]
\end{IEEEproof}

\subsection{Expiry, suffix structure and the compact monitor}

\begin{restated}{Lemma}{lem:expiry}
\StmtExpiry
\end{restated}
\begin{IEEEproof}
Let $d\notin\Cov(k)$ with $k\ge b-d+\ell$. Every witness $s$ with $d\in I(s)$
satisfies $s\le b-d$, so it is completed at $s+\ell\le b-d+\ell\le k$ and would
already be counted in $\Cov(k)$. Hence no such witness exists, and
$d\notin\Cov(k')$ for all $k'$. For $k'\ge k$ we also have $k'\ge b-d+\ell$, so
$d$ stays expired and the task stays failed. If $H\ge b+\ell$, then $d$ is
expired at $H$, so $\Cov(H)\neq\Dset$.

For $k\ge b+\ell$, every $d\in\Dset$ has $b-d+\ell\le k$, so the task is failed
at $k$ iff some $d\in\Dset$ is not in $\Cov(k)$.
\end{IEEEproof}

\begin{restated}{Theorem}{thm:suffix}
\StmtSuffix
\end{restated}
\begin{IEEEproof}
If $\Cov(k)=\varnothing$, take $c=D+1$. Otherwise let $d\in\Cov(k)$ be served
by a witness $s$ with $a\le s+d\le b$ and $s+\ell\le k$, and let $d'\in\Dset$
with $d'>d$. Since $s+d'>s+d\ge a$, the only way to have $d'\notin I(s)$ is
$s+d'>b$, which gives
\[
b-d'+\ell< s+\ell\le k .
\]
If also $d'\notin\Cov(k)$, then $d'$ is expired at $k$, which contradicts
non-failure. Hence $d\in\Cov(k)$ implies $[d,D]\subseteq\Cov(k)$, and
$\Cov(k)=[c,D]$ with $c=\min\Cov(k)$.
\end{IEEEproof}

We use the following reformulation of the failure test. For $c\in[0,D+1]$,
\begin{equation}\label{eq:pf-cbar}
c>\cbar_{k}\iff c\ge1\ \text{and}\ k\ge b-(c-1)+\ell .
\end{equation}
Indeed, if $b+\ell-k\ge D+1$, then $\cbar_k=D+1$ and both sides are false,
because the right side would require $c\ge b+\ell-k+1\ge D+2$. Otherwise
$\cbar_k=\max\{0,b+\ell-k\}$, and $c>\max\{0,b+\ell-k\}$ iff $c\ge1$ and
$c\ge b+\ell-k+1$.

\begin{restated}{Corollary}{cor:compact}
\StmtCompact
\end{restated}
\begin{IEEEproof}
Since both the joint update and failure are componentwise, with $\bot$ as soon
as one component fails, it suffices to prove the following for one task, by
induction on $k\in[-D-1,H]$: either the task is failed at $k$ and
$q_k=\bot$, or it is not failed and $q_k=(c_k,r_k)$ with $\Cov(k)=[c_k,D]$ and
$r_k=r(k)$.

\emph{Base.} At $k=-D-1$ no sample has been read, $\Cov=\varnothing=[D+1,D]$,
$r=0$, and no offset is expired because every deadline satisfies
$b-d+\ell\ge-D>-D-1$. This matches $\qinit=(D+1,0)$.

\emph{Step.} Assume the claim at $k-1$. If the task is failed at $k-1$, it is
failed at $k$ by Lemma~\ref{lem:expiry}, and $\bot$ is absorbing. Otherwise let
$q_{k-1}=(c,r)$ with $\Cov(k-1)=[c,D]$ and $r=r(k-1)$. The run length satisfies
$r(k)=\min\{\ell+1,r(k-1)+1\}$ if $\beta(k)=1$ and $r(k)=0$ otherwise, which is
$r^+$. The witness $s=k-\ell$ is completed at $k$ iff
$\beta(k-\ell)=\dots=\beta(k)=1$ with $k-\ell\ge-D$, i.e., iff $r(k)=\ell+1$.
Its interval is $I(s)=[\lo,\hi]$, which is empty if $\lo>\hi$. Hence
$\Cov(k)=[c,D]\cup[\lo,\hi]$ if $r^+=\ell+1$ and $\Cov(k)=[c,D]$ otherwise.
There are three cases.

\emph{Case A:} $r^+\le\ell$, or $\lo>\hi$. Then $\Cov(k)=[c,D]$ and $c^+=c$.

\emph{Case B:} $r^+=\ell+1$, $\lo\le\hi$ and $\hi\ge c-1$. The intervals
$[\lo,\hi]$ and $[c,D]$ overlap or are adjacent and $\hi\le D$, so
$\Cov(k)=[\min\{c,\lo\},D]=[c^+,D]$.

\emph{Case C:} $r^+=\ell+1$ and $\lo\le\hi<c-1$. Then $0\le c-1\le D$ and
$c-1\notin\Cov(k)$. Because $c-1\le D$, the inequality
$\min\{D,b-s\}<c-1$ gives $b-s<c-1$, i.e., $b-(c-1)+\ell<s+\ell=k$. So $c-1$
is expired and the task is failed at $k$. The update gives $c^+=c$, and by
\eqref{eq:pf-cbar} with $c\ge1$ and $k>b-(c-1)+\ell$, $c^+>\cbar_k$, so
$q_k=\bot$.

In Cases A and B, $\Cov(k)=[c^+,D]$. The uncovered offsets are $[0,c^+-1]$,
and among them $c^+-1$ has the earliest deadline. Hence the task is failed at
$k$ iff $c^+\ge1$ and $k\ge b-(c^+-1)+\ell$, iff $c^+>\cbar_k$ by
\eqref{eq:pf-cbar}, iff $q_k=\bot$. If not failed, $q_k=(c^+,r^+)$ has the
claimed form, and $c^+\le\cbar_k$ means $q_k\in\Qset_k$.

Every non-failed value lies in $\Qset$, and
$|\Qset|=\prod_j(D+2)(\ell_j+2)$; adding $\bot$ gives the bound.
\end{IEEEproof}

\subsection{Monotonicity}

\begin{restated}{Lemma}{lem:mono}
\StmtMono
\end{restated}
\begin{IEEEproof}
If $q=\bot$, then $\Gamma_k(q,\beta)=\bot$ and there is nothing to show.
Otherwise $q'\neq\bot$. Since the joint update is componentwise and returns
$\bot$ if any component does, it suffices to prove, for one task with
$(c,r)\preceq(c',r')$ (i.e., $c\ge c'$, $r\le r'$) and bits $\beta\le\beta'$,
that $\Gamma_k((c,r),\beta)\preceq\Gamma_k((c',r'),\beta')$, and that the right
side is $\bot$ only if the left side is.

\emph{Run length.} If $\beta=0$, then $r^+=0\le r'^+$. If $\beta=1$, then
$\beta'=1$ and $\min\{\ell+1,r+1\}\le\min\{\ell+1,r'+1\}$.

\emph{Coverage index.} Let $u(c)$ be the right side of
\eqref{eq:compact-update}, which applies when a witness completes. Then
$u(c)\le c$, and $u$ is nondecreasing: if $c\ge c'$ and the condition in
\eqref{eq:compact-update} holds for $c$, it holds for $c'$ since
$\hi\ge c-1\ge c'-1$, and then $u(c)=\min\{c,\lo\}\ge\min\{c',\lo\}=u(c')$; if
the condition fails for $c$, then $u(c)=c\ge c'\ge u(c')$. Let $\kappa$ and
$\kappa'$ indicate completion, i.e., $r^+=\ell+1$ and $r'^+=\ell+1$; by the
previous step $\kappa\le\kappa'$. If $\kappa=\kappa'=0$, then
$c^+=c\ge c'=c'^+$. If $\kappa=\kappa'=1$, then $c^+=u(c)\ge u(c')=c'^+$. If
$\kappa=0$ and $\kappa'=1$, then $c^+=c\ge c'\ge u(c')=c'^+$.

\emph{Failure.} The test $c^+>\cbar_k$ is monotone in $c^+$. Since
$c^+\ge c'^+$, if the right side fails, so does the left side. If neither
fails, $(c^+,r^+)\preceq(c'^+,r'^+)$.

Finally, $\Qset_k$ is defined by upper bounds on the $c_j$, so it is upward
closed, and $\Acc$ requires $c_j=0$, which is preserved when moving up.
\end{IEEEproof}

\subsection{Baselines}

\begin{slemma}[Single continuous visit]\label{lem:segment}
\StmtSegment
\end{slemma}
\begin{IEEEproof}
The starts $s\in[p,q-\ell]$ are witnesses. For consecutive starts, the integer
intervals $[a-s-1,b-s-1]$ and $[a-s,b-s]$ overlap or are adjacent because
$b\ge a$. Hence
\[
\bigcup_{s\in[p,q-\ell]}[a-s,b-s]=[a-q+\ell,\ b-p],
\]
and its intersection with $\Dset$ equals $\Dset$ iff $a-q+\ell\le0$ and
$b-p\ge D$, i.e., iff $q\ge a+\ell$ and $p\le b-D$.

\emph{Case $\Delta\ge D$.} If a segment $[p,q]$ covers $\Dset$, let
$s=\max\{p,a\}$. Then $s\le b-D$, because $p\le b-D$ and $a\le b-D$. Also
$s+\ell\le q$: if $s=p$ this is $q-p\ge\ell$, and if $s=a$ it is $q\ge a+\ell$.
So $\beta=1$ on $[s,s+\ell]$ with $s\in[a,b-D]$. Conversely, such an $s$ gives
the segment $[s,s+\ell]$ with $s\le b-D$ and $s+\ell\ge a+\ell$.

\emph{Case $\Delta<D$.} If $[p,q]$ covers $\Dset$, then
$p\le b-D<a\le a+\ell\le q$, so $\beta=1$ on $[b-D,a+\ell]$. Conversely, if
$\beta=1$ on $[b-D,a+\ell]$, take $p=b-D\ge-D$ and $q=a+\ell$; then
$q-p=\ell+D-\Delta>\ell$ and both conditions hold.
\end{IEEEproof}

\begin{restated}{Proposition}{prop:hierarchy}
\StmtHierarchy
\end{restated}
\begin{IEEEproof}
$\Fsi\subseteq\Fex$: the witnesses of the segment are witnesses of $\beta$, and
all of them are completed by $b+\ell$ because the signal ends there.
$\Fsh\subseteq\Fsi$: the segment $[t-D,t+\ell]$ with $t\in[a,b]$ has
$p=t-D\le b-D$, $q=t+\ell\ge a+\ell$, $q-p=D+\ell\ge\ell$ and $p\ge-D$, so
Lemma~\ref{lem:segment} applies. $\Fsk\subseteq\Fsi$: the segment $[s,s+\ell]$
with $s\in[a,b-D]$ satisfies both conditions. If $\Delta\ge D$,
Lemma~\ref{lem:segment} gives $\Fsi=\Fsk$. If $\Delta<D$, then $[a,b-D]$ is
empty and $\Fsk=\varnothing$.
\end{IEEEproof}

\begin{restated}{Proposition}{prop:strict}
\StmtStrict
\end{restated}
\begin{IEEEproof}
($\Leftarrow$) Let $D\ge1$ and $\Delta\ge\ell+1$, and let $\beta=1$ exactly on
$[a-D,a-1+\ell]\cup[b,b+\ell]\subseteq[-D,b+\ell]$. Since
$b\ge a+\ell+1$, the time $a+\ell$ separates the two segments. The first
segment has $\ell+D\ge\ell+1$ samples and witnesses $s\in[a-D,a-1]$; by the
first part of Lemma~\ref{lem:segment} its coverage is
$[a-(a-1),\,b-(a-D)]\cap\Dset=[1,\Delta+D]\cap\Dset=[1,D]$. The second segment
has the witness $s=b$ with $I(b)=[a-b,0]\cap\Dset=\{0\}$. Hence
$\beta\in\Fex$. The first segment violates $q\ge a+\ell$, and the second
violates $p\le b-D$ because $D\ge1$. Both conditions only become harder on
sub-segments, so $\beta\notin\Fsi$.

($\Rightarrow$) We show $\Fex\subseteq\Fsi$ when $D=0$ or $\Delta\le\ell$. If
$D=0$, some witness $s$ has $a\le s\le b$, and the segment $[s,s+\ell]$
satisfies $p\le b=b-D$ and $q\ge a+\ell$.

Let $\Delta\le\ell$ and $\beta\in\Fex$. Let $\sigma_1,\dots,\sigma_m$ be the
maximal true segments that contain witnesses, ordered in time, with
$\sigma_i=[p_i,q_i]$. By Lemma~\ref{lem:segment}, $\sigma_i$ covers
$U_i=J_i\cap\Dset$ with $J_i=[a-q_i+\ell,\ b-p_i]$. For $i<i'$ we have
$p_{i'}\ge q_i+2$, so, using $b\le a+\ell$,
\begin{equation}\label{eq:pf-gap}
\max J_{i'}=b-p_{i'}\le b-q_i-2\le a+\ell-q_i-2=\min J_i-2 .
\end{equation}
Thus the $J_i$ are ordered decreasingly and separated by at least one integer.
Suppose two segments contribute, i.e., $U_i\neq\varnothing\neq U_{i'}$ for
some $i<i'$, and choose $i'$ as the next contributing index after $i$. Let
$g=\min J_i-1$. Since $U_{i'}\neq\varnothing$, $\max J_{i'}\ge0$, and
\eqref{eq:pf-gap} gives $g>\max J_{i'}\ge0$. Since $U_i\neq\varnothing$,
$\min J_i\le D$, so $g\le D-1$. Thus $g\in\Dset$. Now $g\notin J_i$; for
$i''\ge i'$, $\max J_{i''}\le\max J_{i'}<g$; for $i''<i$,
$\min J_{i''}\ge\max J_i+2>g$ by \eqref{eq:pf-gap}; and the segments strictly
between $i$ and $i'$ do not contribute. Hence $g\notin\Cov(b+\ell)$, which
contradicts $\beta\in\Fex$. Therefore exactly one segment contributes, it
covers $\Dset$, and $\beta\in\Fsi$.
\end{IEEEproof}

\section{Proofs for Section~\ref{sec:certs}}

\begin{sproposition}[Monotone exact family]\label{prop:exact-mono}
\StmtExactMono
\end{sproposition}
\begin{IEEEproof}
Backward induction on $k$. At $H$, if $\Tset_H(q)\neq\varnothing$ then
$\Acc(q)$, hence $\Acc(q')$ because $\Acc$ is upward closed
(Lemma~\ref{lem:mono}), and $\Tset_H(q)=\Tset_H(q')$. Let $k<H$ and
$z\in\Tset_k(q)$ with witness input $v$ and successor $z^+=Az+Bv$. Then
$\Adm_k(z,q')$ holds because $\Qset_k$ is upward closed. By
Lemma~\ref{lem:mono},
$\Gamma_{k+1}(q,\chi(Cz^+))\preceq\Gamma_{k+1}(q',\chi(Cz^+))$, so the induction
hypothesis gives
$z^+\in\Tset_{k+1}(\Gamma_{k+1}(q,\chi(Cz^+)))
\subseteq\Tset_{k+1}(\Gamma_{k+1}(q',\chi(Cz^+)))$, and the same $v$ shows
$z\in\Tset_k(q')$.
\end{IEEEproof}

\begin{restated}{Lemma}{lem:minpre}
\StmtMinPre
\end{restated}
\begin{IEEEproof}
\emph{Upward closure.} Let $q\in\Pre_k(g,\sigma)$ and $q\preceq q'\in\Qset$.
Then $q'\in\Qset_k$ by Lemma~\ref{lem:mono}, and
$\Gamma_{k+1}(q',\sigma)\succeq\Gamma_{k+1}(q,\sigma)\succeq g$.

\emph{Reduction to one task.} Membership in $\Qset_k$, the update, the order,
and the condition $\Gamma_{k+1}(q,\sigma)\succeq g$ with $g\neq\bot$ are all
componentwise (the last requires every component to be non-failed). Hence
$\Pre_k(g,\sigma)=\prod_j\Pre_{j}$ with
$\Pre_j=\{q_j\in\Qset_j: c_j\le\cbar_{j,k},\
\Gamma_{j,k+1}(q_j,\sigma_j)\succeq g_j\}$, and minimal elements of a product
of posets are the products of minimal elements. It therefore suffices to show,
for one task (index dropped), that
\begin{equation}\label{eq:pf-minpre}
X_j\subseteq\Pre_j
\quad\text{and}\quad
\forall q\in\Pre_j\ \exists x\in X_j:\ x\preceq q .
\end{equation}
Then $\min\Pre_j=\min X_j$. Indeed, if $m\in\min\Pre_j$, some $x\in X_j$
satisfies $x\preceq m$ and $x\in\Pre_j$, so $x=m$, and $m$ is minimal in
$X_j\subseteq\Pre_j$. Conversely, if $x\in\min X_j$ and $q\in\Pre_j$ with
$q\preceq x$, some $x'\in X_j$ satisfies $x'\preceq q\preceq x$, so $x'=x$ and
$q=x$.

Two facts are used throughout. First, $\cbar_{k+1}\le\cbar_k$, and
$g\in\Qset_{k+1}$ gives $c^*\le\cbar_{k+1}$. Second, if a state has image
$(\hat c^+,\hat r^+)$ before the failure test with $\hat c^+\le c^*$, then the
image is not $\bot$, because $c^*\le\cbar_{k+1}$.

\emph{Candidates are predecessors.} If $\sigma_j=0$ and $r^*=0$, the candidate
$(c^*,0)$ lies in $\Qset_k$ since $c^*\le\cbar_{k+1}\le\cbar_k$; its image is
$(c^*,0)\succeq(c^*,r^*)$. If $\sigma_j=1$, the candidate
$x_A=(c^*,\max\{0,r^*-1\})$ lies in $\Qset_k$ for the same reason; its run
length becomes $\min\{\ell+1,\max\{0,r^*-1\}+1\}\ge r^*$ because
$r^*\le\ell+1$, and its coverage index cannot increase, so its image dominates
$g$. For $x_B=(c_B,\ell)$ with $c_B=\min\{D+1,\hi+1\}$ and $\lo\le\hi$, we have
$\hi\ge0$ and $\hi\le b-s=b+\ell-k-1$, so $c_B\le\min\{D+1,b+\ell-k\}=\cbar_k$
and $x_B\in\Qset_k$. Its image completes the witness, with run length
$\ell+1\ge r^*$; since $\lo\le\hi$ and $\hi\ge c_B-1$, the coverage index
becomes $\min\{c_B,\lo\}\le\lo\le c^*$.

\emph{Every predecessor dominates a candidate.} Let $q=(c,r)\in\Pre_j$ with
image $(c^+,r^+)\ne\bot$, $c^+\le c^*$ and $r^+\ge r^*$.
If $\sigma_j=0$, then $r^+=0\ge r^*$ forces $r^*=0$, and $c=c^+\le c^*$, so
$(c^*,0)\preceq q$.
If $\sigma_j=1$ and $c\le c^*$, then $r^+=\min\{\ell+1,r+1\}\ge r^*$ gives
$r\ge r^*-1$, so $x_A\preceq q$.
If $\sigma_j=1$ and $c>c^*$, then $c^+<c$, which requires a completion, so
$r\ge\ell$, and the first branch of \eqref{eq:compact-update} with $\lo<c$:
$\lo\le\hi$, $\hi\ge c-1$ and $c^+=\lo\le c^*$. Hence $x_B\in X_j$, and
$c\le\min\{D+1,\hi+1\}=c_B$ together with $r\ge\ell$ gives $x_B\preceq q$.

This establishes \eqref{eq:pf-minpre}. Each $X_j$ has at most two
elements, so $|X_1\times\dots\times X_{n_r}|\le2^{n_r}$.
\end{IEEEproof}

\begin{restated}{Theorem}{thm:dp-eq}
\StmtDPEq
\end{restated}
\begin{IEEEproof}
Backward induction on $k$.

\emph{Base.} By \eqref{eq:win-H}, $\Win_H(\rho,q)=1$ iff $\RA_H(\rho)$,
$P_\rho\subseteq\Zpost$ and $\Acc(q)$. Now $\Acc(q)$ iff $q\neq\bot$ and
$c_j=0$ for all $j$, iff $q\succeq((0,0))_j$. This matches the initialization
in \eqref{eq:ordered}.

\emph{Step.} Assume the claim at $k+1\le H$. If $\RA_k(\rho)$ fails, both sides
are false. Otherwise, by \eqref{eq:win-k},
\begin{align*}
&\Win_k(\rho,q)=1\\
&\iff q\in\Qset_k\wedge\exists e:\rho_e=\rho,\
  \Win_{k+1}(\rho'_e,\Gamma_{k+1}(q,\sigma_e))=1\\
&\iff q\in\Qset_k\wedge\exists e,\ \exists g\in M_{k+1}(\rho'_e):\
  g\preceq\Gamma_{k+1}(q,\sigma_e)\\
&\iff \exists e,\ \exists g\in M_{k+1}(\rho'_e):\ q\in\Pre_k(g,\sigma_e).
\end{align*}
The second equivalence is the induction hypothesis; if
$\Gamma_{k+1}(q,\sigma_e)=\bot$, both sides are false, since $\Win$ vanishes at
$\bot$ and no $g\in\Qset$ satisfies $g\preceq\bot$. The elements of
$M_{k+1}(\rho'_e)$ lie in $\Qset_{k+1}$, as required by
Lemma~\ref{lem:minpre}, because $\Win_{k+1}(\rho',g)=1$ implies $g\in\Qset_{k+1}$
(for $k+1<H$ by \eqref{eq:win-k}; for $k+1=H$ because $\Acc(g)$ implies
$c_j=0\le\cbar_{j,H}$). By Lemma~\ref{lem:minpre}, each
$\Pre_k(g,\sigma_e)$ is upward closed in the finite poset $\Qset$, so $q$
belongs to it iff $m\preceq q$ for some $m\in\min\Pre_k(g,\sigma_e)$. Let $U$
be the union of these minimal sets over all $e$ and $g$. Then
$\Win_k(\rho,q)=1$ iff $q$ dominates some element of $U$, iff $q$ dominates
some element of $\min U=M_k(\rho)$.
\end{IEEEproof}

\begin{restated}{Theorem}{thm:sound}
\StmtSound
\end{restated}
\begin{IEEEproof}
(a) By Theorem~\ref{thm:dp-eq}, $\Win_k(\rho,\cdot)$ is the indicator of the
upward closure of $M_k(\rho)$, so it is monotone, and so is the union
$\That_k$.

(b) Let $z\in P_\rho$ with $\Win_k(\rho,q)=1$ and $k<H$. By \eqref{eq:win-k},
$q\in\Qset_k$, $\RA_k(\rho)$ holds, and some edge $e$ with $\rho_e=\rho$
satisfies $\Win_{k+1}(\rho'_e,\Gamma_{k+1}(q,\sigma_e))=1$. $\RA_k(\rho)$ gives
$z\in\mathcal Z$ and $Cz\in\bar S_k$, hence $\Adm_k(z,q)$. (V1) gives
$v=F_ez+g_e\in\mathcal V$. (V2) gives $z^+\in P_{\rho'_e}$. (V3) gives
$Cz^+\in\bar R_j$ whenever $\sigma_{e,j}=1$, i.e., $\sigma_e\le\chi(Cz^+)$. For
every $\beta\ge\sigma_e$, Lemma~\ref{lem:mono} gives
$\Gamma_{k+1}(q,\beta)\succeq\Gamma_{k+1}(q,\sigma_e)$, and by (a) applied to
$\Win_{k+1}(\rho'_e,\cdot)$,
$\Win_{k+1}(\rho'_e,\Gamma_{k+1}(q,\beta))=1$. Therefore
$z^+\in P_{\rho'_e}\subseteq\That_{k+1}(\Gamma_{k+1}(q,\beta))$.

(c) If $\That_H(q)\neq\varnothing$, some $\rho$ with $P_\rho\neq\varnothing$ has
$\Win_H(\rho,q)=1$, so $\Acc(q)$ by \eqref{eq:win-H}. Every such $\rho$ has
$P_\rho\subseteq\Zpost$.

(d) Backward induction on $k$. At $H$, $z\in\That_H(q)$ lies in some $P_\rho$
with $\Win_H(\rho,q)=1$. Then $\Acc(q)$, $z\in\Zpost$, and $Cz\in\bar S_H$
by $\RA_H(\rho)$, so $z\in\Tset_H(q)$. For $k<H$, let $z\in\That_k(q)$. Part (b)
with $\beta=\chi(Cz^+)$ gives $\Adm_k(z,q)$ and an input $v\in\mathcal V$ with
$z^+\in\That_{k+1}(\Gamma_{k+1}(q,\chi(Cz^+)))$, which is contained in
$\Tset_{k+1}(\Gamma_{k+1}(q,\chi(Cz^+)))$ by the induction hypothesis. Hence
$z\in\Tset_k(q)$.
\end{IEEEproof}

\subsection*{Branching libraries}

\begin{restated}{Proposition}{prop:robpre}
\StmtRobPre
\end{restated}
\begin{IEEEproof}
\emph{Upward closure} is as in Lemma~\ref{lem:minpre}: if $q\in\PreA_k$ and
$q\preceq q'\in\Qset$, then $q'\in\Qset_k$ and
$\Gamma_{k+1}(q',\sigma^b)\succeq\Gamma_{k+1}(q,\sigma^b)\succeq g^b$ for every
$b$ by Lemma~\ref{lem:mono}.

\emph{Reduction to one task.} Membership in $\Qset_k$, the update and the
order are componentwise, and every $g^b\neq\bot$, so
$\PreA_k=\prod_j\PreA_j$ with
$\PreA_j=\{q_j\in\Qset_j:c_j\le\cbar_{j,k},\
\Gamma_{j,k+1}(q_j,\sigma^b_j)\succeq g^b_j\ \forall b\}$; minimal elements of
a product are products of minimal elements. Fix $j$ and drop the index.

\emph{Grouping by bit.} $\Gamma_{k+1}(q,\sigma^b_j)$ depends on $b$ only
through the bit, so the condition reads $\Gamma_{k+1}(q,0)\succeq g^b$ for
$b\in I^0$ and $\Gamma_{k+1}(q,1)\succeq g^b$ for $b\in I^1$. On $\Qset_j$
the order $\preceq$ is a lattice with join $(c,r)\vee(c',r')=(\min\{c,c'\},
\max\{r,r'\})$, and an element dominates every $g^b$ of a set iff it
dominates their join; the join of elements of $\Qset_{k+1}$ lies in
$\Qset_{k+1}$. With bit $0$ the run length resets and no witness completes,
so $\Gamma_{k+1}(q,0)=(c,0)$ unless it is $\bot$. Hence the bit-$0$ condition
holds iff $r^b=0$ for all $b\in I^0$ and $c\le c^b$ for all $b\in I^0$ (the
failure test is then implied by $c^b\le\cbar_{k+1}$). If some $r^b>0$,
$\PreA_j=\varnothing=X_j$. Otherwise, together with $q\in\Qset_k$, the
bit-$0$ condition and membership define $U^0=\{(c,r)\in\Qset_j:c\le
C^0\}$, which is upward closed with the single minimal element $(C^0,0)$. If
$I^1=\varnothing$, $\PreA_j=U^0$ and $\min\PreA_j=\{(C^0,0)\}=X_j$.
Otherwise the bit-$1$ condition is $q\in\Pre_j(G^1,1)$ with the joined target
$G^1=(C^1,R^1)\in\Qset_{k+1}$, and $\PreA_j=U^0\cap\Pre_j(G^1,1)$.

\emph{Intersection.} By Lemma~\ref{lem:minpre}, $\Pre_j(G^1,1)$ is upward
closed with minimal elements among $x_A=(C^1,\max\{0,R^1-1\})$ and, if
$\lo\le\hi$ and $\lo\le C^1$, $x_B=(\min\{D+1,\hi+1\},\ell)$. The
intersection of two upward closed sets is generated by the joins of their
generators: every $q\in\PreA_j$ dominates $(C^0,0)$ and some $x\in\{x_A,x_B\}$,
hence $x\vee(C^0,0)\preceq q$; conversely each join dominates a generator of
each set and lies in $\Qset_j$, so it belongs to both. The joins are
$(\min\{C^0,C^1\},\max\{0,R^1-1\})$ and $(\min\{C^0,D+1,\hi+1\},\ell)$, which
is $X_j$. As in Lemma~\ref{lem:minpre}, $X_j\subseteq\PreA_j$ and every element
of $\PreA_j$ dominates an element of $X_j$, so $\min\PreA_j=\min X_j$, and
$|X_j|\le2$.
\end{IEEEproof}

\begin{slemma}[Suffix simulation]\label{lem:suffixsim}
\StmtSuffixSim
\end{slemma}
\begin{IEEEproof}
Fix one task (failure and the update are componentwise) and $q=(c,r)$. An
offset $d$ is expired at $k$ iff $d$ is unserved and $d\ge b+\ell-k$. For
$\iota(q)$ the unserved offsets are $[0,c-1]$, so $\iota(q)$ is failed iff
$c-1\ge\max\{0,b+\ell-k\}$, i.e., iff $c>\cbar_k$ (recall $c\le D+1$); this is
the last claim. The run length evolves identically in both monitors. If no
witness completes at $k$, or the served interval $[\lo,\hi]$ of the completing
witness ($s=k-\ell$) is empty, both monitors keep the served set and the
claim follows from the failure characterization. If $\lo\le\hi$ and $\hi\ge
c-1$, then $[\lo,\hi]\cup[c,D]=[\min\{c,\lo\},D]=\iota(c^+)$, and again both
fail iff $c^+>\cbar_k$. If $\lo\le\hi<c-1$, then $\hi<D$, so $\hi=b-s=
b+\ell-k$; offset $c-1$ is unserved and $c-1>b+\ell-k$, so it is expired and
the explicit monitor fails, while $c^+=c>\max\{0,b+\ell-k\}+1>\cbar_k$ gives
$\Gamma_k(q,\beta)=\bot$.
\end{IEEEproof}

\begin{restated}{Theorem}{thm:robcert}
\StmtRobCert
\end{restated}
\begin{IEEEproof}
(a) Backward induction on $k$. At $H$, $\Win_H(\rho,q)=1$ iff $\RA_H(\rho)$,
$\rho\in\Kpost$ and $q\succeq((0,0))_j$, which matches $M_H$ and is upward
closed. Let $k<H$ and assume the claim at $k+1$. If $\RA_k(\rho)$ fails both
sides vanish. Otherwise, by \eqref{eq:rwin-k} and the induction hypothesis,
$\Win_k(\rho,q)=1$ iff $q\in\Qset_k$ and there are an edge $e\in A(\rho)$ and,
for every branch $b$, a generator $g^b\in M_{k+1}(\rho_b)$ with $g^b\preceq
\Gamma_{k+1}(q,\sigma^b)$ (if some $\Gamma_{k+1}(q,\sigma^b)=\bot$ both sides
vanish). This is $q\in\PreA_k((g^b)_b)$ for some $e$ and some target vector.
Each $\PreA_k$ is upward closed in the finite poset $\Qset$
(Proposition~\ref{prop:robpre}), so membership is domination of one of its
minimal elements, and the union over $e$ and target vectors is generated by
$M_k(\rho)$. Upward closure of $\Win_k(\rho,\cdot)$ follows, and
$\bot$ dominates no generator.

(b) Backward induction. At $H$, $\iota(q)$ serves every offset iff $c_j=0$ for
all $j$, iff $\Acc(q)$. For $k<H$ the explicit recursion requires $\iota(q)$
not failed at $k$, which by Lemma~\ref{lem:suffixsim} is $q\in\Qset_k$, and
evaluates every branch at $\Gamma^{\mathrm{bit}}_{k+1}(\iota(q),\sigma^b)$,
which by the same lemma is failure exactly when $\Gamma_{k+1}(q,\sigma^b)=\bot$
and otherwise $\iota(\Gamma_{k+1}(q,\sigma^b))$; the induction hypothesis
gives equal branch values, hence equal disjunctions.

(c) With one branch, the conjunction in \eqref{eq:rwin-k} has one term, and
$\PreA_k$ is $\Pre_k(g,\sigma_e)$ of Lemma~\ref{lem:minpre}.
\end{IEEEproof}

\begin{restated}{Theorem}{thm:robloop}
\StmtRobLoop
\end{restated}
\begin{IEEEproof}
Induction on $k\in[0,H-1]$ with hypothesis $\Win_k(\rho_k,q_k)=1$ and
$x_k\in P_{\rho_k}$, true at $k=0$. By \eqref{eq:rwin-k}, $q_k\in\Qset_k$,
$\RA_k(\rho_k)$ holds, so the state constraints active at $k$ hold on
$P_{\rho_k}\ni x_k$, and an edge $e\in A(\rho_k)$ with
$\Win_{k+1}(\rho_b,\Gamma_{k+1}(q_k,\sigma^b))=1$ for every $b$ exists.
(B1) gives $u_k=\kappa_e(x_k)\in U(x_k)$. For the realized $w_k$, (B2)
provides a branch $b$ with $x_{k+1}\in P_{\rho_b}$ and
$\sigma^b\le\chi(h(x_{k+1}))$; the controller selects such a branch (if several
match, any of them). By Lemma~\ref{lem:mono},
$q_{k+1}=\Gamma_{k+1}(q_k,\chi(h(x_{k+1})))\succeq\Gamma_{k+1}(q_k,\sigma^b)$,
and Theorem~\ref{thm:robcert}(a) gives $\Win_{k+1}(\rho_b,q_{k+1})=1$. At $H$,
\eqref{eq:rwin-H} gives $\Acc(q_H)$, $\RA_H(\rho_H)$ and $\rho_H\in\Kpost$.
For $k\ge H$, Proposition~\ref{prop:postfp} keeps $\rho_k\in\Kpost$. Since
$q_k$ is the monitor run on the actual bits from the recorded history, the
compact monitor theorem gives $\Cov_j(H)=\Dset$ for every $j$, so every reach
task holds under every offset for the tightened bits. Moreover
$q_k\in\Qset_k$ for every $k\le H$, so no task is failed at any $k\le H$; since
$b_j-d+\ell_j\le H$, offset $d$ is not expired at $b_j-d+\ell_j$, i.e.,
$d\in\Cov_j(b_j-d+\ell_j)$. A safety task under
offset $d$ at time $t$ reads $x_{t-d}$: for $t-d\in[-D,-1]$ the pre-mission
premise gives it, for $t-d\in[0,H]$ the region invariant $x_k\in P_{\rho_k}$
with $\RA_k(\rho_k)$ does, and for $t-d>H$ the regions of $\Kpost$, which lie
in the static safe set $Z_{\rm post}$, do. The clock is the original task clock
throughout.
\end{IEEEproof}

\begin{sproposition}[Multi-successor enlargement]\label{prop:robmono}
\StmtRobMono
\end{sproposition}
\begin{IEEEproof}
Backward induction: the terminal values coincide, and at $k<H$ the
disjunction in \eqref{eq:rwin-k} for $\Blib'$ ranges over a superset of edges
whose conjunctions are pointwise larger by the induction hypothesis.
\end{IEEEproof}

\begin{sproposition}[Robust post-mission set]\label{prop:postfp}
\StmtPostFP
\end{sproposition}
\begin{IEEEproof}
The iteration is decreasing on a finite set, so it stops at a fixed point
$\Kpost$, and every $\rho\in\Kpost$ has an edge $e_{\rm post}(\rho)$ whose
branch destinations all lie in $\Kpost$. If $x_k\in P_{\rho_k}$ with
$\rho_k\in\Kpost$, (B1) makes the input admissible and (B2) places $x_{k+1}$ in
a branch region of $e_{\rm post}(\rho_k)$, which belongs to $\Kpost$. Every
other fixed point below $\mathcal K^0$ is contained in each iterate, so
$\Kpost$ is the greatest one.
\end{IEEEproof}

\begin{sproposition}[No single invariant position box]\label{prop:box}
\StmtBoxObstruction
\end{sproposition}
\begin{IEEEproof}
Let $B=[\underline x,\bar x]\times[\underline y,\bar y]$ and fix
$\theta\in\Theta$. At $x^1=(\bar x,\bar y,\theta)$ and $x^2=(\bar x,
\underline y,\theta)$, let $v_i$ be the speed of $\kappa(x^i)$. With
$w=(\bar w_x,\bar w_y,0)$ at $x^1$, invariance needs $hv_1\cos\theta\le-\bar
w_x<0$ and $hv_1\sin\theta\le-\bar w_y<0$, so $\cos\theta$ and $\sin\theta$
are nonzero with the sign of $-v_1$. With $w=(\bar w_x,-\bar w_y,0)$ at $x^2$,
it needs $hv_2\cos\theta<0<hv_2\sin\theta$, so $\cos\theta$ and $\sin\theta$
have opposite signs, a contradiction. Subdividing $B$ into smaller regions does
not help, because the argument only uses the two corners of $B$.
\end{IEEEproof}

\subsection{Partition cost of single-successor libraries}\label{sec:partcost}
Fix one position coordinate $p$ of $x$ and write $x=(p,x_{-p})$. Consider
libraries whose regions are product boxes, $P_\rho=I_\rho\times
P_\rho^{-p}$ with $I_\rho$ a closed interval of positive width $|I_\rho|$,
for systems in which $\mathcal W=[-\bar w,\bar w]\times\mathcal W_{-p}$,
$\bar w>0$, does not depend on $(x,u)$ and every edge's closed loop has the
form
\begin{equation}\label{eq:partclass}
p^+=p+g_e(x_{-p},w_{-p})+w_p,
\end{equation}
with $g_e$ independent of $p$. Heading primitives of the unicycle
\eqref{eq:unicycle} belong to this class, with $g_e=hv\cos\theta$ for $p=p_x$;
position-feedback primitives do not. An edge is \emph{single-successor} if
$|\Succ(e)|=1$.

\begin{slemma}\label{lem:partwidth}
For a verified single-successor edge from $\rho$ to $\rho'$ in this class,
$|I_{\rho'}|\ge|I_\rho|+2\bar w$, and the $p$-projection of its reachable
set $S_e=\{x^+:x\in P_\rho,\ w\in\mathcal W\}$ contains an interval of width
at least $|I_\rho|+2\bar w$.
\end{slemma}
\begin{IEEEproof}
Fix any $x_{-p}\in P^{-p}_\rho$ and $w_{-p}\in\mathcal W_{-p}$, and let
$c=g_e(x_{-p},w_{-p})$. Because $P_\rho$ and $\mathcal W$ are products,
$p\in I_\rho$ and $w_p\in[-\bar w,\bar w]$ vary freely, so by
\eqref{eq:partclass} the $p$-projection of $S_e$ contains
$[\min I_\rho+c-\bar w,\ \max I_\rho+c+\bar w]$. By (B2) with a single branch,
$S_e\subseteq P_{\rho'}$, so this interval lies in $I_{\rho'}$.
\end{IEEEproof}

\begin{scorollary}[Common terminal set]\label{cor:partA}
Let $\Blib_1\subseteq\Blib$ be the single-successor edges of a verified
library in this class, on the same regions with the same $\RA$ and $\Kpost$.
Let reach task $j$ have $\bar R_j=\bar I\times\bar R_j^{-p}$, and let $q_0$
leave its offset $0$ unserved, so that completion needs a one of task $j$ at
some time $k\ge a_j$. If $2\bar w a_j\ge|\bar I|$, then
$\Win^{\Blib_1}_0(\rho,q_0)=0$ for every region $\rho$, for every partition
into product boxes.
\end{scorollary}
\begin{IEEEproof}
Under $\Blib_1$ every branch choice is forced, so a certificate fixes one
region sequence $\rho_0,\rho_1,\dots$ with edges $e_k$ from $\rho_k$ to
$\rho_{k+1}$, and \eqref{eq:rwin-k} advances the monitor with the labels
$\sigma_k$ of $e_k$ only. By Lemma~\ref{lem:partwidth} and induction,
$|I_{\rho_k}|\ge|I_{\rho_0}|+2\bar wk$, and the $p$-projection of the
reachable set of $e_{k-1}$ contains an interval of width at least
$|I_{\rho_0}|+2\bar wk>2\bar wk$. A label $1$ for task
$j$ requires, by (B2), $\chi_j(h(x^+))=1$ for every $x^+$ in that reachable
set, hence its $p$-projection inside $\bar I$, which is impossible for
$k\ge a_j$ because then $2\bar wk\ge|\bar I|$. So all labels of task $j$ at
times $k\ge a_j$ are $0$, offset $0$ is never served, $\Acc$ fails at $H$, and
\eqref{eq:rwin-H} gives $0$ whatever the terminal set.
\end{IEEEproof}

For R1 and R2, $p=p_x$, $\bar w=\PartWbar$, $a=\PartA$ and
$|\bar I|=\PartRbarWidth$\,m, so $2\bar wa=\PartLHS\ge\PartRbarWidth$: no
single-successor library of heading primitives on any product-box partition
certifies the task, which explains the empty sub-library of R2 independently of
the grid. The corollary does not cover position-feedback primitives or
non-product regions.

\begin{scorollary}[Own post set]\label{cor:partB}
If a single-successor library in this class must provide its own $\Kpost$,
computed as in Proposition~\ref{prop:postfp} from its own edges, then
$\Kpost=\varnothing$.
\end{scorollary}
\begin{IEEEproof}
Each $\rho\in\Kpost$ has a witness edge into $\Kpost$, so following witnesses
gives an infinite region sequence in the finite set $\Kpost$ whose widths
$|I_\rho|$ increase by at least $2\bar w>0$ per step, which is impossible.
\end{IEEEproof}
Corollary~\ref{cor:partB} is narrower in scope (it changes the terminal set)
and is not used for R2, where the sub-library keeps the full library's
$\Kpost$.

\section{Proofs for Section~\ref{sec:mpc}}

\begin{restated}{Theorem}{thm:rf}
\StmtRF
\end{restated}
\begin{IEEEproof}
Write $n=N_k=\min\{N,H-k\}$, which is at least $1$ because $N\ge1$ and $k<H$.
Because $k\le H-2$, the problem $\mathcal P_{k+1}$ exists and $N_{k+1}\ge1$. There are two cases for the horizon: if $k+n=H$,
then $N_{k+1}=\min\{N,H-k-1\}=H-k-1=n-1$ because $H-k=n\le N$; if $k+n<H$, then
$n=N<H-k$, so $N\le H-k-1$ and $N_{k+1}=N=n$.

\emph{Step 1: dominance of the executed monitor.} By
Assumption~\ref{ass:onesided}(a), $\hat\beta^*_{1|k,j}=1$ implies
$Cz^*_{1|k}\in\bar R_j$, i.e., $\chi_j(Cz^*_{1|k})=1$; by (b) this is the same
$\chi$ used by the executed monitor. Hence
$\hat\beta^*_{1|k}\le\chi(Cz^*_{1|k})=\chi(Cz_{k+1})$, where the equality uses
hypothesis (i). Lemma~\ref{lem:mono} with $q=q'=q_k$ gives
\[
q^*_{1|k}=\Gamma_{k+1}(q_k,\hat\beta^*_{1|k})\preceq
\Gamma_{k+1}(q_k,\chi(Cz_{k+1}))=q_{k+1}.
\]

\emph{Step 2: shifted candidate.} For $i\in[0,n-2]$ set
$v_{i|k+1}=v^*_{i+1|k}$, and for $i\in[1,n-1]$ set
$\hat\beta_{i|k+1}=\hat\beta^*_{i+1|k}$. With $z_{0|k+1}=z_{k+1}=z^*_{1|k}$, the
dynamics give $z_{i|k+1}=z^*_{i+1|k}$ for $i\in[0,n-1]$. Let
$q_{0|k+1}=q_{k+1}$ and
$q_{i|k+1}=\Gamma_{k+1+i}(q_{i-1|k+1},\hat\beta_{i|k+1})$. By Step~1 and
induction with Lemma~\ref{lem:mono} (same time index $k+1+i$ and same bits),
\begin{equation}\label{eq:pf-dom}
q_{i|k+1}\succeq q^*_{i+1|k},\qquad i\in[0,n-1].
\end{equation}
For $i\in[0,n-2]$, the constraints of $\mathcal P_{k+1}$ at step $i$ refer to
absolute time $k+1+i$, which is step $i+1$ of $\mathcal P_k$: $v_{i|k+1}\in\mathcal V$,
$z_{i|k+1}\in\mathcal Z$ and $Cz_{i|k+1}\in\bar S_{k+1+i}$ are inherited. The
condition $q_{i|k+1}\in\Qset_{k+1+i}$ follows from \eqref{eq:pf-dom} because
$\Qset_{k+1+i}$ is upward closed. The one-sided implications involve the same
states and bits as before.

\emph{Step 3a: $k+n=H$.} Then $N_{k+1}=n-1$, and the terminal constraint of
$\mathcal P_{k+1}$ is $z_{n-1|k+1}=z^*_{n|k}\in\That_H(q_{n-1|k+1})$. It follows
from $z^*_{n|k}\in\That_H(q^*_{n|k})$, \eqref{eq:pf-dom} and
Theorem~\ref{thm:sound}(a).

\emph{Step 3b: $k+n<H$.} Then $N_{k+1}=n$ and one more step is needed. By
\eqref{eq:pf-dom} and Theorem~\ref{thm:sound}(a),
$z^*_{n|k}\in\That_{k+n}(q^*_{n|k})\subseteq\That_{k+n}(q_{n-1|k+1})$, so
$z^*_{n|k}\in P_\rho$ for some $\rho$ with $\Win_{k+n}(\rho,q_{n-1|k+1})=1$.
Since $k+n<H$, Theorem~\ref{thm:sound}(b) applies. It gives
$\Adm_{k+n}(z^*_{n|k},q_{n-1|k+1})$, which is the admissibility constraint at
the new step $n-1$, and an edge $e$ from $\rho$. Set
$v_{n-1|k+1}=F_ez^*_{n|k}+g_e\in\mathcal V$ and $\hat\beta_{n|k+1}=\sigma_e$.
Then $z_{n|k+1}=Az^*_{n|k}+Bv_{n-1|k+1}$ satisfies $\sigma_e\le\chi(Cz_{n|k+1})$,
which is the one-sided implication, and with $\beta=\sigma_e$,
\[
\begin{aligned}
z_{n|k+1}&\in\That_{k+1+n}\big(\Gamma_{k+1+n}(q_{n-1|k+1},\sigma_e)\big)\\
&=\That_{k+1+n}(q_{n|k+1}),
\end{aligned}
\]
which is the terminal constraint.

All constraints of $\mathcal P_{k+1}(z_{k+1},q_{k+1})$ hold, so it is feasible.
\end{IEEEproof}

\begin{restated}{Theorem}{thm:closedloop}
\StmtClosedLoop
\end{restated}
\begin{IEEEproof}
(i) Induction on $k$. At $k=0$, the stored candidate is the verified initial
candidate, and a newly returned candidate is selected only if it passes the
check of Assumption~\ref{ass:onesided}(a); either way the selected candidate is
feasible for $\mathcal P_0(z_0,q_0)$. Let $k\le H-2$, and let the candidate
selected at $k$ be feasible for $\mathcal P_k(z_k,q_k)$. Algorithm~\ref{alg:closedloop}
applies its first input without resetting the nominal center, so
$z_{k+1}=z^*_{1|k}$, and it updates the monitor with exact bits; these are
hypotheses (i) and (ii) of Theorem~\ref{thm:rf}. Since $k<H-1$, the algorithm
stores the shifted candidate of Theorem~\ref{thm:rf}, which is feasible for
$\mathcal P_{k+1}(z_{k+1},q_{k+1})$. The candidate selected at $k+1$ is either
this stored candidate or a returned candidate that passes the check, so it is
feasible. For $H=1$ only the base case occurs.

(ii) By Assumption~\ref{ass:tube} and induction,
$e_{k+1}=(A+BK)e_k+w_k\in(A+BK)\mathcal E\oplus\mathcal W\subseteq\mathcal E$.
For $k\le H-1$, $\Adm_k(z_k,q_k)$ gives $z_k\in\mathcal Z$, so
$x_k=z_k+e_k\in(\mathcal X\ominus\mathcal E)\oplus\mathcal E\subseteq\mathcal X$
and $u_k=v_{0|k}+Ke_k\in(\mathcal U\ominus K\mathcal E)\oplus K\mathcal E
\subseteq\mathcal U$. At $H$, part (iii) gives $z_H\in\Zpost\subseteq\mathcal Z$,
so $x_H\in\mathcal X$.

(iii) Consider $k=H-1$, where $N_{H-1}=\min\{N,1\}=1$ because $N\ge1$. By (i),
the candidate selected at $H-1$ is feasible, so it has
$z_H=z_{1|H-1}\in\That_H(q_{1|H-1})$. This set is therefore nonempty, and
Theorem~\ref{thm:sound}(c) gives $\Acc(q_{1|H-1})$ and $z_H\in\Zpost$. Step~1 of
the proof of Theorem~\ref{thm:rf} applies verbatim at $k=H-1$ and gives
$q_{1|H-1}\preceq q_H$. Since $\Acc$ is upward closed (Lemma~\ref{lem:mono}),
$\Acc(q_H)$ holds.

(iv) By construction, $q_H$ is the compact monitor
(Definition~\ref{def:compact}) driven by the exact tightened bits
$\beta(t)=\chi(\ybar_t)$ of the executed nominal output, including the recorded
history for $t<0$. Since $q_H\neq\bot$ and $c_j(q_H)=0$,
Corollary~\ref{cor:compact} gives $\Cov_j(H)=\Dset$ for every reach task, and
Lemma~\ref{lem:L1}(a) shows that every reach conjunct of $\varphi_\beta$ holds
for $(\gamma,\beta)(\cdot-d)$, for every $d\in\Dset$. For safety task $i$ and
$t\in[\alpha_i-D,\omega_i]$: if $t<0$, then $\gamma_i(t)=1$ by
Assumption~\ref{ass:history}; if $0\le t\le H-1$, then $\Adm_t(z_t,q_t)$ gives
$Cz_t\in\bar S_t\subseteq\bar S_i$; if $t=H$, then $z_H\in P_\rho$ for some
$\rho$ with $\Win_H(\rho,\cdot)=1$, so $\RA_H(\rho)$ gives $Cz_H\in\bar S_i$.
Lemma~\ref{lem:L1}(b) yields every safety conjunct under every offset. Hence
$((\gamma,\beta)(\cdot-d),0)\models\varphi_\beta$ for all $d\in\Dset$, and
Lemma~\ref{lem:spatial} gives $(\ytil^{d,\eta},0)\models\varphi$.

Finally, for $t\in[0,H]$, $y_t=Cz_t+Ce_t=\ybar_t+\eta(t)$ with
$\eta(t)=Ce_t\in C\mathcal E\subseteq\Ball$, so $y=\ytil^{0,\eta}$ on $[0,H]$,
and $\varphi$ evaluated at time $0$ reads only $[0,H]$.
\end{IEEEproof}

\subsection{Encoding of the online problem and the full-horizon arms}\label{sec:encoding}
\emph{Monitor rows.} The update $q_{i+1|k}=\Gamma_{k+i+1}(q_{i|k},
\hat\beta_{i+1|k})$ in \eqref{eq:ocp} is encoded per task $j$ and absolute time
$t=k+i+1$; $\lo,\hi$ (with $s=t-\ell$) and $\cbar_{j,t}$ of
Definition~\ref{def:compact} are constants. For incoming $(c,r)$ (constants at
$i=0$, variables after) and bit $\beta$, the new integers $c'\in[0,D+1]$,
$r'\in[0,\ell+1]$ and binaries $u,w,\mathrm{adj},g,\tau,h$ satisfy
\begin{align*}
&(\ell{+}1)u\le r\le \ell+u,\quad (\ell{+}1)w\le r'\le \ell+w,\\
&r'\le(\ell{+}1)\beta,\quad r'\le r+1-u,\\
&r'\ge r+1-u-(\ell{+}2)(1-\beta),
\end{align*}
so that $u=[r=\ell+1]$, $w=[r'=\ell+1]$ and $r'=\beta\min\{r+1,\ell+1\}$;
and, if $\lo\le\hi$,
\begin{align*}
&(\hi{+}2)(1-\mathrm{adj})\le c\le \hi+1+(D-\hi)(1-\mathrm{adj}),\\
&(\lo{+}1)\tau\le c\le \lo+(D{+}1-\lo)\tau,\\
&g\le w,\ \ g\le\mathrm{adj},\ \ g\ge w+\mathrm{adj}-1,\\
&h\le g,\ \ h\le\tau,\ \ h\ge g+\tau-1,\\
&c-(D{+}1)h\le c'\le c,\quad |c'-\lo|\le(D{+}1)(1-h),
\end{align*}
so that $\mathrm{adj}=[c\le\hi+1]$, $\tau=[c\ge\lo+1]$, $g=w\wedge\mathrm{adj}$,
$h=g\wedge\tau$, and $c'=\min\{c,\lo\}$ if $h=1$, $c'=c$ otherwise. Here $\mathrm{adj}\equiv1$ when $\hi+1\ge D+1$; if $\lo>\hi$, the single row
$c'=c$ replaces the last five lines. Expiry is the hard row $c'\le\cbar_{j,t}$,
so a transition to $\bot$ is infeasible rather than represented. All constants
follow from $c\in[0,D+1]$ and $r\in[0,\ell+1]$, so the rows are exact for
integral values: they hold iff $(c',r')=\Gamma_{j,t}((c,r),\beta)\neq\bot$. Per
task and step this is at most \EncMonVars{} variables (two integer) and
\EncMonRows{} rows, independent of $D$ and $\ell$; the script checks this bound
on \EncMonCases{} task/time cases and finds it attained on the E2 models. A
one-hot transition table is kept only as a test oracle.

\emph{Bits, safety, terminal and solver.} A claimed bit adds one binary and
one row per face of $\bar R_j$, $G_{jh}Cz\le f_{jh}+M_{jh}(1-\hat\beta_j)$ with
$M_{jh}=\max\{0,\max_{z\in\operatorname{vert}\mathcal Z}(G_{jh}Cz-f_{jh})\}$;
the converse is not encoded (Assumption~\ref{ass:onesided}). Each active
safety union adds one selector per cell when there are several cells, with the
same exact constants. The terminal set $\That_{k+N_k}$ is a disjunction over
the options $(\rho,m)$ with $m\in M_{k+N_k}(\rho)$: one selector $s$ per option,
the rows of $P_\rho$ relaxed by exact constants when $s=0$,
$c_j\le m_{c,j}+(D+1-m_{c,j})(1-s)$ and $r_j\ge m_{r,j}s$, and $\sum s=1$.
HiGHS runs single-threaded with relative gap $10^{-6}$, primal and integer
feasibility tolerances $10^{-7}$, and the rows of geometric sets tightened by
$\delta\|a_{\rm cont}\|_1$ with $\delta=10^{-6}$; infeasibility of the
tightened model is reported as such, never as infeasibility of
\eqref{eq:ocp}. A candidate is used only after an exact rational check of
every constraint and of the monitor, on the solver's inputs converted exactly or, if that fails, rounded
to small rationals and checked again.

\emph{Size.} For $n_z$ states, $n_v$ inputs and $n_r$ tasks, the online model
has, without cost auxiliaries, at most $N_k(n_z+n_v+n_r+8n_r)$ variables plus
the safety and terminal selectors, and at most $N_k(n_z+|\mathcal V|+|\mathcal Z|+\sum_j|\bar R_j|+22n_r)$
rows plus the safety rows and $\sum_{(\rho,m)}(|P_\rho|+2n_r)+1$ terminal rows;
the last two depend on the geometry and the library. Table~\ref{tab:encsize}
gives the actual counts on the \EncQueries{} E2 test queries (medians, $k=0$,
before-patch library, $n_z=4$, $n_v=2$, $n_r=1$, $D=8$, $\ell=2$, $H=30$).

\begin{table}[t]
\caption{Variables/rows of the E2 models by component (medians over the
\EncQueries{} test queries). Online: $\mathcal P_0$ with $N=5,10,30$
(\EncOptionsTen{} terminal options for $N=10$). Full-H: the timed STL arm;
two-sided: the external arm \texttt{raman2}. Cost auxiliaries: absolute-value
variables and rows of the stage cost.}
\label{tab:encsize}
\centering\footnotesize\setlength{\tabcolsep}{2.5pt}
\begin{tabular}{@{}lrrrrr@{}}
\toprule
variables / rows & $N{=}5$ & $N{=}10$ & $N{=}30$ & full-H & two-sided \\
\midrule
dynamics, $V$, $Z$ & 30/80 & 60/160 & 180/480 & 180/512 & 180/512 \\
one-sided bits & 5/20 & 10/40 & 30/120 & 30/120 & 30/120 \\
converse bits & -- & -- & -- & -- & 120/150 \\
safety cells & --/16 & --/36 & --/116 & 60/150 & 60/150 \\
converse safety & -- & -- & -- & -- & 120/180 \\
monitor (13) & 35/100 & 70/200 & 194/528 & -- & -- \\
terminal selector & 25/851 & 23/783 & 1/35 & -- & -- \\
STL tree & -- & -- & -- & 175/706 & 175/1019 \\
cost auxiliaries & 10/20 & 20/40 & 60/120 & 60/120 & 60/120 \\
total & 105/1087 & 183/1259 & 465/1399 & 505/1608 & 745/2251 \\
\midrule
binary (total) & 55 & 83 & 165 & 265 & 505 \\
\bottomrule
\end{tabular}

\end{table}

\emph{Full-horizon arm and the closest published encoding.} The timed
full-horizon arm of E2 (method \texttt{stl}) encodes $\G_{[0,D]}\varphi$ at
$-D$, i.e.\ $\bigwedge_{d\in\Dset}\varphi$ under the shift $d$ (synchronous
shifts of all outputs), by a generic syntax-tree traversal in the style of
Raman et al.~\cite{raman2014mpc}, with the history samples $t\le0$ as
constants. Row by row:
\begin{center}\footnotesize
\begin{tabular}{@{}p{0.2\columnwidth}p{0.33\columnwidth}p{0.33\columnwidth}@{}}
\toprule
element & \cite{raman2014mpc} & \texttt{stl} (timed) \\
\midrule
predicate & binary per half-space, both directions & one bit per box, $\beta=1\Rightarrow Cz\in\bar R_j$ \\
safety atom & OR over cells of AND over half-spaces & $\gamma\le\sum_c s_c$, $s_c\Rightarrow Cz\in$ cell $c$ \\
AND, $\G$ & $b\le c_i$, $b\ge\sum c_i-(m-1)$ & $b\le c_i$ \\
OR, $\F$ & $b\ge c_i$, $b\le\sum c_i$ & $b\le\sum c_i$ \\
temporal & unrolled, one auxiliary per occurrence & same \\
dynamics, $\mathcal V$, $\mathcal Z$, $\Zpost$ & same & same \\
\bottomrule
\end{tabular}
\end{center}
So \texttt{stl} is that encoding with the converse half of every equivalence
removed (the one-directional form~\cite{kurtz2022fewer}); the conjunction of
a cell's half-spaces under one selector is equivalent once only one direction
is kept. Without solver margins both have the same feasible set in $(z,v)$,
but not the same LP relaxation, so we also ran the two-sided form
(\texttt{raman2}: the converse rows above, with complements of boxes and
cells as disjunctions of non-strict faces) under the E2 protocol: same
\ExtDen{} queries, budget, tolerances, cold start and exact validation, arms
alternating per query. As solved, the converse rows carry the same safety
margin as every tightened row, which removes a thin band near each predicate
boundary from the feasible set of \texttt{raman2}; the equivalence therefore
holds for the models without margins only. \texttt{raman2} verified
\ExtRamVer/\ExtDen{} with median \ExtRamMs\,ms (95th percentile
\ExtRamPninetyfiveMs\,ms) and was slower than \texttt{stl} on
\ExtRamSlower/\ExtPaired{} queries; \texttt{stl} re-run in the same session
had median \ExtSTLMs\,ms (recorded: \ExtSTLMainMs\,ms). These times describe
our implementation and this protocol, not the two encodings in general; E2
compares against the faster form. We did not implement the counter-based encoding
of~\cite{rodionova2023tecs}, which maximizes a time-robustness value instead of
certifying a fixed $D$. E2 uses no warm start in any arm.

The method \texttt{exact} (one dwell indicator $\omega_s\le\beta_u$,
$u\in[s,s+\ell]$, per start $s$, shared by all offsets, and
$\sum_{t\in[a,b]}\omega_{t-d}\ge1$ per $d$) is not an online performance
arm. It only decides,
with a 30\,s limit, which test queries are feasible (the denominator) and
supplies the references along which the library is patched.

\section{Experimental Protocol and Additional Results}\label{sec:experiment-details}

Every number, figure, table and video is generated from the formal records
of one result set (\texttt{results/main3}) through its manifest, which lists
each adopted record with its experiment object, expected library identity
and file hashes; the asset and video scripts read records only through it and
abort if any record, or the manifest's source digest, differs from the
working tree. Earlier result sets are kept as read-only archives and are
cited only where marked as historical. A record is keyed by the digest of its experiment parameters, the source
digest of the controller package, the verification identity of the library
and the identity of its certificates; changing any of them produces a new
record. Raw per-tick logs (nominal and measured states, inputs, bits, monitor
and bitset states, interface excess, candidate events, timing and the MuJoCo
trace) are kept with each record. Every formal run is reported. Development
configurations differ from the formal ones. Host: Apple M4 Max (16 logical
CPUs), macOS 26.6.2; Python 3.11.16, NumPy 2.4.6, HiGHS 1.15.1 (one thread),
MuJoCo 3.14.0, Matplotlib 3.11.2. Timings are wall-clock measurements on this
host, not bounds.

\subsection{Platform and interface}
MuJoCo executes the Crazyflie model pinned at Menagerie commit
\texttt{2cd6b7be15440787\allowbreak d12f7eb4c34ae1a1a03237c4}~\cite{menageriePinned}
with a modified actuator configuration: physics at 0.002\,s, an attitude loop
of natural frequency 50\,s$^{-1}$ and damping 1, height gains 16 and 8, body
moment limit $\pm0.006$\,N\,m (the upstream model allows $10^{-5}$\,N\,m),
thrust cap 0.35\,N and commanded acceleration clipped at $\pm2$\,m/s$^2$;
wind is a bounded random acceleration of amplitude 0.01\,m/s$^2$ evaluated at
physical time. These values are the selected configuration, validated
empirically as described below; they are not a hardware calibration, so the
platform is a simulated model, not a specific vehicle. Each actuator is a
unit-gain motor acting through one gear component, so the body thrust or
moment is the control multiplied by its gear; the controller sets the control
to the physical value divided by the gear. The limits above are control
limits (MuJoCo \texttt{ctrlrange}, $[0,0.35]$ for thrust, $\pm600$ for each
moment control at gear $-10^{-5}$); no actuator output limit is set. A
script checks this mapping on the runtime model with one positive and one
negative command per moment axis and legal thrust commands, at two
attitudes, and checks that commands outside the range are clamped. Over all
physics steps of the E1 and E3 runs, recomputed from the logged controls,
the peak thrust is \PlatThrustPct\,\% of the cap and the peak body moment
\PlatMomentMaxNm\,N\,m (\PlatMomentPct\,\% of the limit); \PlatAtLimit{}
steps reach a limit. The outer model is a
planar double integrator per axis with $h=0.2$\,s,
$\mathcal X=[0,6]\times[0,5]\times[-1,1]^2$, $\|u\|_\infty\le0.9$,
$\mathcal W$ a box with half-widths $1/125$ (position) and $1/40$ (velocity),
and $\varepsilon=1/25$. Each axis uses the gain $K=(-49/4,-231/40)$ (a double
closed-loop pole at $3/10$); the tail condition $A_K^4\mathcal W\subseteq
\frac15\mathcal W$ is checked exactly and
$\mathcal E=\frac54\bigoplus_{i<4}A_K^i\mathcal W$. Inside a sample the inner
loop tracks $v_k+Ke(\tau)$; the residual $w_k=x_{k+1}-Ax_k-Bu_k$ is measured
against $u_k=v_k+Ke_k$ at the sample boundary, so inner-loop lag and wind both
enter $w$. Every sample of every run checks $e\in\mathcal E$, $w\in\mathcal
W$, $u\in\mathcal U$ and $x\in\mathcal X$ exactly and counts the physics
steps at which the thrust or moment limits or the acceleration clip are
active, or a contact exists. A run is \emph{compliant} only if the four
memberships hold at every sample and all three counts are zero.

The interface was chosen on development routes; as empirical validation of
the selected configuration it was run on
\IfRuns{} runs (three wind seeds, seven routes) not used for tuning; all were
compliant. A one-factor sweep around a wind-free baseline varied reference
acceleration and speed, gain, zero-order hold, attitude bandwidth, moment and
acceleration limits and wind (\IfSweep{} settings, four runs each); only the
moment limit 0.0005\,N\,m broke compliance. Compliance on finitely many runs
is not a nonlinear tracking certificate.

\subsection{Scenes and libraries}
Both scenes share the platform. Library regions are translates $c\oplus T$ of
one invariant shape $T$ with law $v=g+K(z-c)$ along routes; the final region
is the post set $z^\star\oplus T$. Every region, edge, label and the post law
is verified in exact rational arithmetic. The \emph{revisit} scene has
$R_1$ centred at $(2,2.5)$ with half-width 0.3, task
$\F_{[3,20]}\G_{[0,2]}\mu_{R_1}$, $D=8$, $H=30$, an initial speed of
0.8\,m/s into $R_1$ and a zero-input history; its library has
\ETwoRegionsBefore{} regions. The \emph{branching} scene adds an obstacle
between two channels and a second region $R_2$ at $(5,2.5)$ to be reached and
held for 2 samples within $[\,\text{arrival}-4,\text{arrival}+D\,]$ of the
channel route ($H=107$, 164 regions, 166 edges).

\subsection{Costs}
Table~\ref{tab:cost} collects every timed stage. Development search is
repeated until a configuration certifies; preprocessing runs once per
library and task; online rows are per query or tick. E2 medians and tails are
over verified candidates only, with the other outcomes counted in the
denominator column. All times are wall time of one process on the host above.
R6 times are method times inside a fresh worker process, excluding library
loading, and are reported with peak memory in Table~\ref{tab:r6}.

\begin{table*}[t]
\caption{Costs, read from the formal records by \texttt{scripts/cost\_table.py}.}
\label{tab:cost}
\centering\footnotesize
\setlength{\tabcolsep}{4pt}
\begin{tabular}{@{}p{0.19\textwidth}p{0.33\textwidth}p{0.23\textwidth}p{0.19\textwidth}@{}}
\toprule
stage (scene) & timer covers & denominator & time \\
\midrule
\multicolumn{4}{@{}l}{\emph{Development search}} \\
R4 window search (quadrotor) & one recursion per window $a$, stop at the first certified & 5 windows ($a=10$--14) & 173.0\,s (28.4\,s--41.0\,s each) \\
\midrule
\multicolumn{4}{@{}l}{\emph{Fixed-instance preprocessing}} \\
R1 library (unicycle) & verify every edge on whole cells for all of $\mathcal W$ & 27\,076 edges & 7.6\,s \\
R1 recursion (unicycle) & ordered recursion, one task & 1 task & 3.4\,s \\
R4 library (quadrotor) & as R1, intermediate samples and inputs included & 224\,094 edges & 133.9\,s \\
R4 recursion (quadrotor) & ordered recursion for the chosen window $a=14$ & 423\,772 generators & 40.4\,s \\
R5 library (two-target) & as R1, robust and safety-only variants & 75\,327 edges & 39.4\,s \\
R5 recursion (two-target) & ordered recursion, two tasks & 62\,872 generators & 20.2\,s \\
E2 library (revisit) & scene, route library with exact labels, verification, certificates & 23 regions & 5.6\,s \\
E2 patching (revisit) & whole procedure: $N{=}5$ attempts, references, checks, rebuilt certificates & 12 queries, 4 patches & 63.7\,s (53.7\,s certificates) \\
E3 library (branching) & as E2 library & 164 regions & 28.5\,s \\
\midrule
\multicolumn{4}{@{}l}{\emph{Online queries}} \\
$\mathcal P_0$, $N{=}10$ (revisit) & build, solve, exact check, first complete plan & 60/60 verified & 58\,ms median, 72\,ms p95 \\
$\mathcal P_0$, $N{=}5$ (revisit) & build, solve, exact check, first complete plan & 38/60 verified (22 infeasible after tightening) & 45\,ms median, 54\,ms p95 \\
full-H (\texttt{stl}) (revisit) & build, solve, exact check & 60/60 verified & 37\,ms median, 53\,ms p95 \\
two-sided (\texttt{raman2}) (revisit) & as full-H & 60/60 verified & 176\,ms median, 287\,ms p95 \\
R3 filter tick (unicycle) & proposal, projection and filter query per tick & 60 runs (3 arms), 0 overruns of 100\,ms & 0.75\,ms max \\
\bottomrule
\end{tabular}

\end{table*}

\subsection{E1}
Exact evidence: an exact full-horizon reference verifies a plan; the
single-witness encoding is infeasible as a solver claim (no tightening); an
exact lower bound on the $x$-position over all admissible nominal inputs
proves that the position exceeds $\bar R_1$ at samples \EOneForcedOut, which
blocks every dwell start in $[a,b-D]$ and every single continuous visit.
Ten closed-loop MuJoCo runs (seeds 400--409, fixed before the runs) use the
online loop of E3 with $N=10$ and lookahead 1, adopting short candidates and
extending them by the backup rule. For each run the compact monitor and the full bitset
are evaluated tick by tick on the nominal bits, the bits of the measured
positions, and a counterfactual history with the second visit removed from
sample 13 on. The failure verdicts are compared at every sample; coverage
against the suffix $[c,D]$ and the dwell counters are compared at every
sample at which the compact monitor has not failed, since afterwards it has
no state to compare. Panel (d) of Fig.~\ref{fig:mechanism} is chosen by a
fixed rule among the adopted candidates of all ten runs: those with
$\tau<b+\ell=22$ whose predicted monitor still has $c_\tau>0$
(\EOneTailPool{} candidates), the one adopted closest to $k=8$ (seed
\EOneTailSeed). Its tail is the default chain of its validated terminal pair
(the witness the backup rule appends at every step), checked equal to the inputs the backup rule
executes, and the whole plan is validated exactly up to $H$.

\subsection{E2}
\begin{figure}[t]
\centering
\includegraphics[width=\columnwidth]{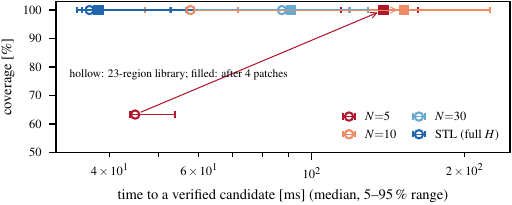}
\caption{E2: share of the \ETwoFeasible{} reference-feasible test queries
with an exactly verified candidate against median time (bars 5--95\,\%),
before and after patching. STL: full horizon $H=30$.}
\label{fig:coverage-latency}
\end{figure}

\emph{E2 results.} An exact full-horizon reference verifies
\ETwoFeasible{} of \ETwoQueries{} test queries around the start; these form
the denominator. With a 2\,s budget and identical semantics, history and
constraints, library-certified horizons $N=5,10,30$ verify \ETwoNFiveBefore,
\ETwoNTenBefore{} and \ETwoNThirtyBefore\,\% (median \ETwoNFiveBeforeMs,
\ETwoNTenBeforeMs, \ETwoNThirtyBeforeMs\,ms) and the full-horizon STL
encoding \ETwoSTLBefore\,\% (median \ETwoSTLBeforeMs\,ms;
Fig.~\ref{fig:coverage-latency}). Patching the library along references of
\ETwoPatches{} separate queries (regions $z_t\oplus T$, law $v_t+K(z-z_t)$,
exactly verified, old winning set kept; \ETwoRegionsBefore{} to
\ETwoRegionsAfter{} regions) lifts $N=5$ to \ETwoNFiveAfter\,\% and $N=10$ to
\ETwoNTenAfter\,\%, at median \ETwoNFiveAfterMs{} and \ETwoNTenAfterMs\,ms.
At $H=30$ the full-horizon encoding plans fastest; its two-sided form
(Section~\ref{sec:encoding}) verifies \ExtRamVer/\ExtDen{} at median
\ExtRamMs\,ms and is slower on every query in our implementation. With one
reach task every reachable bitset is a
suffix, so bitset and compact recursions store the same states%
\ifcomputationtable{} (Table~\ref{tab:computation})\fi; ordered generators
store fewer on the branching library (\ETwoStoredFullOrdered{} vs.\
\ETwoStoredFullBitset) in similar time. All answers agree and no recursion
exceeds \ETwoMaxDPMs\,ms. Offline, building and verifying the initial
library takes \ETwoLibraryS\,s and patching \ETwoPatchS\,s
(\ETwoPatchCertS\,s re-verifying it).

\ifcomputationtable
\begin{table}[t]
\caption{E2 recursions on identical graphs and query domains, per library
(regions, $H$): total time [ms] / stored states / peak resident memory of
the isolated process [MiB].
Fewer stored states need not mean less memory or time.}
\label{tab:computation}
\centering\footnotesize
\setlength{\tabcolsep}{2pt}
\begin{tabular}{@{}lrrr@{}}
\toprule
library & bitset & compact & ordered \\
\midrule
revisit (23, 30) & 1.0/544/53 & 0.8/544/53 & 2.2/627/53 \\
patched (143, 30) & 5.6/3546/56 & 6.3/3546/55 & 14.4/3337/54 \\
branching (164, 107) & 33.3/15742/65 & 38.2/15742/65 & 39.6/11487/59 \\
\bottomrule
\end{tabular}

\end{table}
\fi

\emph{E2 protocol.} Test queries perturb the start by at most $(3/50,1/10)$\,m in position and
$1/20$\,m/s in velocity on a rational grid (seed 401); patch queries are drawn
separately (seed 400). A reference with a 30\,s limit defines the
denominator. Every method gets one solve with a 2\,s limit and exact
validation; for the short horizons the first complete plan (candidate and
certificate rollout) must also verify. Outcomes are kept apart: verified,
timeout, infeasible only for the tightened model, solver-infeasible, rejected.
Patches are added for patch queries that $N=5$ misses, at most five; each
solves a reference with constraints tightened by $T$ and target
$z^\star\oplus\frac1{10}T$, adds $z_t\oplus T$ with law $v_t+K(z-z_t)$, verifies
the new edges exactly, and rejects the patch if any old winning state is lost.
The recursions in Table~\ref{tab:computation} use the same certificate graph
and query domain (every region at $k=0$ with a neutral history), count
forward reachability for the sparse methods, and run in isolated processes.
Offline cost is measured in the same run: constructing the revisit scene
(tube sets, post set and the route library with exact labels) and verifying
its library and certificates takes \ETwoLibraryS\,s, and
\ETwoLibraryFullS\,s for the branching scene. The patching procedure takes
\ETwoPatchS\,s in total, including the $N=5$ attempts on all patch queries,
the tightened references and the exact checks of the new regions and edges;
\ETwoPatchCertS\,s of it re-verify the whole grown library and rebuild its
certificates after each patch.

\subsection{E3}
\emph{E3 results.} A
branching scene adds a second region behind an obstacle, reached through one
of two channels ($H=107$). A soft cost prefers one channel and is replaced at
$k=\EThreeUpdateK$ by a preference for the other. All arms start from one
plan that is both a verified short candidate and a valid full-horizon plan,
and apply the same inputs until the update. \emph{Ours} re-solves
$\mathcal P_k$ ($N=10$) \EThreeLookOurs{} samples ahead and, within the same
request, appends the library tail that R4 would execute and validates the
plan exactly to $H$; \emph{full-$H$} re-solves the exact full-horizon
encoding with the executed prefix fixed and a complete warm start, with its
tuned \EThreeLookFull-sample window or with ours (solver limit tuned on
development seeds); \emph{no-solve} executes only the backup. Ours and
six-sample full-$H$ also run with the optimizer lost over samples
\EThreeOutage. Every request has a budget of \EThreeBudgetS\,s from issue,
capped by its target boundary (on average \EThreeBeffOursMs\,ms two and
\EThreeBeffFullMs\,ms six samples ahead); adopted plans reach $H$ and are
then only shifted. Seeds 400--409 use the original obstacle and 500--509 a
shifted one fixed by preset conditions before any arm ran on it; intervals
are bootstraps per geometry, stratified by revealed direction. Every run
completes with the interface intact (Table~\ref{tab:robot}); \RobotLateInputs{}
of \RobotTicks{} inputs reached the plant late (\RobotLateMs\,ms; work per
sample below \RobotWorkMs\,ms, host scheduling unbounded).

\begin{figure}[t]
\centering
\includegraphics[width=\columnwidth]{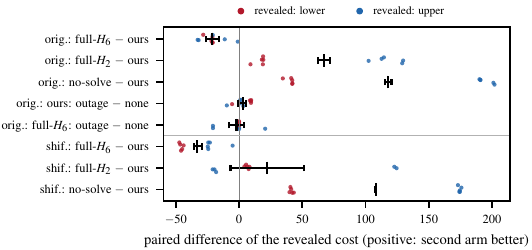}
\caption{E3: paired differences of the revealed cost per configuration
(dots, colored by the revealed channel) with mean and 95\,\% bootstrap
interval stratified by direction, per geometry; full-$H_6$, full-$H_2$: six- and
two-sample windows. Positive values favour the second arm.}
\label{fig:e3}
\end{figure}

Re-solving lowers the revealed cost relative to the backup alone by
\EThreeNoSolveOurs{} [\EThreeNoSolveOursLo, \EThreeNoSolveOursHi] (original)
and \EThreeShNoSolveOurs{} [\EThreeShNoSolveOursLo, \EThreeShNoSolveOursHi]
(shifted), and ours switches channel in every configuration whose backup
takes the other (Fig.~\ref{fig:e3}, Table~\ref{tab:e3}). A \emph{useful}
update is an adopted plan whose full revealed cost is at least
\EThreeUsefulDelta{} below the backup it replaces. Two samples ahead,
full-$H$ is worse than ours in every original configuration,
$J(\text{full-}H_2)-J(\text{ours})=\EThreeWTwoOurs$ [\EThreeWTwoOursLo,
\EThreeWTwoOursHi] (\EThreeWTwoOursUp{} when the upper channel is revealed,
\EThreeWTwoOursLow{} otherwise): both adopt a revealed-cost plan
\EThreeRespOursMin{} samples after the update, but its first useful update
comes after \EThreeUsefulWTwoMin--\EThreeUsefulWTwoMax{} samples (ours
\EThreeUsefulOursMin--\EThreeUsefulOursMax) and it never switches. On the
shifted geometry the difference is \EThreeShWTwoOurs{} [\EThreeShWTwoOursLo,
\EThreeShWTwoOursHi]. Six samples ahead full-$H$ is better on both
geometries, by \EThreeFullOurs{} [\EThreeFullOursLo, \EThreeFullOursHi] and
\EThreeShFullOurs{} [\EThreeShFullOursLo, \EThreeShFullOursHi], and updates
usefully first in \EThreeUsefulFullFirst{} and \EThreeShUsefulFullFirst{} of
10 configurations. In this scene the certificate-based controller thus
updates usefully under a two-sample deadline more reliably than full-horizon
replanning, which costs less given a longer window. Ours selects by the check alone and
does not see its tail's cost (\EThreeCostlierOursPct\,\% of its adoptions
raise the full cost; full-$H$ \EThreeCostlierFullPct\,\%); pricing the
library tail lowered the full cost by about 1 in an offline diagnosis
(below) and was not adopted, and this experiment does not separate that
mismatch from library restrictions and timing. Every arm keeps a legal
backup, which carries all \EThreeOutageTicks{} outage samples; the outage
changes the cost by \EThreeOutageOurs{} [\EThreeOutageOursLo,
\EThreeOutageOursHi] (ours) and \EThreeOutageFull{} [\EThreeOutageFullLo,
\EThreeOutageFullHi] (full-$H$).

\begin{table}[t]
\caption{E3 per arm and geometry over \EThreeConfigs{} configurations each:
mean revealed cost on the executed nominal trajectory; switches to the
revealed channel among the \EThreeSwitchable{} configurations whose backup
takes the other; range of samples from the update to the first useful update
(full revealed cost at least \EThreeUsefulDelta{} below the replaced backup).}
\label{tab:e3}
\centering\footnotesize
\setlength{\tabcolsep}{2.5pt}
\begin{tabular}{@{}lrrrrrr@{}}
\toprule
 & \multicolumn{3}{c}{original geometry} & \multicolumn{3}{c}{shifted geometry} \\
\cmidrule(lr){2-4}\cmidrule(lr){5-7}
arm & $J$ & switch & useful & $J$ & switch & useful \\
\midrule
ours & \ensuremath{-116.6} & 5/5 & 7--24 & \ensuremath{-106.7} & 5/5 & 6--31 \\
full-$H_6$ & \ensuremath{-137.7} & 5/5 & 6--15 & \ensuremath{-139.7} & 5/5 & 6--15 \\
full-$H_2$ & \ensuremath{-49.3} & 0/5 & 31--37 & \ensuremath{-84.6} & 3/5 & 8--41 \\
no-solve & \ensuremath{1.1} & 0/5 & -- & \ensuremath{1.2} & 0/5 & -- \\
\bottomrule
\end{tabular}

\end{table}

\begin{table}[!b]
\caption{Robot runs, totals (E3 over both geometries): adopted plans /
requests; samples $k=1,\dots,H-1$ from a new plan / the backup; inputs more
than 10\,ms late / all samples.}
\label{tab:robot}
\centering\footnotesize
\setlength{\tabcolsep}{1.7pt}
\begin{tabular}{@{}lrrrrr@{}}
\toprule
runs & compl. & compliant & adopted/req. & new/backup & late \\
\midrule
interface check (21) & -- & 21 & -- & -- & -- \\
E1 online (10) & 10 & 10 & 254/290 & 254/36 & 0/350 \\
E3 ours (20) & 20 & 20 & 830/1900 & 830/1290 & 0/2240 \\
E3 full-$H_6$ (20) & 20 & 20 & 422/1820 & 422/1698 & 1/2240 \\
E3 full-$H_2$ (20) & 20 & 20 & 1102/1900 & 1102/1018 & 0/2240 \\
E3 no-solve (20) & 20 & 20 & -- & 0/2120 & 0/2240 \\
E3 ours, outage (10) & 10 & 10 & 384/800 & 384/676 & 0/1120 \\
E3 full-$H_6$, outage (10) & 10 & 10 & 231/760 & 231/829 & 0/1120 \\
\bottomrule
\end{tabular}

\end{table}

\emph{E3 protocol.} Each geometry has \EThreeConfigs{} formal configurations alternating the
initial and revealed channel preference: wind seeds 400--409 on the original
obstacle and 500--509 on the shifted one, fixed before the runs; the soft
cost is $\frac1{20}\|v\|_1\pm y$. The shifted obstacle was fixed before any
arm ran on it, as the first of a short pre-listed sequence of candidate
geometries meeting four conditions written down in advance: its library and
certificates verify exactly; the common plan is both a short candidate and a
complete plan; in both directions, at the update, the other channel admits
an exact full-horizon reference through its gate and a library witness; and
the two channels are told apart by their gates. In both geometries the
common plan takes the lower channel, so only configurations revealing the
upper one require a switch. The common initial plan is the initial short
candidate, extended to $H$ by the backup rule (R4) and validated exactly as a
full plan; every arm starts from it and requests are issued only from
$k=\EThreeUpdateK$ on, so all arms apply the same inputs before the update.

\begin{figure}[t]
\centering
\includegraphics[width=\columnwidth]{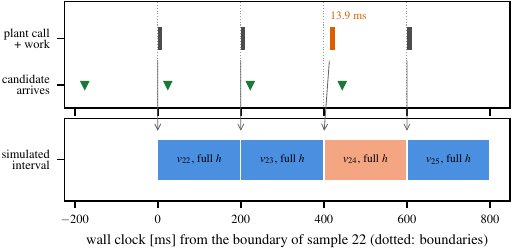}
\caption{Historical: the one late input of the archived main3-baseline records (E1, seed 306,
sample 24; historical record, archive key in
\texttt{generated/late\_tick\_timeline.json}). Top: plant call and the work
after it (nominal update, backup, next request) and the arrival of the next
candidate, in wall-clock time. The candidate for sample 24 was checked
167\,ms before its deadline, so the loop was idle and the host woke it
13.9\,ms late. Bottom: each command drives its full simulated interval
regardless of the clock reading.}
\label{fig:late-tick}
\end{figure}

\emph{Ours} solves $\mathcal P_k$ with $N=10$; in the same solver process
the validated candidate is extended along the default chain of its terminal
pair (the library tail R4 would append, identified by a digest of its edges)
and the resulting plan is validated exactly up to $H$; construction, solve,
extension and validation all count against the request's budget, and a
failed extension leaves the backup in place. \emph{Full-$H$} solves the exact
full-horizon encoding with the executed prefix fixed, warm-started with the
stored plan and all its auxiliary binaries, and its result is validated as an
online prefix to $H$. Every adopted plan therefore reaches $H$, becomes the
backup and is then only shifted. All arms share the backup rule, the exact
check and a budget of \EThreeBudgetS\,s per request from its issue (deadline
$\min\{\text{issue}+B,\ \text{boundary}-10\,\text{ms}\}$, the solver stops
30\,ms before it). Each request is logged at issue with its target sample,
boundary, deadline, effective budget $B_{\mathrm{eff}}$ and cost version, and
ends in exactly one recorded outcome. Lookahead and solver limit were tuned
per arm on development seeds and frozen: ours \EThreeLookOurs{} samples and
\EThreeLimitOursMs\,ms; full-$H$ \EThreeLookFull{} samples and
\EThreeLimitFullMs\,ms; and full-$H$ with ours' window, \EThreeLookWTwo{}
samples, with the limit chosen among 150, 250 and 350\,ms on development
seeds 200--203 by mean revealed cost (\EThreeLimitWTwoMs\,ms; longer limits
let fewer results arrive in time). $B_{\mathrm{eff}}$ averages
\EThreeBeffOursMs\,ms for the two-sample windows and \EThreeBeffFullMs\,ms
for the six-sample one. Period 0.2\,s, cost update at $k=\EThreeUpdateK$,
outage over samples \EThreeOutage, five post-mission samples; the arms of a
configuration run in a random order, and the wind is a function of physical
time, so its value at each sample start is equal across arms (checked). The
solver process skips requests made obsolete by a newer request or by an
adoption. A request issued during sample $k$ uses the cost known at $k$, so
the revealed cost first enters the request issued at $k=\EThreeUpdateK$; each
result carries a unique identifier, the cost version and the issue sample of
its request (outcomes on the original geometry, ours:
\EThreeOutcomesOurs; full-$H$: \EThreeOutcomesFull; full-$H$, two-sample
window: \EThreeOutcomesWTwo; \emph{superseded}: made obsolete by an adoption,
\emph{skipped in queue}: replaced by a newer request, \emph{stale}: returned
after its target sample, \emph{over budget}: its check ended after issue
plus $B$). A result is \emph{ready} if its target is the current sample, its
$(z,q)$ equal the executed nominal and monitor states exactly, the inputs it
assumed equal the executed nominal inputs $v$ (not $u$), its validation is
bound to the same $(k,z,q)$, and this check ends before the guard; a result
whose check ends later is discarded as late (\RobotLateCandidates{} in the
reported runs). The \emph{first version} is the first applied plan whose
request carried the revealed cost; the \emph{first useful update} is the
first such plan whose exact full revealed cost is at least
$\delta=\EThreeUsefulDelta$ below that of the backup it replaced at the same
$(k,z,q)$ ($\delta$ fixed before the runs; both are plans to $H$). The
command for a sample is decided before its boundary; at the boundary the
clock is read and the simulator is advanced with that command with no
computation in between, and the nominal update, the backup for the next
sample and the next request follow. A late input is one for which this clock
reading is more than 10\,ms after the boundary. The loop is a
wall-clock-paced simulation with measured dispatch latency: simulated time
does not advance while the host waits, and a late command still drives the
full sample interval, so holding the old input during the delay is not
simulated (Fig.~\ref{fig:late-tick}, a late input of an earlier, archived
run). Lateness therefore measures host scheduling, not an executed delay,
and no robustness to input delay is claimed.

\emph{Tail cost (diagnosis, not adopted).} The short objective does not see
the cost of the library tail. An experimental objective adds, per terminal
option, the exact cost of its default chain and of up to two chains with
another stored first witness, through an epigraph variable with a common
lower bound; the selected chain is the one extended and validated, so its
recursive feasibility argument is unchanged. Before any formal run, an
offline diagnosis re-solved, with both objectives and a 5\,s limit, all
states of the archived formal E3 runs (ours and full-$H$) in the ten samples
after the update, rebuilt exactly from the recorded adoptions (220 states;
all verified under both objectives). The tail-priced plans had an exact full
cost lower by 1.1 on average (median 0.9, at most 4.8); the threshold set in
advance for a development ablation, at least 5 on average in most
configurations, was met in none of 10, so the objective was not adopted and
the formal runs use the short objective. The code, tests and diagnosis are
kept with the records.

The revealed cost is evaluated on the executed nominal trajectory from
$k=\EThreeUpdateK$ to $H$ (the planners' objective; the same stage cost on the
applied inputs and measured states gives the same signs); intervals are
percentile bootstraps over the configurations of one geometry, resampled
within each revealed direction, and per-direction means are reported
alongside. Selection is by the exact check alone: on the original geometry
\EThreeCostlierOurs{} of the \EThreeCostlierOursOf{} plans adopted by ours
(\EThreeCostlierOursPct\,\%) had an exact full cost above that of the backup
they replaced at the same $(k,z,q)$, against \EThreeCostlierFull{} of
\EThreeCostlierFullOf{} for full-$H$ and \EThreeCostlierWTwo{} of
\EThreeCostlierWTwoOf{} for its two-sample variant (shifted: ours
\EThreeShCostlierOurs{} of \EThreeShCostlierOursOf), so no decrease of the
remaining cost is claimed.

\subsection{R1--R3 protocol}\label{sec:robust-protocol}

\begin{figure}[t]
\centering
\includegraphics[width=\columnwidth]{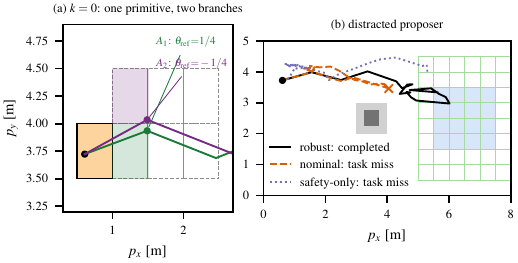}
\caption{(a) R1 gate~3: two admissible disturbances put the successor of
\ROneEvidencePrim{} in different cells with disjoint feasible sets $A_1$,
$A_2$. (b) R3 distracted proposer (cross: nominal certificate lost); blue:
$\bar R$, green: $\Kpost$ (projected); grey: obstacle and clearance.}
\label{fig:branching}
\end{figure}

\emph{Code and identities.} The robust experiments live in a separate
package (\texttt{src/mjctrl/robust}); the main3 code is unchanged. Each record
carries a verification identity (digest of model, $\mathcal W$, grid,
obstacles, clearance, primitives, tightened regions, task, subdivision depths,
$Z_{\rm post}$ and the verifier sources), a recursion identity (verification
identity, task, algorithm, recursion sources), a source digest of the robust
package, and the full parsed experiment and scene configuration with its
digest, which also enters every run key. The assets script refuses records
whose source digest or configuration snapshot differs from the working tree.

\emph{Verifier.} Endpoints are rationals; products and transcendental values
are rounded outward to dyadics with 64 fractional bits; $\sin$ and $\cos$ at
rational points use Taylor polynomials with Lagrange remainder, and over an
interval the hull of the endpoint enclosures plus the extreme value whenever
a critical point (located with rational bounds on $\pi$) may lie inside. The
heading update $(1-hk_\theta)\theta+hk_\theta\theta_{\rm ref}$ is evaluated in
this affine form to avoid the dependency between $\theta$ and $\omega$. Unit
tests compare the enclosures with sampled values, reproduce the $1-x^2$
counterexample, check a disturbance-induced branching example, match sampled
successors of randomly chosen verified edges to branches, check that a
single position box is rejected for a pure heading law and a position law,
and run the $\Kpost$ switching law for 100 steps from 20 sampled states.

\emph{Scenes.} Obstacles are closed boxes, and a region is admissible only if
its closed cell keeps $\ell_\infty$ distance at least $\tfrac14$\,m from every
obstacle, so a cell touching an obstacle is never admissible; $\RA$ is
computed from this test. Geometry A has the obstacle
$[\tfrac{13}4,\tfrac{15}4]\times[\tfrac94,\tfrac{11}4]$; geometry B the
obstacle $[\tfrac94,\tfrac{11}4]^2$ and a block $[\tfrac{21}4,\tfrac{23}4]
\times[\tfrac{19}4,5]$ at the top edge. With the clearance they cover
$[3,4]\times[2,3]$, respectively $[2,3]\times[2,3]$ and $[5,6]\times[4.5,5]$.
A cell may touch the boundary of such a band but not its interior; for
example $[2.5,3]\times[1.5,2]$ keeps distance exactly $\tfrac14$ from the
obstacle of geometry A. Declaring the bands themselves as closed obstacles
with the same clearance would leave the declared start uncertified for every
window with $a\le39$ in either geometry. Headings stay in
$[-\tfrac98,\tfrac98]$, and on the verified library with the widened band the
recursion finds no certified passage into the winning set; this is a result
on this library, not a reachability claim for the continuous system. The negative-time history is
checked against the same predicate. Both use $\bar R=[5.5,7.5]\times[1.5,3.5]$, $Z_{\rm post}=[5,8]\times
[0.5,4.5]$ and the start $(\tfrac18,\tfrac{18}5,\tfrac14)$ propagated by two
recorded history steps with $v=\tfrac14$, $\theta_{\rm ref}=\tfrac14$, $w=0$.
The window is the earliest certifiable one: $[13,16]$ for A and $[11,14]$ for
B ($H=19$, $17$). A start aligned with the obstacle row is not certifiable
for any tested window, because a committed heading must serve all three
possible next rows; the start therefore lies in the upper lane.

\emph{R1 search.} Gate~3 searches exactly reachable box states. It starts
from the rational enclosure $X_0$ of the history endpoint (no float
conversion) and extends 40 rollouts: at each step it picks a random
completion-preserving edge and a branch cell, steers a rational disturbance to
the midpoint of $\mathcal W\cap(P_{\rho_b}-F_e(X_k))$, with $F_e(X_k)$ the
interval enclosure of $f(x,\kappa_e(x),0)$ over $X_k$, and keeps the
successor enclosure $X_{k+1}$ if it lies strictly inside $P_{\rho_b}$ with
decided bits and a winning successor state. At every $(k,X_k,q_k)$ with
$k\le H-2$ and every completion-preserving edge, it applies the same steering
to each branch cell, requires a strictly positive margin and decided bits for
the whole box, advances the monitor on those bits and enumerates all outgoing
edges at $k+1$ with the all-successor check. The evidence is therefore valid
for every state in $X_k$, which contains the actual state under the recorded
history and disturbances. Primitive identities are digests of the law
parameters, independent of the source cell.

\emph{R3 protocol.} The configuration file (2 geometries $\times$ 10 seeds,
proposer parameters, disturbance law, budget and the four targeted cases)
was frozen before any arm ran. Each coordinate of $w_k$ is a vertex of
$\mathcal W$ with probability $1/2$ and uniform otherwise, drawn from the seed
independently of the actions. The proposer is a 4-iteration CEM (population
48, 8 elites) over 5-step $(v,\theta_{\rm ref})$ sequences on the nominal
model, with an approach cost before $a$, a window cost on $\bar R$ shrunk by
0.5\,m, obstacle and boundary penalties, and, after completion, attraction to
the centre of $Z_{\rm post}$; the safety-only arm uses the same cost. The
tick deadline covers proposal, projection and filtering and is checked after
the final selection, before execution; on an overrun the certified witness
edge is executed instead. In the targeted \emph{distracted} case
(Fig.~\ref{fig:branching}b), the
proposer's task objective is replaced for the entire mission (ticks
$0$--$H-1$) by attraction to the safe waypoint $(1,4.2)$ far from $\bar R$
(obstacle and boundary penalties unchanged); the case tests the filter against a proposer that never pursues the task,
not the shared proposer. Only the robust arm completes, after
\RThreeDistPhys{} interventions on physically safe proposals. The one-step filter acts only from certified
states: at $q_0$, \RTwoFullCertified{} of the \RAdmRegions{} admissible
regions are certified, and from any other state it has no completion-preserving
action to select. Every tick logs the raw proposal, the projected primitive, the executed primitive,
the matched branch and the actual bits. Outcomes are classified as physical
violation, task miss, post-set violation or completed; certificate loss is
recorded separately. Completability is lost at the first tick (or at $H$) at
which no region containing the actual state is winning, at the actual monitor
state, in the robust recursion of the full library. The same criterion and
robust winning family are used for all three arms. The loss time is evaluated
separately on each arm's actual trajectory.

\begin{table}[t]
\caption{R3, \RThreeConfigs{} frozen configurations, all reported.}
\label{tab:r3}
\centering\footnotesize
\setlength{\tabcolsep}{3.2pt}
\begin{tabular}{@{}llrrrrrrr@{}}
\toprule
geom. & arm & done & safe & cert. & compl. & interv. & acc. & $J_{\rm path}$ \\
 & & & & lost & lost & & \% & m \\
\midrule
A & robust & 10/10 & 10/10 & 0 & 0 & 59 & 69 & 9.5 \\
A & nominal & 10/10 & 10/10 & 2 & 5 & 6 & 97 & 9.4 \\
A & safety-only & 10/10 & 10/10 & 0 & 4 & 12 & 94 & 9.4 \\
B & robust & 10/10 & 10/10 & 0 & 0 & 41 & 76 & 8.9 \\
B & nominal & 10/10 & 10/10 & 3 & 1 & 15 & 91 & 8.9 \\
B & safety-only & 10/10 & 10/10 & 0 & 4 & 8 & 95 & 8.8 \\
\bottomrule
\end{tabular}

\end{table}

Table~\ref{tab:r3} also reports
the acceptance rate (ticks on which the executed primitive is the projected
proposal), and the logs count branch-dependent ticks (the complete feasible
sets at $k+1$ differ across the executed edge's branches: \RThreeBranchTicks{}
for the robust arm) and the path length $J_{\rm path}$ over all $H$ ticks,
from which no cost claim is made. In the distracted case completability is
lost at $k=\RThreeDistComplNominal$ (nominal) and $k=\RThreeDistComplSafety$
(safety-only), before the nominal certificate is lost at
$k=\RThreeDistLostK$. In the other three targeted cases, vertex disturbances
leave every nominal branch at $k=\RThreeUncLostK$; replaying the gate-3
disturbances changes the executed primitive at $k=1$; and through an
optimizer outage on ticks 3--8 the robust arm completes on
\RThreeOutFallback{} certified fallbacks. The targeted cases are written
through the record store with run keys.

\emph{R2 details.} Table~\ref{tab:r2} varies $D$ and adds a second obligation
on the same region ($a=14$, $b=16$, $\ell=2$). Compact states grow linearly
in $D$ per task and explicit ones exponentially; antichains stay small. The
smaller representation does not pay off in resources here: with one task and
$D=2$ the ordered recursion takes \RTwoOrdS\,s and \RTwoOrdMB\,MB peak against
\RTwoCompS\,s and \RTwoCompMB\,MB for the compact table (two tasks:
\RTwoOrdTwoS\,s, \RTwoOrdTwoMB\,MB against \RTwoCompTwoS\,s,
\RTwoCompTwoMB\,MB). The ordered implementation stores generator lists and
witnesses per entry; Section~\ref{sec:r6-protocol} measures and profiles the
difference on matched queries. On
the same partition, the single-successor sub-library has \RTwoSingleEdges{}
edges: every verified one-step enclosure is wider than a cell, so each edge
has \RTwoBranchHist{} branches, and even the $w=0$ library has
\RTwoNominalSingle{} single-successor edges. Corollary~\ref{cor:partA} makes
this independent of the grid: with heading primitives and product-box
regions, $2\bar wa=\PartLHS\ge\PartRbarWidth$\,m rules out a single-successor
certificate on every partition; position-feedback primitives and non-product
regions are outside that statement.

\begin{table}[t]
\caption{R2 on the R1 graph: compact and explicit state counts, stored
generators, largest antichain, evaluated label combinations and the naive
number of target vectors $\sum\prod_b|M_{k+1}(\rho_b)|$. All three recursions
agree in every row. Bottom: untraced wall time and tracemalloc peak (separate
run) for the traced rows.}
\label{tab:r2}
\centering\footnotesize
\begin{tabular}{@{}rrrrrrrr@{}}
\toprule
$n_r$ & $D$ & $|\mathcal Q|$ & bitset & gen. & width & combos & naive \\
\midrule
1 & 1 & 9 & 12 & 19348 & 2 & 306122 & 5.6e5 \\
1 & 2 & 12 & 24 & 19478 & 2 & 308780 & 6.9e5 \\
1 & 4 & 18 & 96 & 19738 & 2 & 314096 & 9.6e5 \\
1 & 6 & 24 & 384 & 19998 & 2 & 319412 & 1.2e6 \\
2 & 1 & 108 & 192 & 20202 & 6 & 327408 & 5.4e8 \\
2 & 2 & 192 & 768 & 20820 & 6 & 343222 & 1.1e9 \\
2 & 3 & 300 & 3072 & 21438 & 6 & 359216 & 1.6e9 \\
\bottomrule
\end{tabular}

\smallskip

\begin{tabular}{@{}rrrrrrrr@{}}
\toprule
& & \multicolumn{2}{c}{compact} & \multicolumn{2}{c}{bitset} & \multicolumn{2}{c}{ordered} \\
$n_r$ & $D$ & s & MB & s & MB & s & MB \\
\midrule
1 & 2 & 3.21 & 4.6 & 3.24 & 4.9 & 3.52 & 39.8 \\
2 & 2 & 3.9 & 9.7 & 6.26 & 26.1 & 6.04 & 42.6 \\
\bottomrule
\end{tabular}

\end{table}

\subsection{R4 protocol: branching policy on the quadrotor}\label{sec:r4-protocol}
\emph{Model.} The certified system is the platform's declared outer model
(Section~\ref{sec:eval}) per axis, $x_{j+1}=Ax_j+Bu_j+w_j$ with $h=\tfrac15$\,s,
$|w_p|\le\tfrac1{125}$\,m, $|w_\nu|\le\tfrac1{40}$\,m/s and $|u|\le\tfrac9{10}$,
grouped into a macro tick of 5 samples (1\,s). All task times ($a$, $b$,
$\ell$, $D$, $H$) are macro ticks, and target bits are read only at macro
boundaries. A primitive is a velocity reference $v_{\rm ref}\in\{0,\pm\tfrac14,
\pm\tfrac12\}^2$ (25 primitives) applied at every sample as
$u_j=\tfrac94(v_{\rm ref}-\nu_j)$ on the measured velocity. The macro map is
the exact composition of the 5 steps; its disturbance enclosure is the exact
interval hull of the sum of the 5 propagated per-sample boxes. An edge is
verified on its whole source cell and for all disturbances if the inputs
satisfy $|u_j|\le\tfrac9{10}$ at all 5 samples, every intermediate sample keeps
the clearance $\tfrac1{10}$\,m from the closed obstacle
$[3,3.8]\times[1.5,3.5]$ and stays in the workspace $[0,6]\times[0,5]$
(safety between macro boundaries is thus an additional verified property of
each edge), and the successor enclosure is covered by present cells with
conservative labels, as in R1. Regions are product cells of
$\tfrac18$\,m in position and $\tfrac14$\,m/s in velocity over
$[-\tfrac58,\tfrac58]^2$. The task is $\F_{[a,a+3]}\G_{[0,1]}\mu_R$ with
$\bar R=[4.5,5.5]\times[2,3]$, $D=\RFourD$ and $H=a+7$; $\Kpost$ is the safety
fixed point inside $Z_{\rm post}=[4.25,5.75]\times[1.75,3.25]$ (all velocity
cells), \RFourKpost{} of \RFourKpostCand{} candidates. The deadline was
searched on the model before any robot run from the commanded start
$(\tfrac{33}{16},\tfrac{41}{16},0,0)$ with an all-zero history:
$a=\RFourSearchLo,\dots,\RFourSearchFail$ are not certified, $a=\RFourA$ is
($H=\RFourH$). Library construction takes \RFourBuildS\,s and the ordered
recursion \RFourDPs\,s.

\emph{Design log.} The planned $0.5$\,m cells were replaced before any robot
run; this is the one refinement we allowed. On $0.5$\,m cells the macro
position image of a cell overlaps the cell itself even at full speed (about
$[0.36,1.07]$ from $[0,0.5]$), so every moving region keeps a branch back to
itself. With $\tfrac14$\,m cells (we tried gains $\tfrac54$ and $\tfrac94$,
hold laws and a braking gain), the abstraction admitted neither a robust stop
inside $\bar R$ nor entry into the channels beside the obstacle. With
$\tfrac18$\,m cells the macro displacement at $|v_{\rm ref}|\ge\tfrac14$
exceeds one cell on the whole source cell. The refinement was chosen by
certification on the model alone.

\emph{Execution.} Each fine sample $j$ of macro tick $k$ calls the unchanged
robot interface with the measured outer state $z=x_j$, the input $u_j$, the
tube gain $K$ and the global sample index $5k+j$, so $e_j=0$ at every boundary
and the residual $w_j=x_{j+1}-Ax_j-Bu_j$ is the quantity the platform
evidence already reports. Before the mission the vehicle hovers under
$v_{\rm ref}=0$ for $D$ macro ticks from the commanded start; the measured
states give the history bits and $q_0$, each is checked against the safety
predicate, $x_0$ must lie strictly inside one cell $\rho_0$, and gate~2 is
$\Win_0(\rho_0,q_0)$. The policy executes witness edges of the ordered
recursion, matches the branch on the measured macro successor and advances
the monitor on the actual bits; after $H$ it executes the $\Kpost$ witness
edges for \RFourPostTicks{} further macro ticks and checks residence in
$\Kpost$. A run counts as guaranteed only if gate~2 holds, every successor
up to $H$ and during the post ticks matches a certified branch, and every
residual lies in $\mathcal W$ (no slack). The 10 wind seeds were fixed in the configuration
before any run; the record keys include the MuJoCo version (\RFourMujoco), the
scene hash and the upstream model hash.

\begin{figure*}[t]
\centering
\includegraphics[width=\textwidth]{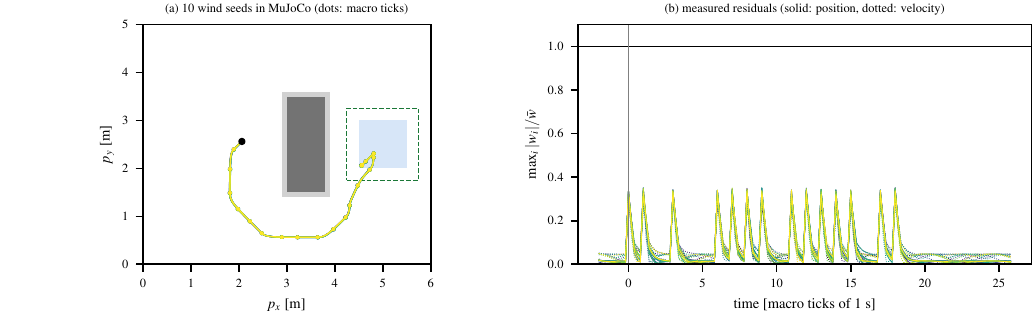}
\caption{R4 in MuJoCo. (a) Measured paths of the \RFourSeeds{} wind seeds
(dots: macro boundaries) from the hover start (black) around the obstacle
(grey, clearance band light grey) into $\bar R$ (blue); dashed:
$Z_{\rm post}$. (b) Largest measured residual per sample relative to the
declared bound (line at 1); negative times are the hover pre-run.}
\label{fig:r4}
\end{figure*}

\emph{Results.} Of the \RFourSeeds{} seeds, \RFourGateTwo{} pass gate~2,
\RFourCompleted{} complete, \RFourSafe{} are safe at every sample with no
obstacle contact, \RFourPost{} stay in $\Kpost$ and \RFourResidualOK{} keep
every residual in $\mathcal W$; the largest residual is \RFourMaxWp\,\% of
$\bar w_p$ and \RFourMaxWnu\,\% of $\bar w_\nu$, with peaks at changes of
$v_{\rm ref}$ (Fig.~\ref{fig:r4}b). The executed edges have up to 16 branches
and the feasible sets at $k+1$ differ across branches at \RFourBDTicks{} of
\RFourBDOf{} executed ticks, but the hover pre-run ends within a few
millimetres for all seeds and every seed realized the same branch at each of
the \RFourKeys{} keys $(k,\rho,q,e)$ (Fig.~\ref{fig:r4}a). The runs therefore
show that the robot executed the branching policy, not that successors or
continuations differed between seeds.

\emph{Paired witness.} We searched, on the declared fine-sample model, for two
legal 5-sample disturbance sequences (every $w_i\in\mathcal W$, propagated
exactly with the feedback evaluated on the propagated states) from the
measured $x_0$, taken as an exact rational, that end strictly inside
different cells with decided bits and nonempty disjoint complete feasible
sets: first at $k=0$ over all winning edges, then along the executed edges
with a common exactly steered prefix. None was found from $x_0$
(witness found: \RFourFineFound); on the region graph alone, the executed edge at
$k=\RFourWitnessK$ has two abstract branches with nonempty disjoint certified
action sets at $k+1$.

The second search starts from measured decision states: from the measured
states at all \RPairPoints{} recorded decision points of the \RFourSeeds{} seeds,
\RPairCands{} candidate pairs of branches were classified exactly on the same
fine-step outer dynamics (support-function margins on rational zonogons, no
floating-point LP): \RPairUnreach{} are unreachable, \RPairBoundary{} reach
only a cell boundary, \RPairUnknown{} are undecided, and \RPairWit{} are
witnesses, i.e., two legal disturbance sequences from one measured state end
strictly inside two cells whose certified next-action sets are nonempty and
disjoint. All witnesses lie at $k=\RPairWitK$ on \RPairWitSeeds{} seeds and
use the same certified primitive \RPairPrim, which the original runs did not
execute; the margins inside the target cells are
\RPairMarginMinMM--\RPairMarginMaxMM\,mm. The two searches do not conflict:
the first starts at $x_0$, the second from measured decision states, and all
\RPairWit{} witnesses occur at $k=\RPairWitK$.

Each witness side was then replayed once in MuJoCo from the recorded
snapshot with a frozen gust (\RPairAttempts{} sides): in all \RPairLandDistinct{}
pairs the two trajectories reach different certified regions, but no side
ends strictly inside its target cell (misses \RPairMissMinMM--\RPairMissMaxMM\,mm),
and in all \RPairLandNested{} pairs the certified next-action sets of the two
landing regions are strictly nested, so \RPairLandDisjoint{} pairs are
disjoint. All \RPairCompleted{} continuations complete with every residual in
$\mathcal W$. The outer model therefore admits successors from an actual
state that require incompatible certified actions; the MuJoCo replays did not
reproduce such a pair.

\subsection{R5 protocol: same state, different histories}\label{sec:r5-protocol}
\emph{Scene.} Model, $\mathcal W$, primitives and cell sizes are those of R1;
the workspace is $[0,12]\times[0,7.5]$\,m without obstacles
(\RFiveRegions{} regions, \RFiveEdges{} verified edges). The two tasks are
$\F_{[\RFiveA,\RFiveB]}\G_{[0,\RFiveEll]}\mu_{R_j}$ with
$\bar R_{\rm A}=[3,11]\times[5,7]$ and $\bar R_{\rm B}=[3,11]\times
[\tfrac12,\tfrac52]$, $D=\RFiveD$, $H=\RFiveH$, and $\Kpost$ is the fixed
point inside $Z_{\rm post}=[10.5,12]\times[0.5,7]$ (\RFiveKpost{} regions).
The scene is mirror-symmetric about $p_y=\tfrac{15}4$. The targets are 2\,m
wide because a whole-cell successor enclosure is about 0.8\,m wide; with
1\,m targets no robust two-sample dwell was certified. Plain, explicit-bitset
and ordered recursions agree at every $(k,\rho,q)$ of the formal domain
(\RFivePlainStates{} compact and \RFiveBitsetStates{} bitset states).

\emph{Histories.} Three prefixes start at the declared
$x_0=(\tfrac34,\tfrac{15}4,0)$, reached by a recorded zero-disturbance
history, and end at $x^\star=\RFiveXstar$ at $k=\RFiveKstar$ in the region
$\rho^\star=[6,6.5]\times[3.5,4]\times[-\tfrac18,\tfrac18]$, which contains
$x^\star$ with margin \RFiveMargin{} (Fig.~\ref{fig:r5}). A visits only
$\bar R_{\rm A}$; B, its mirror image, only $\bar R_{\rm B}$; AB both. Each
step applies one library primitive and is enclosed in rational interval
arithmetic; inputs, the safety predicate and decided bits are checked on
every enclosure. Every step but the last uses a rational disturbance in
$\mathcal W$. The last disturbance is the exact residual
$w^\star=x^\star-f(x_{\rm prev},\kappa_e(x_{\rm prev}),0)$ of the additive
model for the actual $x_{\rm prev}$, so every prefix ends exactly at
$x^\star$ in every state component. It need not be rational and is never
rounded; $w^\star\in\mathcal W$ holds because its rational interval
enclosure lies in $\mathcal W$ (largest
component \RFiveWlastA\,m for A and B, \RFiveWlastAB\,m for AB). At $x^\star$ the clock,
region, bits and dwell counters coincide, and $q_{\rm A}=((0,0),(3,0))$,
$q_{\rm B}=((3,0),(0,0))$, $q_{\rm AB}=((0,0),(0,0))$. The prefixes are legal
but not certified throughout: A and B are certified from $k=\RFiveCertFromA$,
AB from $k=\RFiveCertFromAB$.

\emph{Certified sets.} Enumerating every outgoing edge of $\rho^\star$ at
$k=\RFiveKstar$, $q_{\rm A}$ certifies \RFiveSetA{} and $q_{\rm B}$
\RFiveSetB{} (primitives written $(v,\theta_{\rm ref})$): both nonempty,
disjoint. $q_{\rm AB}$ certifies all \RFiveNprims{} primitives, a strict
superset; each of the \RFiveDiffN{} actions in the difference has
\RFiveDiffFailMin--\RFiveDiffFailMax{} branches whose successor state is not
winning under $q_{\rm A}$. The reset state $((3,0),(3,0))$ lies below
$q_{\rm A}$ and $q_{\rm B}$, so its certified set is contained in their
intersection and is empty. Any rule that maps $(k,x)$ to an action takes the
same action at $x^\star$ after A and after B and therefore leaves the
certified set of at least one; this is a statement about certificates on
this library, not about every controller.

\emph{Filter and continuation.} All arms share one task-unaware proposer
(speed $\tfrac12$, heading towards the centre of $Z_{\rm post}$ clipped to
$[-1,1]$), one filter rule (keep the projected proposal if it is certified,
else the nearest certified primitive) and seeded disturbance sequences
drawn as in R3 (\RFiveSeeds{} seeds, fixed in the configuration). At $x^\star$ the
proposal projects to \RFiveProposal; the filter executes \RFiveDecA{} after A,
\RFiveDecB{} after B and \RFiveDecAB{} after AB. The \emph{all-monitor-memory
reset} controller re-initialises coverage and dwell of both tasks at
$x^\star$ but keeps the task clock, so windows and expiry are unchanged;
without a completion certificate it filters with the robust safety-only
certificate and executes \RFiveDecReset. Task outcomes are judged by an
independent monitor that is never reset and has read every actual bit since
the declared start. With memory, \RFiveMemCompleted/\RFiveMemRuns{} runs
complete, keep the certificate, stay safe and remain in $\Kpost$ for 5 ticks
after $H$. The reset controller loses its certificate at $k=\RFiveResetLostK$
in every run; the independent monitor records completion in
\RFiveResetCompA/\RFiveSeeds{} runs after A, \RFiveResetCompB/\RFiveSeeds{}
after B and \RFiveResetCompAB/\RFiveSeeds{} after AB, whose tasks were already
complete. All reset runs are safe and end in $\Kpost$. These failures depend
on the proposer and the fallback; they do not show that no controller
without memory can complete.

\emph{Partial coverage.} On the unchanged R1 library and task
($\F_{[\RFivePA,\RFivePB]}\G_{[0,\RFivePEll]}\mu_R$, $D=\ROneD$, $H=\ROneH$,
the same declared start and recorded history), two prefixes P and Q end
exactly at $x^\star=\RFivePXstar$ at $k=\RFivePKstar$, inside one region
with margin \RFivePMargin. Both have visited $\bar R$ and are outside it at
sample \RFivePKstar{} with dwell counter 0. P leaves one sample earlier, so
its last completed dwell starts at $s=\RFivePDwellP$ and serves offsets
\RFivePServedP{} ($c=\RFivePCP$); Q's starts at $s=\RFivePDwellQ$ and serves
\RFivePServedQ{} ($c=\RFivePCQ$). P's earlier witness serves no offset, and
Q's earlier witness at $s=\RFivePDwellP$ serves \RFivePServedP{}, already
inside its last set. Every step
but the last uses zero disturbance; the last is the exact residual, inside
$\mathcal W$ by its interval enclosure (largest component \RFivePWlastP\,m
for P, \RFivePWlastQ\,m for Q). The explicit bitset monitor, run on the same
bits, gives the same served sets, and from $k=\RFivePCertFrom$ onward every
region--monitor state reached along either prefix is winning. Enumerating every outgoing edge at $x^\star$, the
compact monitor on the ordered recursion and the bitset monitor on its table
both give \RFivePSetP{} after P and \RFivePSetQ{} after Q. The order
guarantees only the inclusion; the complete enumeration shows that here P's
set is nonempty and the inclusion strict. The action \RFivePProposal{} in the
difference has \RFivePDiffFail{} branches with the conservative label
$\sigma=0$ at $k=\RFivePKstar+1$ (not certified inside $\bar R$, which does
not place the actual successor outside); through them the library recursion
cannot certify offset 1 by its deadline $b-1+\ell=16$, which Q has already
met. The R5
proposer projects to \RFivePProposal{} at $x^\star$; the filter keeps it after
Q and executes \RFivePDecP{} after P. With memory,
\RFivePMemCompleted/\RFivePMemRuns{} seeded continuations complete, keep
the certificate, stay safe and remain in $\Kpost$; the reset state is not
certified at $x^\star$.

\subsection{R6 protocol: what the compact and ordered states cost}\label{sec:r6-protocol}
\emph{Scene and settings.} Model, $\mathcal W$, primitives, cells and
workspace are those of R1, without the obstacle (\RSixRegions{} regions,
\RSixEdges{} verified edges, $H=\RSixH$). There are four tasks
$\F_{[2,26]}\G_{[0,1]}\mu_{R_j}$ with upper targets
$[\tfrac32,\tfrac72]\times[2,\tfrac92]$, $[\tfrac92,\tfrac{13}2]\times
[2,\tfrac92]$ and lower targets $[\tfrac52,\tfrac92]\times[\tfrac12,3]$,
$[\tfrac{11}2,\tfrac{15}2]\times[\tfrac12,3]$; upper and lower targets overlap
in $p_y\in[2,3]$, so one visit can serve two tasks. $\Kpost$ is the fixed
point inside $[6,8]\times[\tfrac12,\tfrac92]$. A setting $(n,D)$ keeps the
first $n$ tasks, projects the labels on them (branches that then coincide
are merged) and sets the offset bound to $D$. The sweeps vary $n=1,\dots,4$
at $D=4$ and $D\in\{2,4,8,16\}$ at $n=2$, with $(1,2)$, $(3,8)$, $(4,8)$
and $(3,16)$ added. The code and the scene were developed on the
\RSixDevSettings{} settings marked \emph{dev}; the remaining ones were run
once after the configuration was frozen.

\emph{Queries and methods.} Every method answers the same queries: for each
region $\rho$, whether $(0,\rho,\Gamma(h))$ is winning, where $h$ has no
target bit on $-D,\dots,0$. The two \emph{sparse} methods traverse forward
from the queried states over every edge and branch to the states reachable
on the graph and evaluate the robust AND/OR recursion backward on these
only, with explicit $(D+1)$-bit masks (sparse bitset) or the compact pair
$(c,r)$ (sparse compact). The \emph{ordered} method computes minimal
generators and witnesses on the whole compact domain at every $k$
(Theorem~\ref{thm:robcert}) and answers by lookup. Two auxiliary methods
tabulate every monitor state at every $(k,\rho)$, in compact form (full
compact) or as bitsets (full bitset); they show the cost of whole domains
and are not part of the matched comparison.

\begin{figure*}[t]
\centering
\includegraphics[width=\textwidth]{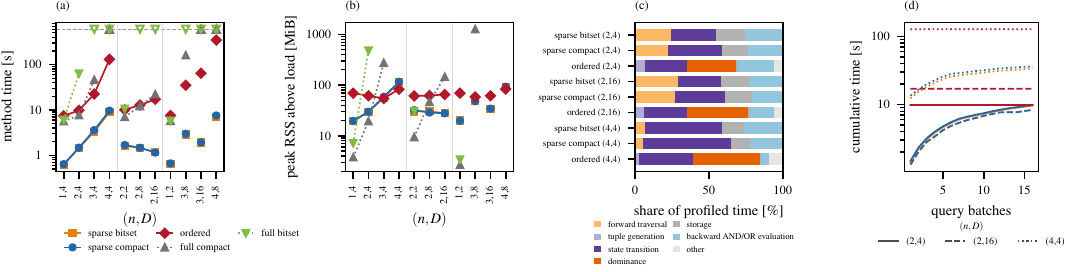}
\caption{R6. (a) Method time per setting $(n,D)$ (open markers: stopped at
the budget, dashed line). (b) Peak resident memory above the load-only run.
(c) Share of profiled time by category (separate cProfile runs). (d)
Cumulative time over the online query batches, ordered (red) against sparse
bitset (orange) and sparse compact (blue).}
\label{fig:r6}
\end{figure*}

\emph{Measurement.} Each run is a fresh worker process, the only child of a
launcher. The worker loads the pickled library, then times the method alone;
peak resident memory is the launcher's
\texttt{getrusage(RUSAGE\_CHILDREN).ru\_maxrss} after the worker exits, and
a load-only run per setting gives the baseline. A watchdog in the worker
stops it at \RSixTimeLimit\,s of method time or \RSixRSSLimit\,GiB resident
memory and records the run as over budget. Function profiles (cProfile,
grouped into the categories of Fig.~\ref{fig:r6}c by the calling function)
and tracemalloc peaks are separate runs; no time in the table comes from
them.

\emph{Checks.} Answers agree across all methods that finish, in all
\RSixAnswersEqual{} settings. In \RSixChecked{} completed checks,
\RSixCompared{} reachable sparse states were compared with the ordered
table (all of them on the \RSixFullChecked{} smallest settings, a seeded
sample of 2000 per method elsewhere), every ordered witness was checked to
be a winning action, and on the smallest settings the full compact and full
bitset tables were compared with the ordered table on the whole domain; there
was no mismatch.

\emph{Online batches.} A second workload sends \RSixBatches{} query batches
in sequence: batch $j$ asks every region at $k=j$ with one seeded random bit
history on $-D,\dots,j$ (each bit 1 with probability 0.3). The sparse methods
traverse again for each batch; the ordered method builds its table once,
inside the first batch.

\emph{Results.} Table~\ref{tab:r6} and Fig.~\ref{fig:r6} give times and
memory. Sparse bitset and sparse compact reach the same number of states in
\RSixSameReach{} settings. This is Theorem~\ref{thm:suffix}: on a non-failed
history the served offsets form a suffix, so a bitset state and its compact
pair determine each other and the two reachable sets are in bijection, so
the compact state does not reduce the number of visited states. With bitsets
stored as integer masks, the sweep shows no clear resource gain of the
compact state in sparse evaluation (times and memory in
Table~\ref{tab:r6}); this is a statement about this implementation and these
settings. The main difference appears when whole domains are stored: at $(2,4)$ the
full bitset table needs \RSixVal{fb}{2}{4}{s}\,s and \RSixVal{fb}{2}{4}{mib}\,MiB
against \RSixVal{fp}{2}{4}{s}\,s and \RSixVal{fp}{2}{4}{mib}\,MiB for full
compact, and it exceeds the budget in \RSixFullBitsetOver{} of
\RSixSettings{} settings (full compact in \RSixFullPlainOver).

On the fixed $k=0$ queries the ordered method takes
\RSixRatioMin--\RSixRatioMax{} times as long as sparse compact. It computes
generators for every state of the domain at every $k$, not only for the
states reachable from the queries, and \RSixOrdCoreMin--\RSixOrdCoreMax\,\% of
its profiled time is spent in tuple generation, state transitions and
dominance tests. Its table answers later queries at any $k$ and history by
lookup, but in this workload that at best draws level: over the
\RSixBatches{} online batches sparse compact needs \RSixOnlineSparseA,
\RSixOnlineSparseB{} and \RSixOnlineSparseC\,s in total at \RSixOnlineA,
\RSixOnlineB{} and \RSixOnlineC, against \RSixOnlineOrdA, \RSixOnlineOrdB{}
and \RSixOnlineOrdC\,s for the ordered method. Later batches start closer to
$H$, and some random histories are already failed, so each forward traversal
is shorter than the first (Fig.~\ref{fig:r6}d). The ordered method's
advantage is memory, which grows more slowly with the task count. It is
below sparse compact only at $(3,4)$ and with four tasks: at $(3,4)$ its peak
is \RSixVal{ord}{3}{4}{extra}\,MiB above load against
\RSixVal{sc}{3}{4}{extra}\,MiB, at $(4,4)$ \RSixVal{ord}{4}{4}{extra} against
\RSixVal{sc}{4}{4}{extra}\,MiB, and in the online workload
at \RSixOnlineC{} its peak is \RSixOnlineOrdMiBC{} against
\RSixOnlineSparseMiBC\,MiB. At $(3,16)$ and $(4,8)$ no region is certified
from the neutral history; every method still evaluates its recursion in
full. The code is
Python and not optimized; the ratios describe this implementation.

\begin{table*}[t]
\caption{R6: certified regions at $k=0$ (of \RSixRegions), reachable sparse
states $(k,\rho,q)$, method time and peak resident memory (including the
loaded library) per method and setting. $^*$Auxiliary whole-domain
tabulations. ``budget'': stopped at \RSixTimeLimit\,s or
\RSixRSSLimit\,GiB.}
\label{tab:r6}
\centering\footnotesize\setlength{\tabcolsep}{3.2pt}
\begin{tabular}{@{}rrcrrrrrrrrrrrr@{}}
\toprule
$n$ & $D$ & split & cert. & reach. & \multicolumn{2}{c}{sparse bitset} & \multicolumn{2}{c}{sparse compact} & \multicolumn{2}{c}{ordered} & \multicolumn{2}{c}{full compact$^*$} & \multicolumn{2}{c}{full bitset$^*$} \\
 & & & & states & s & MiB & s & MiB & s & MiB & s & MiB & s & MiB \\
\midrule
1 & 4 & hold & 358 & 85\,104 & 0.62 & 146 & 0.64 & 147 & 7.39 & 196 & 5.63 & 131 & 6.04 & 134 \\
2 & 4 & hold & 230 & 231\,187 & 1.44 & 160 & 1.48 & 160 & 9.79 & 192 & 7.63 & 149 & 63.4 & 607 \\
3 & 4 & hold & 230 & 508\,408 & 3.32 & 186 & 3.59 & 189 & 22.4 & 185 & 45.9 & 405 & \multicolumn{2}{c}{budget (time)} \\
4 & 4 & hold & 28 & 1\,181\,283 & 8.89 & 235 & 9.65 & 235 & 130 & 202 & \multicolumn{2}{c}{budget (time)} & \multicolumn{2}{c}{budget (mem)} \\
2 & 2 & dev & 230 & 242\,947 & 1.63 & 160 & 1.68 & 162 & 10.1 & 191 & 6.98 & 139 & 10.6 & 161 \\
2 & 8 & hold & 230 & 207\,667 & 1.44 & 161 & 1.47 & 159 & 12.8 & 192 & 12.0 & 177 & \multicolumn{2}{c}{budget (time)} \\
2 & 16 & hold & 150 & 160\,627 & 1.19 & 158 & 1.15 & 158 & 17.1 & 195 & 22.2 & 276 & \multicolumn{2}{c}{budget (mem)} \\
1 & 2 & dev & 358 & 87\,344 & 0.64 & 147 & 0.67 & 147 & 7.55 & 196 & 5.75 & 130 & 5.73 & 131 \\
3 & 8 & hold & 181 & 423\,736 & 2.85 & 178 & 2.98 & 178 & 35.1 & 188 & 160 & 1393 & \multicolumn{2}{c}{budget (mem)} \\
3 & 16 & hold & 0 & 255\,470 & 1.93 & 164 & 1.93 & 164 & 64.5 & 192 & \multicolumn{2}{c}{budget (time)} & \multicolumn{2}{c}{budget (mem)} \\
4 & 8 & hold & 0 & 922\,563 & 7.07 & 210 & 7.53 & 212 & 346 & 202 & \multicolumn{2}{c}{budget (time)} & \multicolumn{2}{c}{budget (mem)} \\
\bottomrule
\end{tabular}

\end{table*}

\clearpage
\bibliographystyle{plainnat}
\bibliography{references}

\end{document}